\documentclass[sigconf]{acmart}
\AtBeginDocument{%
  \providecommand\BibTeX{{%
    \normalfont B\kern-0.5em{\scshape i\kern-0.25em b}\kern-0.8em\TeX}}}

\setcopyright{acmlicensed}
\copyrightyear{2024}
\acmYear{2024}
\setcopyright{rightsretained}
\acmConference[SIGGRAPH Conference Papers '24]{Special Interest Group on Computer Graphics and Interactive Techniques Conference Conference Papers '24}{July 27-August 1, 2024}{Denver, CO, USA}
\acmBooktitle{Special Interest Group on Computer Graphics and Interactive Techniques Conference Conference Papers '24 (SIGGRAPH Conference Papers '24), July 27-August 1, 2024, Denver, CO, USA}
\acmDOI{10.1145/3641519.3657416}
\acmISBN{979-8-4007-0525-0/24/07}

\acmBooktitle{Special Interest Group on Computer Graphics and Interactive Techniques Conference Conference Papers ’24 (SIGGRAPH Conference Papers ’24), July 27–August 01, 2024, Denver, CO, USA} 
\acmISBN{979-8-4007-0525-0/24/07}

\usepackage{multirow}
\usepackage{makecell}
\usepackage{algorithm}
\usepackage{algorithmic}
\usepackage{arydshln}

\begin{document}

\title{BoostMVSNeRFs: Boosting MVS-based NeRFs to Generalizable View Synthesis in Large-scale Scenes}


\author{Chih-Hai Su}
\authornote{Authors contributed equally to the paper.}
\email{oceani.c@nycu.edu.tw}
\orcid{0009-0002-8500-9161}
\affiliation{%
  \institution{National Yang Ming Chiao Tung University}
  \country{Taiwan}
}

\author{Chih-Yao Hu}
\authornotemark[1]
\email{jj2095813@gmail.com}
\orcid{0009-0008-5968-6322}
\affiliation{%
  \institution{National Taiwan University}
  \country{Taiwan}
}

\author{Shr-Ruei Tsai}
\authornotemark[1]
\email{0410330@gmail.com}
\orcid{0009-0002-2707-0095}
\affiliation{%
  \institution{National Yang Ming Chiao Tung University}
  \country{Taiwan}
}

\author{Jie-Ying Lee}
\authornotemark[1]
\email{jayinnn.cs10@nycu.edu.tw}
\orcid{0009-0008-0826-4664}
\affiliation{%
  \institution{National Yang Ming Chiao Tung University}
  \country{Taiwan}
}

\author{Chin-Yang Lin}
\email{linjohn0903@gmail.com}
\orcid{0009-0007-1945-2014}
\affiliation{%
  \institution{National Yang Ming Chiao Tung University}
  \country{Taiwan}
}

\author{Yu-Lun Liu}
\email{yulunliu@cs.nycu.edu.tw}
\orcid{0000-0002-7561-6884}
\affiliation{%
  \institution{National Yang Ming Chiao Tung University}
  \country{Taiwan}
}

\def\eg{\emph{e.g.}}
\def\Eg{\emph{E.g.}}
\def\ie{\emph{i.e.}}
\def\Ie{\emph{I.e.}}
\def\cf{\emph{c.f.}}
\def\Cf{\emph{C.f.}}
\def\etc{\emph{etc.}} 
\def\vs{\emph{vs.}}


\newcommand{\yulunliu}[1]{{\textcolor{red}{[yulunliu: #1]}}}
\newcommand{\jayinnn}[1]{{\textcolor{blue}{[jayinnn: #1]}}}
\newcommand{\hai}[1]{{\textcolor{orange}{[hai: #1]}}}
\newcommand{\ray}[1]{{\textcolor{pink}{[ray: #1]}}}

\renewcommand{\algorithmiccomment}[1]{\hfill$\triangleright$ #1}
\renewcommand{\algorithmicrequire}{\textbf{Input:}}
\renewcommand{\algorithmicensure}{\textbf{Output:}}
\begin{abstract}
While Neural Radiance Fields (NeRFs) have demonstrated exceptional quality, their protracted training duration remains a limitation. Generalizable and MVS-based NeRFs, although capable of mitigating training time, often incur tradeoffs in quality. This paper presents a novel approach called \textbf{BoostMVSNeRFs} to enhance the rendering quality of MVS-based NeRFs in large-scale scenes. We first identify limitations in MVS-based NeRF methods, such as restricted viewport coverage and artifacts due to limited input views. Then, we address these limitations by proposing a new method that selects and combines multiple cost volumes during volume rendering. Our method does not require training and can adapt to any MVS-based NeRF methods in a feed-forward fashion to improve rendering quality. Furthermore, our approach is also end-to-end trainable, allowing fine-tuning on specific scenes. We demonstrate the effectiveness of our method through experiments on large-scale datasets, showing significant rendering quality improvements in large-scale scenes and unbounded outdoor scenarios. 
We release the source code of BoostMVSNeRFs at {\urlstyle{tt}\url{https://su-terry.github.io/BoostMVSNeRFs}}.
\end{abstract}



\begin{teaserfigure}
{\Large\urlstyle{tt}\url{https://su-terry.github.io/BoostMVSNeRFs}}\\[3mm]
\vspace{3mm}
\includegraphics[width=1.0\textwidth]{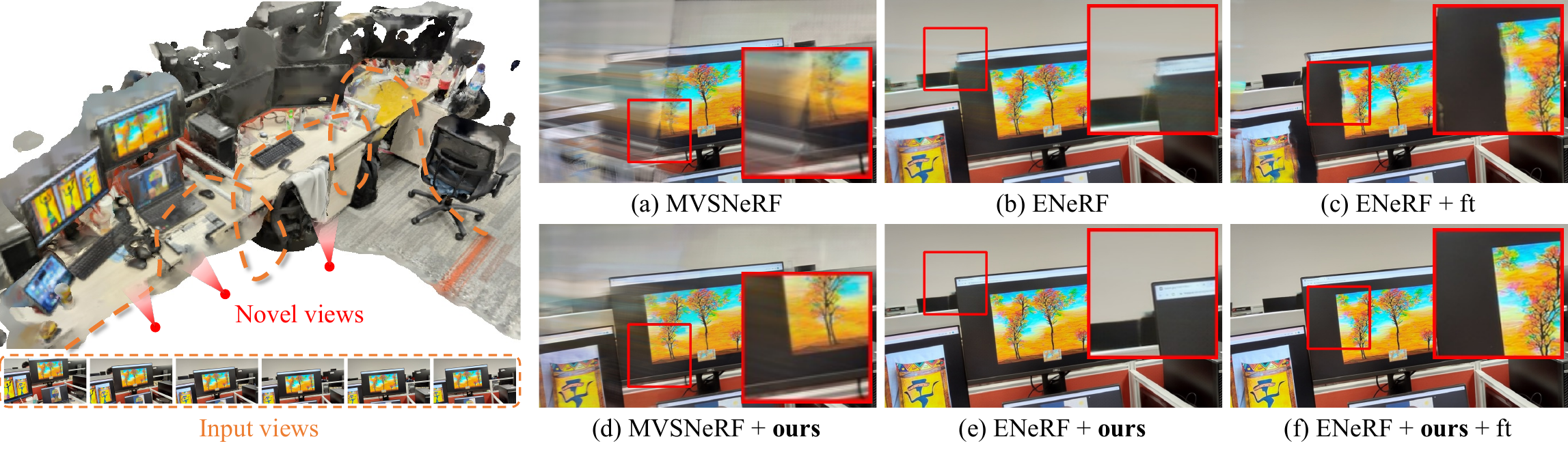}
  \vspace{-10mm}
  \caption{\textbf{Our BoostMVSNeRFs enhances the novel view synthesis quality of MVS-based NeRFs in large-scale scenes.} 
  MVS-based NeRF methods often suffer from (a) limited viewport coverage from novel views or (b) artifacts due to limited input views for constructing cost volumes. (c) These drawbacks cannot be resolved even by per-scene fine-tuning. Our approach selects those cost volumes that contribute the most to the novel view and combines multiple selected cost volumes with volume rendering. (d, e) Our method does not require any training and is compatible with existing MVS-based NeRFs in a feed-forward fashion to improve the rendering quality. (f) The scene can be further fine-tuned as our method supports end-to-end fine-tuning.}
  \label{fig:teaser}
\end{teaserfigure}


\maketitle

\section{Introduction}
%
%
In computer vision, 3D reconstruction and novel view synthesis are crucial, with widespread applications from photogrammetry to AR/VR. Traditional methods relied on photo-geometry for 3D scene reconstruction using meshes. Recently, the task of novel view synthesis has advanced drastically since the emergence of the Neural Radiance Field (NeRF) and its variants \cite{mildenhall2021nerf,barron2021mip,barron2022mip,tancik2022block,Meuleman_2023_CVPR,barron2023zip,cheng2024improving}. NeRF encodes 3D information into a Multi-layer Perceptron (MLP) network to represent a scene. Despite such methods providing photorealistic rendering quality, these models have a huge downside as they require per-scene training with a long training time.

Recent advances in Generalizable NeRFs \cite{cao2022fwd,chen2021mvsnerf,wang2021ibrnet,xu2022point,yu2021plenoxels,yu2021pixelnerf} improve scene adaptation by extracting input image features via 2D CNNs and utilizing large datasets for training, allowing for rapid scene adaptation and enhanced rendering through fine-tuning. 
MVS-based methods such as MVSNeRF \cite{chen2021mvsnerf} and ENeRF \cite{lin2022efficient} synthesize high-quality novel views by constructing cost volumes from a few input images, leveraging 3D CNNs and volume rendering in a feed-forward fashion. However, they are constrained by using a fixed number of input views and often struggle to reconstruct large-scale and unbounded scenes, resulting in padding artifacts at image boundaries (Fig. \ref{fig:teaser}(a)) and wrongly reconstructed geometry in disocclusion regions (Fig. \ref{fig:teaser}(b)). Furthermore, these issues could hardly be resolved by per-scene fine-tuning (Fig. \ref{fig:teaser}(c)).

To address the problems, we propose BoostMVSNeRFs, a pipeline that is compatible with any MVS-based NeRFs to improve their rendering quality in large-scale and unbounded scenes. 
We first present 3D visibility scores for each sampled 3D point to indicate the proportion of contributions from individual input views. We then volume render the 3D visibility scores into 2D visibility masks to determine the contribution of each cost volume to the target novel view. 
Next, we combine multiple cost volumes during volume rendering to effectively expand the coverage of the novel view viewport and reduce artifacts by constructing more consistent geometry and thus alleviate the aforementioned MVS-based NeRFs' issues. 
Additionally, to optimize the novel view visibility coverage, we further propose a greedy algorithm to approximate the optimal support cost volume set selection for the multiple-cost volume combined rendering.
Our proposed pipeline is compatible with any MVS-based NeRFs to improve their rendering quality (Fig. \ref{fig:teaser}(d, e)) and is end-to-end trainable. Therefore, our method also inherits this property from MVS-based NeRFs and can be fine-tuned to a specific scene to further improve the rendering quality (Fig. \ref{fig:teaser}(f)).

We conduct experiments on two large-scale datasets, Free~\cite{wang2023f2} and ScanNet~\cite{dai2017scannet} datasets, which contain unbounded scenes with free camera trajectories and large-scale indoor scenes with complex structures, respectively.
Experiments demonstrate that our proposed method performs favorably against other per-scene training or generalizable NeRFs in different dataset scenarios. Most importantly, our method is able to improve any MVS-based NeRF rendering quality through our extensive experiments, especially in free camera trajectories and unbounded outdoor scenes, which are the most common use cases in real-world applications. 
\section{RELATED WORK}
\paragraph{Novel view synthesis}
Novel view synthesis is a core challenge in computer vision, addressed through various techniques like image-based rendering~\cite{chaurasia2013depth,flynn2016deepstereo,kalantari2016learning,penner2017soft,riegler2020free} or multiplane image (MPI)~\cite{zhou2018stereo,flynn2019deepview,li2020crowdsampling,mildenhall2019local,srinivasan2019pushing,tucker2020single}, and explicit 3D representations, including
meshs~\cite{debevec2023modeling,thies2019deferred,waechter2014let,wood2023surface}, voxels~\cite{sitzmann2019deepvoxels,lombardi2019neural,lombardi2021mixture}, point clouds~\cite{aliev2020neural,xu2022point}, depth maps~\cite{dhamo2019peeking,gortler1998layered,shih20203d,tulsiani2018layer,hedman2021baking}.
Recently, neural representations~\cite{jiang2020sdfdiff,shih20203d,liu2019neural,lombardi2019neural,sitzmann2019deepvoxels,wizadwongsa2021nex,zhou2018stereo}, particularly Neural Radiance Fields (NeRF)~\cite{mildenhall2021nerf,barron2021mip,barron2022mip,tancik2022block,Meuleman_2023_CVPR,barron2023zip,park2021nerfies,zhang2020nerf++}, have achieved photorealistic rendering by representing scenes with continuous fields.
Despite the advancements in areas like relighting~\cite{boss2021nerd,munkberg2022extracting,zhang2021physg,zhang2021nerfactor,yu2021plenoctrees,yu2021pixelnerf}, dynamic scenes~\cite{li2022neural,xian2021space,park2021nerfies,pumarola2021d,liu2023robust}, and multi-view reconstruction~\cite{oechsle2021unisurf,wang2021neus,yariv2021volume,yariv2020multiview}, these methods although speed up training using hash grid~\cite{muller2022instant} or voxel~\cite{chen2022tensorf,sun2022direct} as representations, still require intensive per-scene optimization, thus limiting their generalizability. 
In contrast, our generalizable approach balances rendering quality and speed through feed-forward inference efficiently.

\paragraph{Multi-view stereo and generalizable radiance fields}
Neural Radiance Fields (NeRF) offer photorealistic rendering but are limited by costly per-scene optimization. 
Recently, generalizable NeRFs~\cite{cao2022fwd,chen2021mvsnerf,wang2021ibrnet,xu2022point,yu2021plenoxels,yu2021pixelnerf} provide efficient approaches to synthesize novel views without per-scene optimization. 
Techniques like PixelNeRF \cite{yu2021pixelnerf} and IBRNet \cite{wang2021ibrnet} merge features from adjacent views for volume rendering, while PointNeRF \cite{xu2022point} constructs point-based fields for this purpose.
Multi-view stereo (MVS) methods estimate depth using cost volumes~\cite{oechsle2021unisurf}, with MVSNet~\cite{yao2018mvsnet} utilizing 3D CNNs for feature extraction and cost volume construction, enabling end-to-end training and further novel view synthesis. 
Despite amazing results from learning-based MVS, these methods are memory-intensive, prompting innovations like plane sweep \cite{yao2019recurrent} and coarse-to-fine strategies \cite{gu2020cascade,chen2019point,yu2020fast} for efficiency.
Other works, such as MVSNeRF \cite{chen2021mvsnerf}, ENeRF \cite{lin2022efficient} and Im4D~\cite{lin2023im4d}, further bridge MVS methods with NeRF, introducing volumetric representations and depth-guided sampling for speed and dynamic reconstruction. 
Although these works advance the performance of generalizable NeRF, their rendering qualities are hindered by the limited visibility coverage of a single cost volume, leading to poor synthesis quality and visible padding artifacts near the image boundaries on large-scale or unbounded scenes. Additional research endeavors have been suggested to address these challenges. For instance, GeoNeRF~\cite{johari2022geonerf}) introduces a novel approach to handle occlusions, while Neural Rays~\cite{liu2022neural} presents an occlusion-aware representation aimed at mitigating this problem. Although these methods tackle occlusions issues, the view coverage problem originated from MVS-based methods still exists. Our method overcomes this issue by selecting and combining multiple cost volumes to improve coverage and rendering confidence, enhancing the performance and robustness of MVS-based NeRF methods without any cost compared with previous methods.

\paragraph{Few-shot NeRFs}
Prior work utilized mainly two different approaches to reconstruct scenes with sparse input views~\cite{kim2022infonerf}: introducing regularization priors and training generalized model. Regularization-based methods~\cite{deng2022depth,niemeyer2022regnerf,somraj2023vip,roessle2022dense,uy2023scade,wang2023sparsenerf,wynn2023diffusionerf,jain2021putting,yang2023freenerf,seo2023mixnerf,zhu2023vdn,wu2023reconfusion} such as Vip-NeRF~\cite{somraj2023vip} attempt to tackle this problem by obtaining visibility prior to regularize the scenes' relative depth. Training generalized models~\cite{yu2021pixelnerf,wang2021ibrnet,chen2021mvsnerf,trevithick2021grf,chibane2021stereo,johari2022geonerf,shi2022garf,chen2023explicit,lin2023vision} on large datasets such as MVSNeRF~\cite{chen2021mvsnerf} constructs cost volume to gain cross-view insight to tackle this goal. 
Different from this line of work, we present a novel visibility mask in a 3D fashion and serve as a visibility score to blend features while performing volume rendering. 

\paragraph{Radiance fields fusion}
Recently, several works propose to tackle scene fusion and intend to achieve large-scale reconstruction. NeRFusion~\cite{zhang2022nerfusion} performs sequential data fusion on voxels with GRU on the image level. SurfelNeRF~\cite{gao2023surfelnerf} fuses scenes after converting them to surfels~\cite{pfister2000surfels} representation. Our approach seamlessly integrates cost volume without requiring training, thereby harnessing the capabilities of all MVS-based pre-trained models. Instead of concentrating solely on large-scale fusion, our method functions as a readily applicable tool to enhance various cost volume-based MVS applications.
\section{Method}
Given multi-view images in an unbounded scene, the same as other MVS-based NeRF methods~(Sec.~\ref{sec:preliminaries}), our task is to synthesize novel view images without per-scene training.
In order to tackle limited viewport coverage from a single cost volume created by a fixed number of few (\eg{}, 3) input images, we propose \emph{BoostMVSNeRFs}, an algorithm to consider multiple cost volumes while rendering.
We first introduce a 3D visibility score for each sampled 3D point, which is used to render volume into 2D visibility masks (Sec.~\ref{sec:visibility_mask}).
Given a rendered 2D visibility mask for each cost volume, we combine multiple cost volumes in a support set to render novel views (Sec.~\ref{sec: render}).
Finally, we present a greedy algorithm to iteratively select cost volumes and update the support set to maximize the viewport coverage and confidence of novel views (Sec.~\ref{sec:CV_select}). 
Our pipeline is end-to-end trainable and thus can be fine-tuned on a new scene (Sec.~\ref{sec:end_to_end}).
Our method is model-agnostic and applicable to any MVS-based NeRFs to boost the rendering quality.

\subsection{MVS-based NeRFs Preliminaries} \label{sec:preliminaries}
Given multi-view images with camera parameters, MVS-based NeRFs \cite{chen2021mvsnerf,lin2022efficient,gu2020cascade} use a shared 2D CNN to extract features for input images.
Then, following MVSNet~\cite{yao2018mvsnet}, we construct a feature volume by warping the input features into the target view. The warped features would be used to construct the encoding volume by computing the variance of multi-view features.
Next, we apply a 3D CNN to regularize the encoding volume to build the cost volume $\text{CV}$ to smooth the noise in the feature volume.
Given a novel viewpoint, we query the color $c$ and density $\sigma$ using an MLP with sampled 3D point coordinates $x$, viewing directions $v$, trilinear interpolated cost volume values at location $p$, and projected colors from input views $\text{\bf{C}}_\text{in}$ as input:
\begin{equation}
( c, \sigma ) = \text{MLP}_\theta ( p, v, \text{CV}(p), \text{\bf{C}}_\text{in} ),
\end{equation}
where $\theta$ denotes the parameter of the MLP.
Finally, we can volume render along rays to get the pixel colors in novel views.

The volume rendering equation in NeRF or MVSNeRF is evaluated by differentiable ray marching for novel view synthesis. A pixel color is computed by accumulating sample point values through ray marching. Here we consider a given ray $\textbf{r}$ from the camera center $o$ through a given pixel on the image plane as $\textbf{r}=o+u_jd$, where $d$ is the normalized viewing direction, and $u_j$ is the quadrature point constrained within the bounds of the near plane $u_n$ and the far plane $u_f$. The final color is given by:  
\begin{equation} \label{eq:original}
C(\textbf{r})=\sum_{j=1}^J T(j) \alpha(\sigma_j \delta_j )c_{j},\
\end{equation}
where $T(j)=\exp(-\sum_{s=1}^{j-1}\sigma_{s}\delta_s)$ is the accumulated transmittance, \( \alpha(x) = 1 - \exp(-x) \) is the opacity of the point, and \( \delta_j = u_{j+1} - u_j \) is the distance between two quadrature points.


The existing MVS-based NeRFs only utilize a single cost volume from a few viewpoints (\eg{} 3 input views). As a result, these methods often fall into limited viewport coverage, wrong geometry, and rendering artifacts (Fig.~\ref{fig:teaser}(a, b)). To overcome these problems, a naive solution would be training another MVS-based NeRF with more input views to construct the cost volume. Nevertheless, this solution requires training a new model with larger memory consumption, but even so, the input views could still be insufficient in inference time. Therefore, we proposed a novel method considering multiple cost volumes while rendering novel views.






\begin{figure}[t]
\centering
\includegraphics[width=1.0\columnwidth]{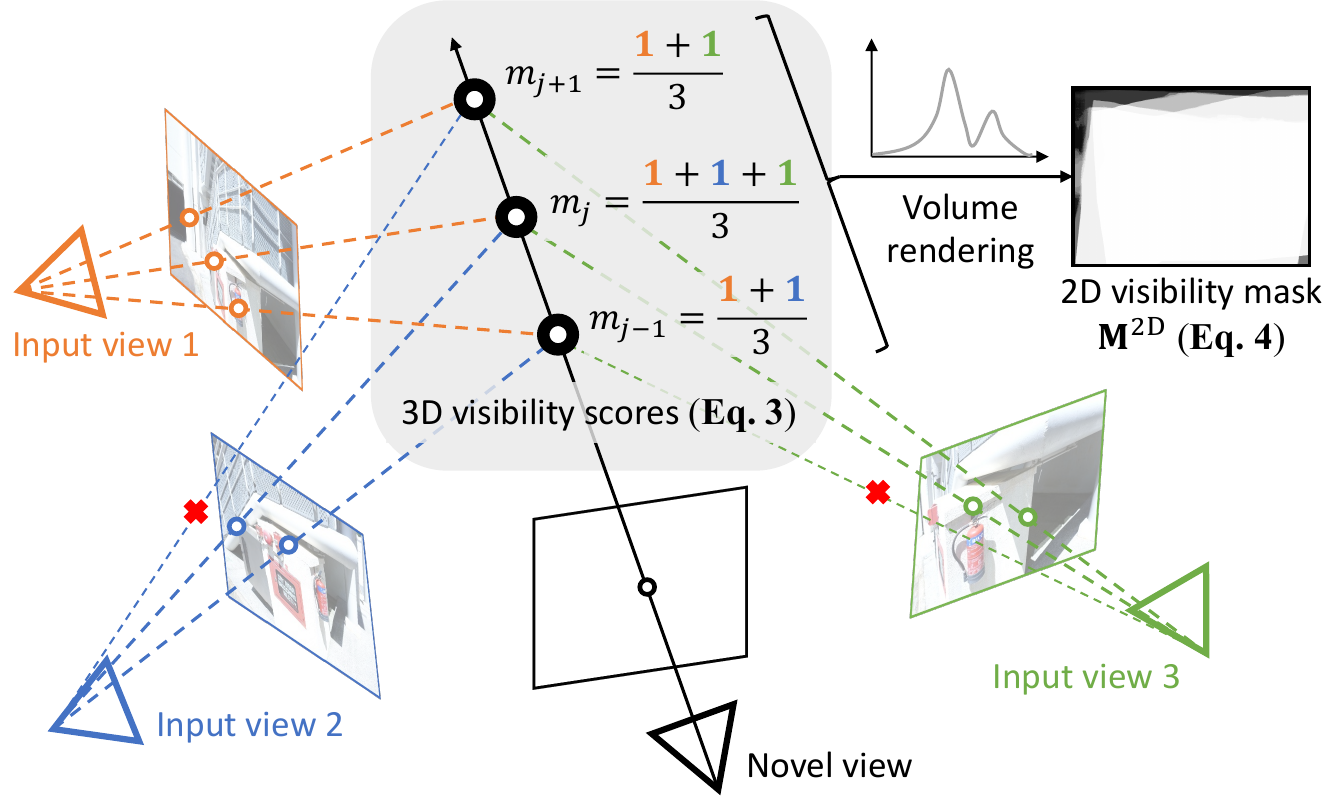}
\caption{\textbf{3D visibility scores and 2D visibility masks.}
For a novel view, depth distribution is estimated from three input views, from which 3D points are sampled and projected onto each view to determine visibility. 
These projections yield 3D visibility scores $m_j$, normalized across the views, and are subsequently volume rendered into a 2D visibility mask $\textbf{M}^{\text{2D}}$. 
This mask highlights the contribution of each input view to the cost volume and guides the rendering process, aiding in the selection of input views that optimize rendering quality and field of view coverage.}
%
\label{fig:visibility_mask}
\end{figure}

\subsection{3D Visibility Scores and 2D Visibility Masks} \label{sec:visibility_mask}

By taking $I$ reference views into account in constructing a single cost volume, the maximum number of cost volumes we can refer to is $C^N_I=\binom{N}{I}=\frac{N(N-1) \cdots (N-I+1)}{I(I-1) \cdots 1}$ for each target view, where $N$ is the number of reference views. However, utilizing all cost volumes results in high memory consumption and also leads to inefficient rendering. To tackle this challenge, we propose a method to select those cost volumes with the largest contribution to viewport coverage and potential enhancement of rendering quality for novel views. To evaluate the contribution of each cost volume, we present \emph{multi-view 3D visibility scores} as a metric.

For each sample point in a cost volume, we calculate its corresponding 3D visibility scores (the gray-shaded part in Fig.~\ref{fig:visibility_mask}). These scores quantify the level of observation from various cost volumes, serving as a measurement of visibility. To calculate the 3D visibility scores of a single cost volume in a rendered view, we sample rays from the rendered view and aggregate the visibility weight from the reference views. Let $I$ represent the total number of reference views. We use $\vmathbb{1}_i(p)$ to indicate whether a sample point $p$ is in the viewport of reference view $i$ (bottom part in Fig.~\ref{fig:visibility_mask}). The 3D visibility scores $m_j$ are calculated using the formula:
\begin{equation} \label{eq:3D_mask}
m_j = \frac{\sum_{i=1}^{I} \vmathbb{1}_i(p)}{I},
\end{equation}
where the subscript $j$ denotes the sampled 3D point index along the ray, and the output 3D visibility scores range from 0 to 1. Each point on the mask indicates its 3D visibility score, with larger values reflecting higher confidence in the information at a specific sample point. The visibility score can be utilized as the weight for the feature of a point on a specific cost volume. Therefore, with the 3D visibility scores, we can combine the results from different cost volumes when volume rendering. 

After obtaining 3D visibility scores for each cost volume, we propose the \textit{2D visibility mask}. The 2D visibility is constructed by volume rendering the 3D metrics scores to novel view, as shown in Fig. \ref{fig:visibility_mask}. Similar to Eq. \ref{eq:original}, given ray $\textbf{r}$ from the camera center $o$ with direction $d$, the value of 2D visibility mask is given by:

\begin{equation} \label{eq:2D_mask}
    \textbf{M}^{\text{2D}}(\textbf{r}) = \sum_{j=1}^J T'(j) \alpha\left(m_j \delta_j \right)m_{j}, 
\end{equation}
where $T'(j)=\exp(-\sum_{s=1}^{j-1} m_s\delta_s)$ is the transmitte considering 3D visibility scores. 
The 2D visibility mask will be used in cost volume selection; we will thoroughly discuss it in Sec. \ref{sec:CV_select}.


%

\begin{figure}[t]
\centering
\includegraphics[width=1.0\columnwidth]{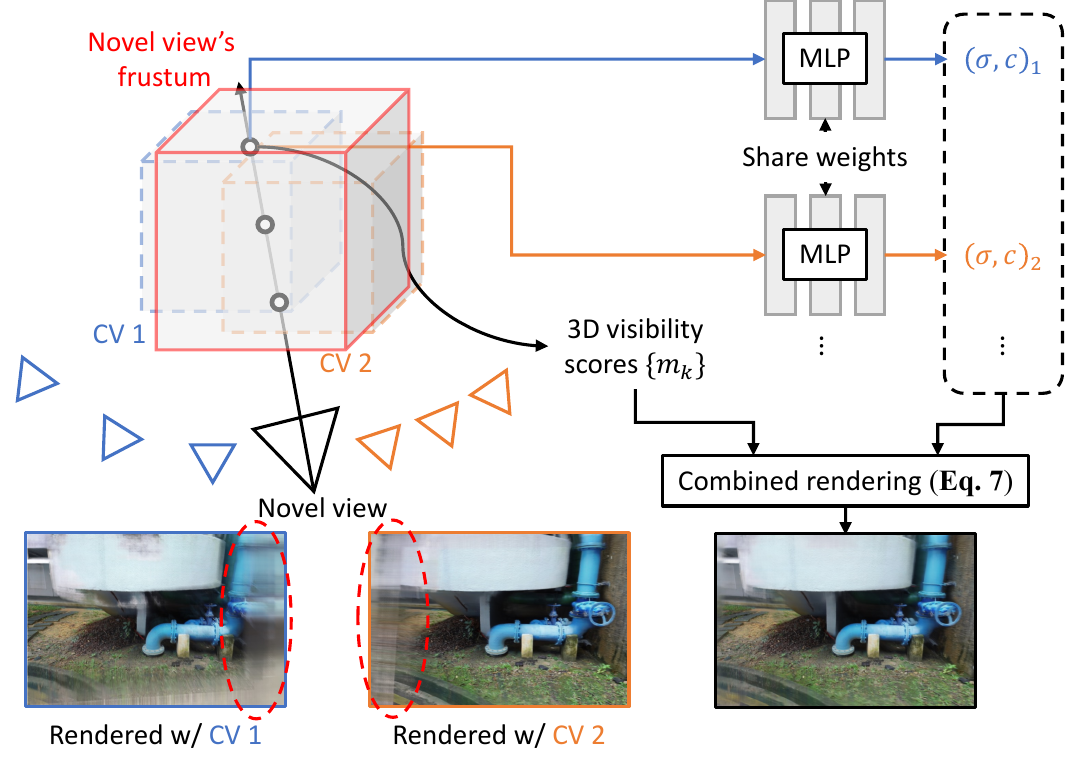}
\caption{\textbf{Combined rendering from multiple cost volumes.} 
Using a single cost volume, as in traditional MVS-based NeRFs, often introduces padding artifacts or incorrect geometry, as indicated by the \textcolor{red}{\textbf{red}} dashed circles. Our method warps selected cost volumes to the novel view's frustum and applies 3D visibility scores $m_j$ as weights to blend multiple cost volumes during volume rendering. Combined rendering provides broader viewport coverage and combines information from multiple cost volumes, leading to improved image synthesis and alleviating artifacts.}
\label{fig:combined_rendering}
\end{figure}

\subsection{Rendering by Combining Multiple Cost Volumes \label{sec: render}}
Our proposed rendering differs from the traditional one (Eq. \ref{eq:original}) by considering 3D visibility scores and combining multiple cost volumes.
Below, we explain the modifications we make.
First, let us only consider a single cost volume for simplicity. The pixel color output by considering only a single cost volume is given by:
\begin{equation} \label{eq:single_cv}
    C_{\text{single}}(\textbf{r})= \sum_{j=1}^{J} T_{\text{single}}(j) \alpha\left(\sigma_j \delta_j \right) m_j c_j, 
\end{equation}
\begin{align}
    T_{\text{single}}(j) &= \exp \left( - \sum_{s=1}^{j-1} \left( \sigma_s \delta_s - \ln m_s \right)\right).
\end{align}
Please refer to the supplementary material for the derivation of the transmittance considering only a single cost volume $T_{\text{single}}(j)$.

To further consider multiple cost volumes and also utilize their corresponding 3D visibility scores, we modify Eq. \ref{eq:single_cv} to combine the result across multiple cost volumes. The final proposed volume rendering is given by:

\begin{equation} \label{eq:combine}
C(\textbf{r}) = \sum_{k=1}^K \sum_{j=1}^{J} T_{\text{combined}}(j) \alpha\left(\sigma^k_j \delta_j \right) M^k_j c^k_j, 
\end{equation}

\begin{equation}
T_{\text{combined}}(j) = \sum_{k=1}^K \exp \left( - \sum_{s=1}^{j-1} \left( \sigma_s^k \delta_s - \ln M_s^k \right)\right),
\end{equation}
%
where $K$ is the number of selected cost volumes,
and $M_j^k = \frac{m^k_j}{\sum_{k=1}^K m^k_j}$ is the normalized 3D visibility score so that the summation of 3D visibility scores over selected cost volumes equals 1.

The illustration and effect of combining multiple cost volumes in rendering is shown in Fig.~\ref{fig:combined_rendering}. 
Existing MVS-based NeRFs use a single cost volume to render novel views that contain padding artifacts and wrong geometry.
Combining multiple cost volumes in rendering alleviates these artifacts and broadens the viewport coverage of novel views, thus improving the rendering quality.



\subsection{Support Cost Volume Set Selection} \label{sec:CV_select}
As mentioned in Sec. \ref{sec: render}, we only select $K$ cost volumes for combined rendering to optimize rendering efficiency. Ideally, combining selected $K$ cost volumes should provide maximum coverage for the rendered view. This problem can be formulated as \emph{maximum coverage problem}, which is NP-hard. Thus, to complete view selection in polynomial time, we propose a greedy algorithm to construct a support set $\mathbf{S}$ of $K$ cost volumes in \textbf{Algorithm~\ref{alg:view_selection}}. Nemhauser \textit{et al.} \cite{nemhauser1978analysis} also proved that the greedy algorithm is the optimal algorithm in polynomial time.





\begin{algorithm}[t]
\small
\caption{Support cost volume set selection algorithm}\label{alg:view_selection}
\begin{algorithmic}[1]
\REQUIRE $\{\text{\bf{CV}}_n\}_{n=1}^N$: $N$ candidate cost volumes
\REQUIRE $\{\text{\bf{M}}^{\text{2D}}_n\}_{n=1}^N$: 2D visibility masks
\ENSURE $\mathbf{S}$: a support set of $K$ cost volumes

\STATE $\mathbf{S} \gets \varnothing$ \COMMENT{Initialize the support CV set as an empty set}
\STATE $\text{\bf{P}}_0 \gets \text{2D Mask filled with ones}$ \COMMENT{Initialize the view coverage}
\WHILE{$\left | \mathbf{S} \right | < K$}
    \STATE $\text{best\_idx} \gets 0$
    \STATE $\text{max\_ratio} \gets 0$
    \STATE $i \gets 1$ \COMMENT{Initialize selection iteration}
    \WHILE{$i \leq N$}
        \IF[Consider remaining views only]{$\text{\bf{CV}}_i \not\in \mathbf{S}$} 
            \STATE $\text{ratio} \gets \sum{(\text{\bf{P}}_{i-1} \cdot \text{\bf{M}}^\text{2D}_{i})} $
            \IF {$\text{ratio} > \text{max\_ratio}$}
                \STATE $\text{max\_ratio} \gets \text{ratio}$
                \STATE $\text{best\_idx} \gets i$
            \ENDIF
        \ENDIF
        \STATE $i \gets i+1$
    \ENDWHILE
    \STATE $\text{\bf{P}}_i \gets \text{\bf{P}}_{i-1} \cdot (1 - \text{\bf{M}}^\text{2D}_{{\text{best\_idx}}})$ \COMMENT{Update the view coverage}
    \STATE $\mathbf{S} \gets \mathbf{S} \cup \{\text{\bf{CV}}_\text{best\_idx}\}$ \COMMENT{Add the best CV to the set}
\ENDWHILE
\end{algorithmic}
\end{algorithm}

\begin{figure}[t]
\centering
\includegraphics[width=1.0\columnwidth]{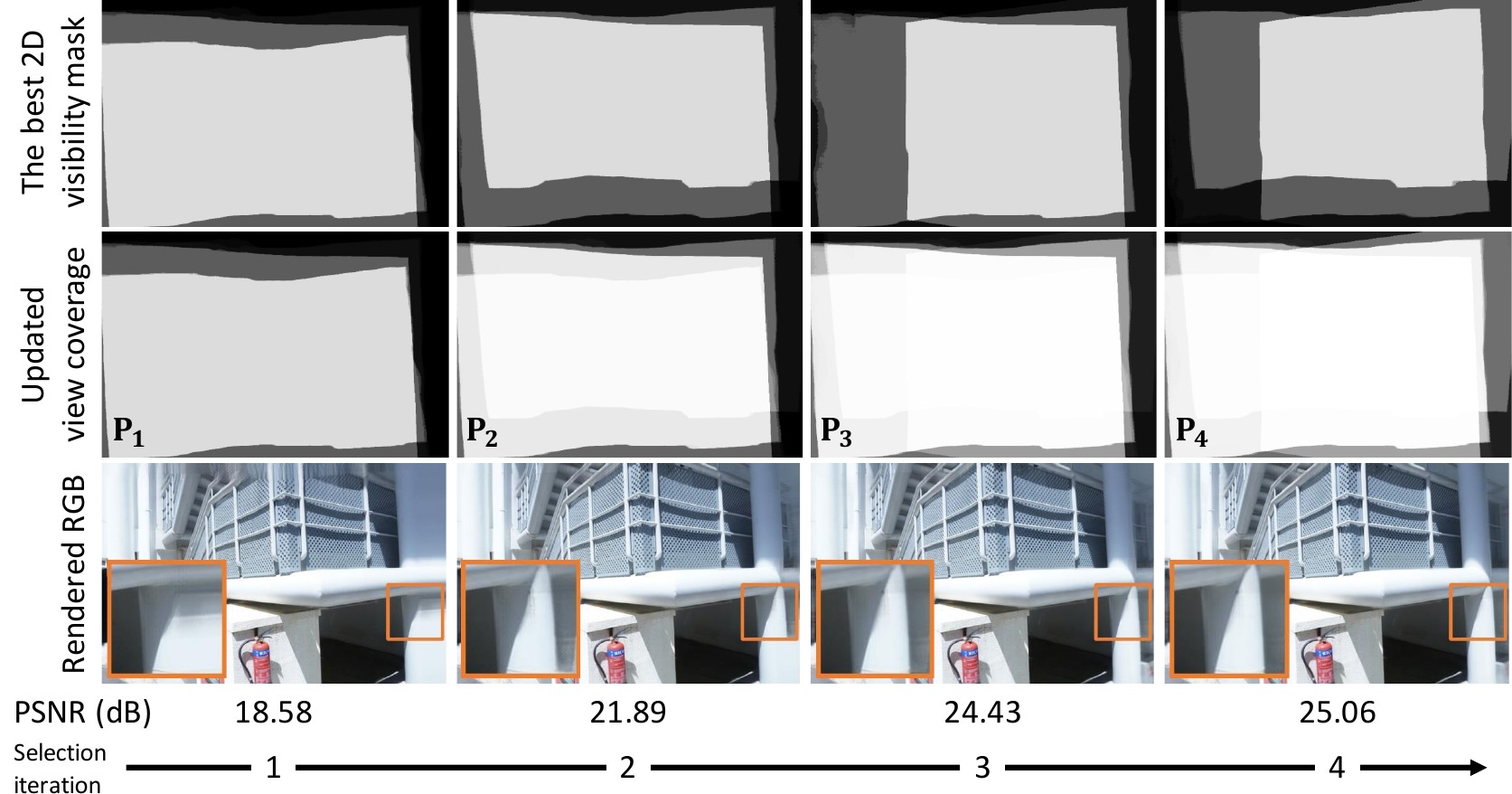}
\caption{\textbf{Support cost volume set selection.} 
Initially, our greedy algorithm selects a single cost volume, providing maximum coverage yet insufficient to prevent padding artifacts (\textcolor{orange}{\textbf{orange}} boxes). Subsequent iterations incorporate additional cost volumes, progressively expanding view coverage, and improving image quality, as indicated by the increasing PSNR values.
}
\label{fig:progressive_selection}
\end{figure}

We show an example of the proposed selection algorithm in Fig.~\ref{fig:progressive_selection}. At the beginning of the algorithm, our method selects the cost volume with the largest coverage score of the corresponding 2D visibility mask.
The rendered image contains padding artifacts near the image boundaries as the viewport of this single cost volume is limited.
Later on, our selection algorithm gradually selects the cost volumes that could maximize the visibility coverage and, therefore, enlarge the valid region of the rendered view.
As a result, the rendering quality of novel views progressively grows as more cost volumes are selected and combined in the volume rendering.

\subsection{End-to-end Fine-tuning} \label{sec:end_to_end}

Our method is compatible with any MVS-based NeRFs
to boost the rendering quality. Moreover, our approach is not optimized for a specific scene and could be generalized to new scenes, allowing it to enhance any end-to-end fine-tunable model. Fine-tuning refines geometry and color consistency within cost volumes and eliminates padding artifacts through combined rendering from multiple cost volumes. Thus, our method could augment the capabilities of advanced MVS-based NeRFs beyond ENeRF and MVS-NeRF. Please refer to the supplementary material for the fine-tuning details.

\section{Experiments}

\subsection{Experimental Settings}
\paragraph{Datasets.}

We evaluate two datasets: (1) the Free dataset collected by F2-NeRF~\cite{wang2023f2} and (2) the ScanNet~\cite{dai2017scannet} dataset. The Free dataset consists of seven challenging scenes featuring narrow, long camera trajectories and focused foreground objects. Our evaluations on the Free dataset follow the train/test split in F2-NeRF~\cite{wang2023f2} by using one-eighth of the images for testing and the rest for training. As for the ScanNet dataset, we strictly follow the train/test splits as defined in NeRFusion~\cite{zhang2022nerfusion}, NerfingMVS~\cite{wei2021nerfingmvs}, and SurfelNeRF~\cite{gao2023surfelnerf}, with eight large-scale indoor scenes.
We assess the rendering quality with PSNR, SSIM~\cite{wang2004image}, and LPIPS~\cite{zhang2018unreasonable} metrics. 

\paragraph{Baselines.}
We compare BoostMVSNeRFs with various state-of-the-art NeRFs, including fast per-scene optimization NeRFs such as F2-NeRF~\cite{wang2023f2} and Zip-NeRF~\cite{barron2023zip} and generalizable NeRFs such as MVSNeRF~\cite{chen2021mvsnerf}, 
ENeRF~\cite{lin2022efficient} and SurfelNeRF~\cite{gao2023surfelnerf}.

In particular, F2-NeRF excels in outdoor scenes with free camera trajectories.
Our method employs cost volume representations similar to MVSNeRF and ENeRF but enlarges valid visible regions by fusing multiple cost volumes.
Although SurfelNeRF also proposes fusing multiple surfels as a type of 3D representation, the fusion method and its scene representation differ from BoostMVSNeRFs.
To ensure fairness, we used the same experimental settings as in previous studies and used official codes where possible.
All the training, fine-tuning, and evaluations are done on a single RTX 4090 GPU.

Our method is compatible with MVS-based techniques, allowing us to employ pre-trained models such as MVSNeRF and ENeRF in our experiments. 
Unless otherwise specified, we use ENeRF as our backbone MVS-based NeRF method in all the experiments. We optimize the parameters, $N = 6$, $I = 3$, and $K = 4$, for efficient rendering and high quality. Our method achieves similar runtime performance in rendering and fine-tuning as other generalizable NeRF methods but renders significantly improved quality.
Additional implementation details of the proposed BoostMVSNeRFs, such as the sensitivity analysis of the number of selected cost volumes $K$ and the numerical representation (continuous or binary) of 2D visibility masks, are provided in the supplementary material.

\subsection{Comparison with State-of-the-art Methods}


\begin{table}[t]
\centering
\small
\caption{\textbf{Quantitative comparisons with state-of-the-art methods on the Free~\cite{wang2023f2} dataset.}}
\label{tab:quantitative_free}
\resizebox{\columnwidth}{!}{%
\begin{tabular}{lccccc}
\toprule
Method & Setting & PSNR $\uparrow$ & SSIM $\uparrow$ & LPIPS $\downarrow$ & FPS $\uparrow$ \\
\midrule
MVSNeRF~\cite{chen2021mvsnerf} & \multirow{4}{*}{\makecell{No per-scene\\optimization}} & 20.06 & 0.721 & 0.469 & 1.79 \\
MVSNeRF + Ours &  & 20.52 & 0.722 & 0.470 & 1.26 \\
ENeRF~\cite{lin2022efficient} &  & 23.24 & 0.844 & 0.225 & \textbf{9.90} \\
ENeRF+Ours &  & \textbf{24.21} & \textbf{0.862} & \textbf{0.218} & 5.51 \\
\midrule
F2-NeRF~\cite{wang2023f2} & \multirow{6}{*}{\makecell{Per-scene\\optimization}} & 25.55 & 0.776 & 0.278 & 3.75 \\
Zip-NeRF~\cite{barron2023zip} & & 25.90 & 0.772 & 0.241 & 0.66 \\
$\text{MVSNeRF}_\text{ft}$~\cite{chen2021mvsnerf} & & 20.49 & 0.698 & 0.425 & 1.79 \\
MVSNeRF + $\text{Ours}_\text{ft}$ &  & 21.59 & 0.759 & 0.265 & 1.26 \\
$\text{ENeRF}_\text{ft}$~\cite{lin2022efficient} &  & 25.19 & 0.880 & 0.180 & \textbf{9.90} \\
$\text{ENeRF+Ours}_\text{ft}$ &  & \textbf{26.14} & \textbf{0.894} & \textbf{0.171} & 5.51 \\
\bottomrule
\end{tabular}%
}
\end{table}

\begin{figure*}[t]
\centering
\small
\setlength{\tabcolsep}{1pt}
\renewcommand{\arraystretch}{1}
\resizebox{1.0\textwidth}{!} 
{
\begin{tabular}{c|cc|cccc}
& \multicolumn{2}{c|}{No per-scene optimization} & \multicolumn{4}{c}{Per-scene optimization} \\
\includegraphics[width=0.143\textwidth]{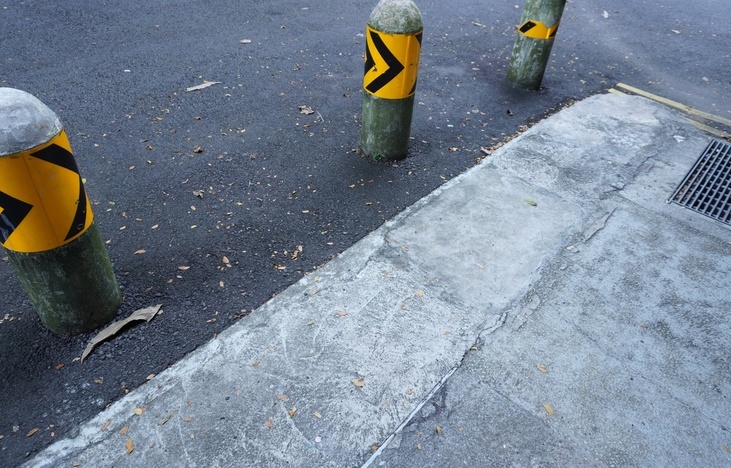} & \includegraphics[width=0.143\textwidth]{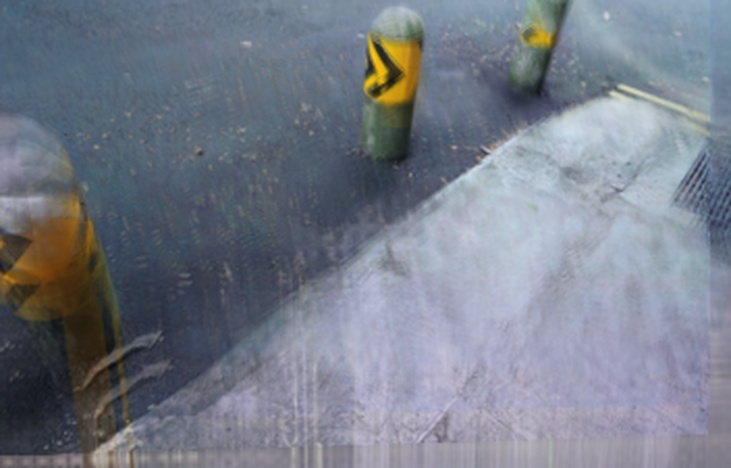} & \includegraphics[width=0.143\textwidth]{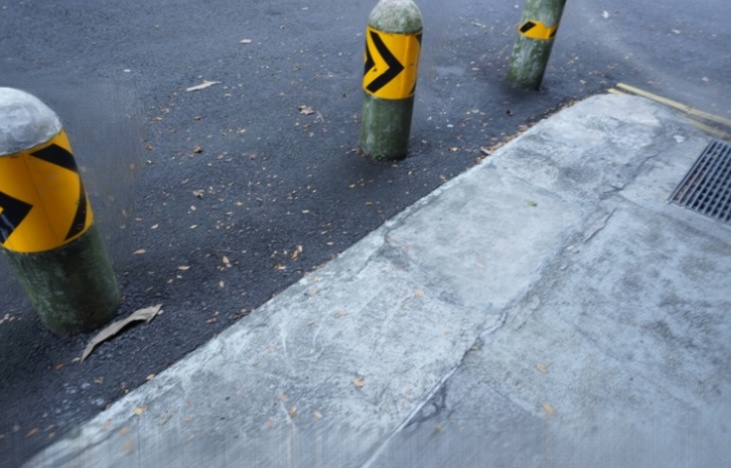} & \includegraphics[width=0.143\textwidth]{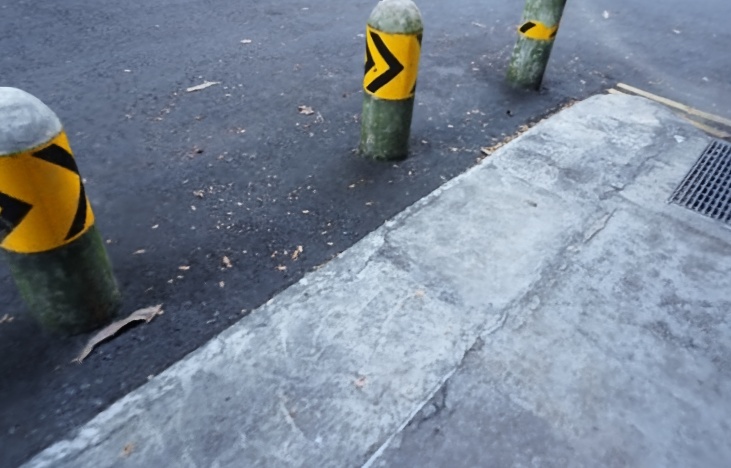} & \includegraphics[width=0.143\textwidth]{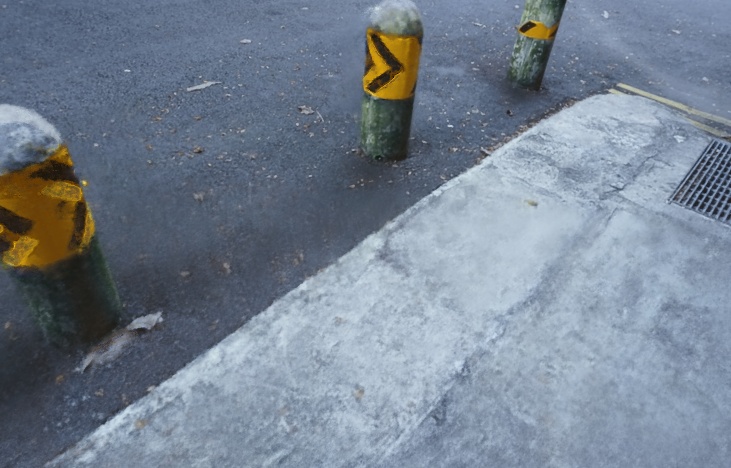} & \includegraphics[width=0.143\textwidth]{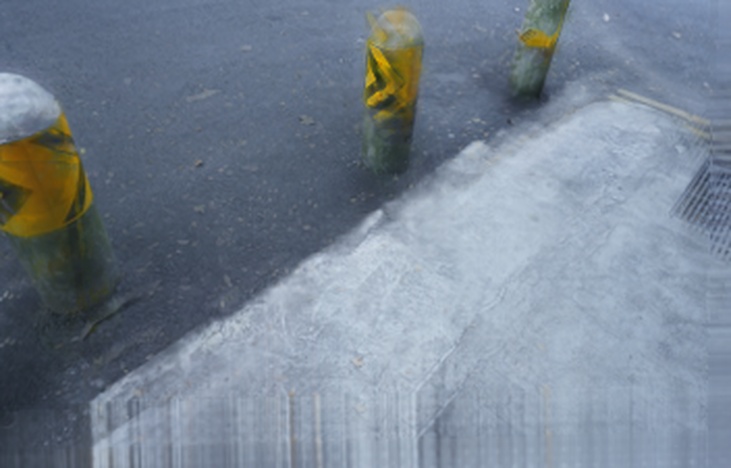} & \includegraphics[width=0.143\textwidth]{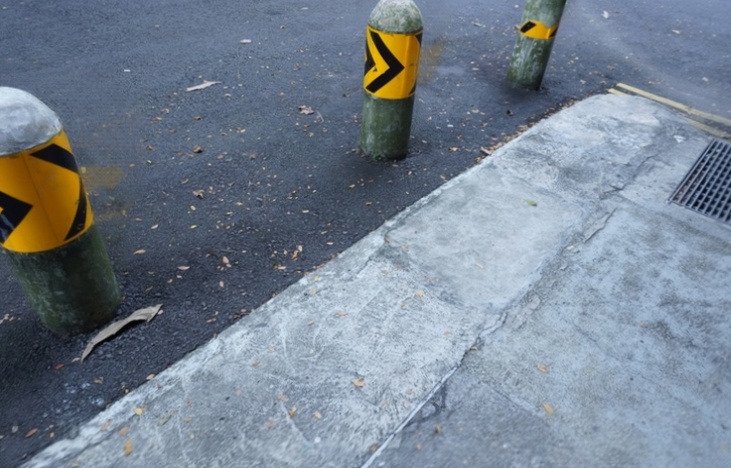} \\
\includegraphics[width=0.143\textwidth]{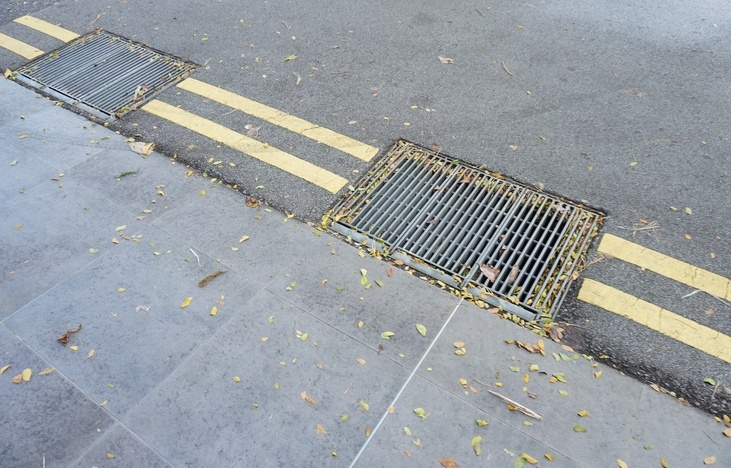} & \includegraphics[width=0.143\textwidth]{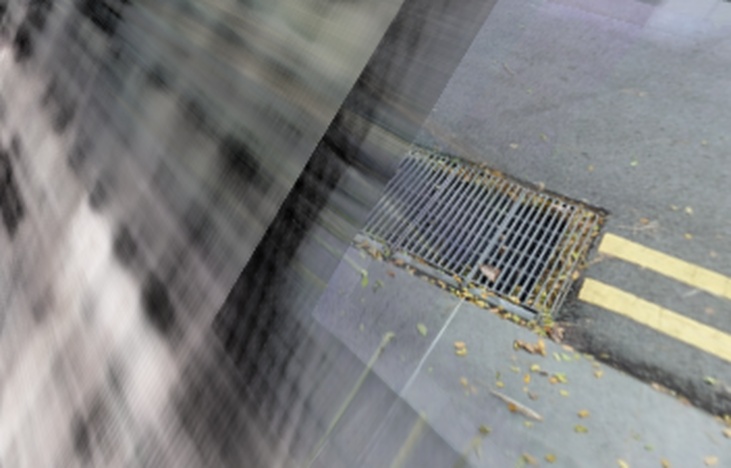} & \includegraphics[width=0.143\textwidth]{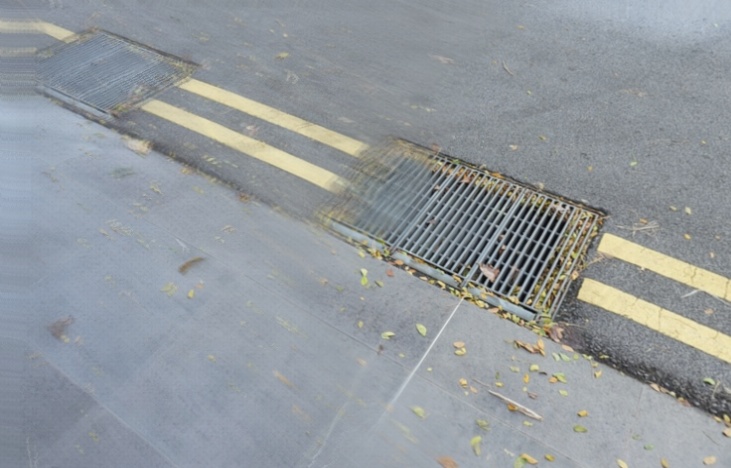} & \includegraphics[width=0.143\textwidth]{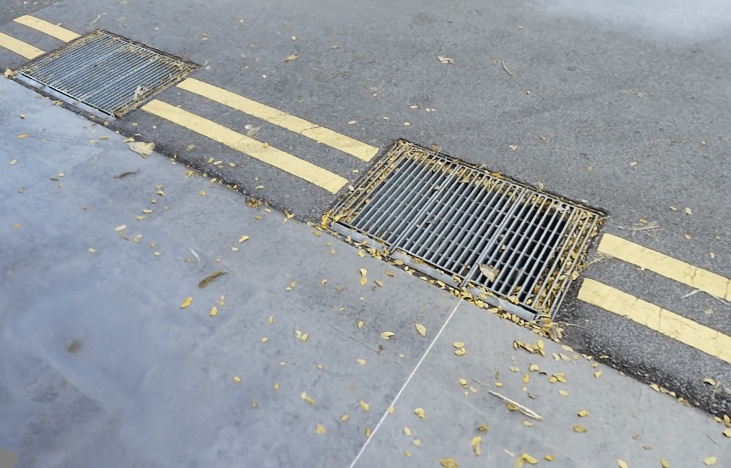} & \includegraphics[width=0.143\textwidth]{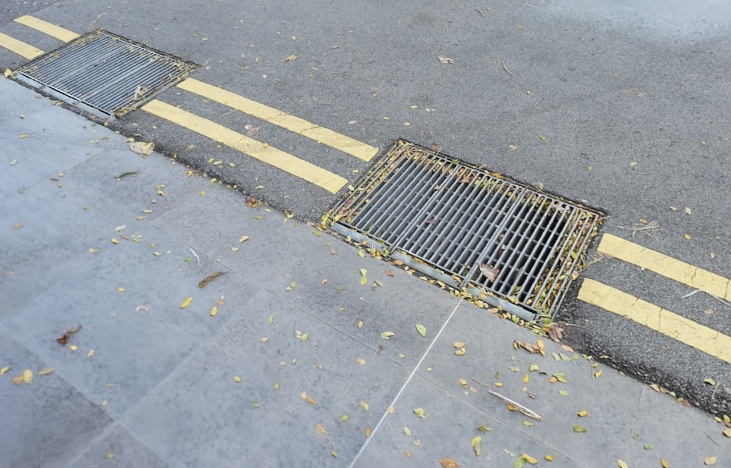} & \includegraphics[width=0.143\textwidth]{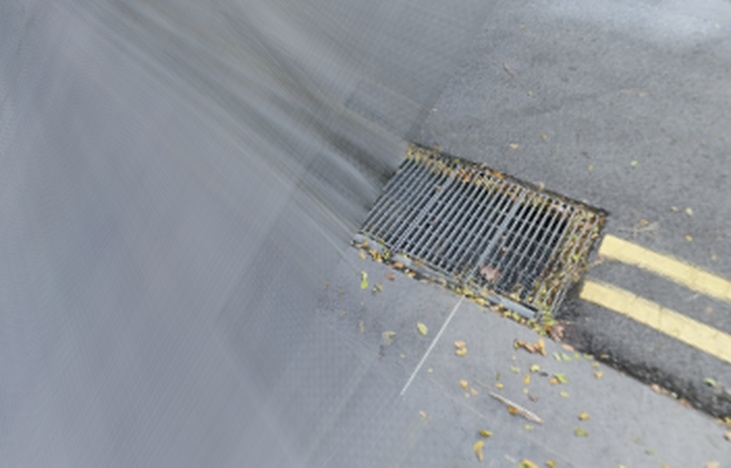} & \includegraphics[width=0.143\textwidth]{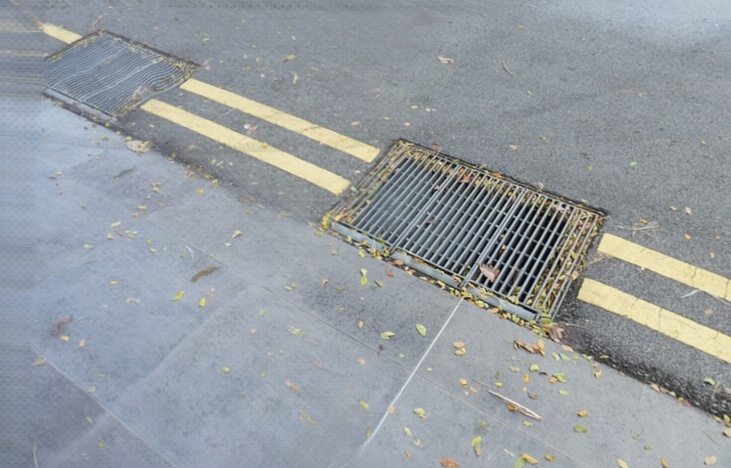} \\
Ground truth & MVSNeRF & Ours & F2-NeRF & Zip-NeRF & MVSNeRF + ft & Ours + ft \\
& \cite{chen2021mvsnerf} & & \cite{wang2023f2} & \cite{barron2023zip} & \cite{chen2021mvsnerf}\\
\end{tabular}%
}
\caption{\textbf{Qualitative comparisons of rendering quality on the Free~\cite{wang2023f2} dataset.}}
\label{fig:qualitative}
\end{figure*}

\paragraph{Free dataset.}
On the Free dataset, BoostMVSNeRFs emerges as the best among no per-scene and per-scene optimization NeRF methods as shown in Table~\ref{tab:quantitative_free} and Fig.~\ref{fig:qualitative}. Compared to F2-NeRF and SurfelNeRF, which produced blurred images, BoostMVSNeRFs leverages multiple cost volume fusion and view selection based on visibility maps for superior rendering quality. Our method demonstrates compatibility with various camera trajectories and achieves results comparable to those of existing methods.

\begin{figure*}[t]
\centering
\small
\setlength{\tabcolsep}{1pt}
\renewcommand{\arraystretch}{1}
\resizebox{1.0\textwidth}{!} 
{
\begin{tabular}{ccccc}
\includegraphics[width=0.2\textwidth]{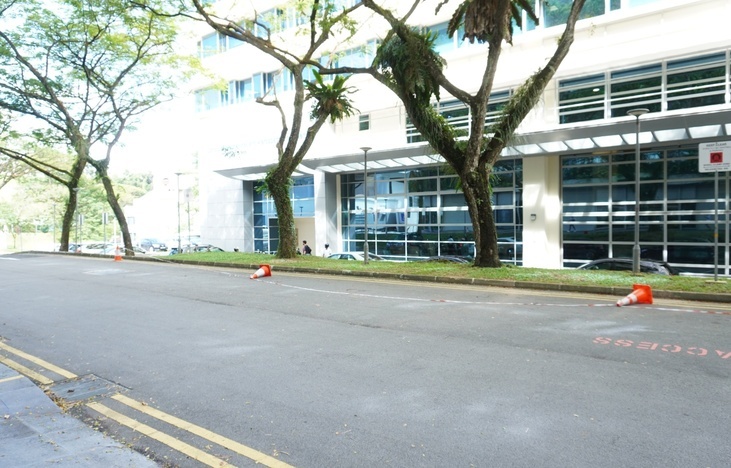} & \includegraphics[width=0.2\textwidth]{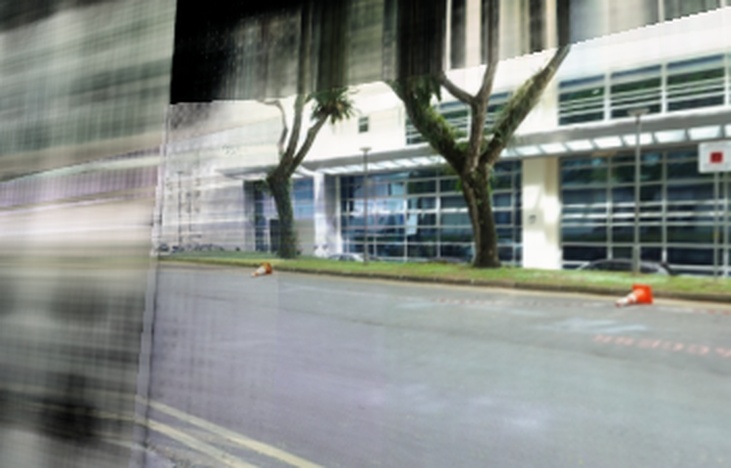} & \includegraphics[width=0.2\textwidth]{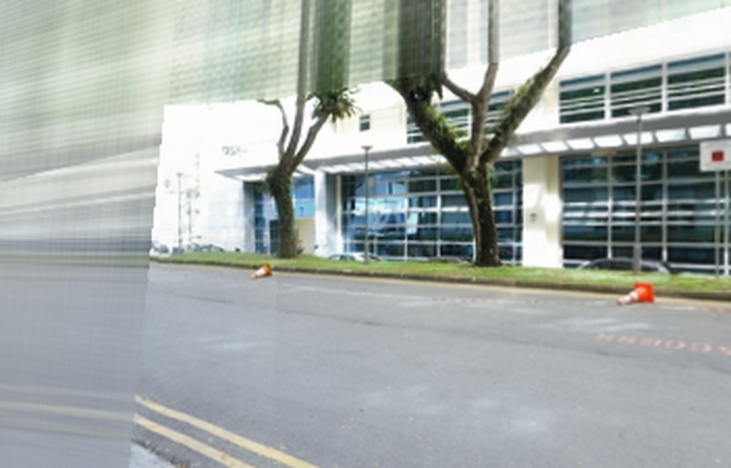} & \includegraphics[width=0.2\textwidth]{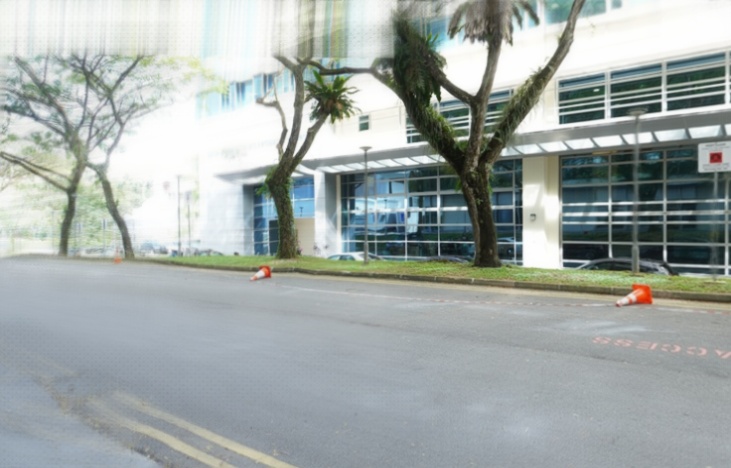} & \includegraphics[width=0.2\textwidth]{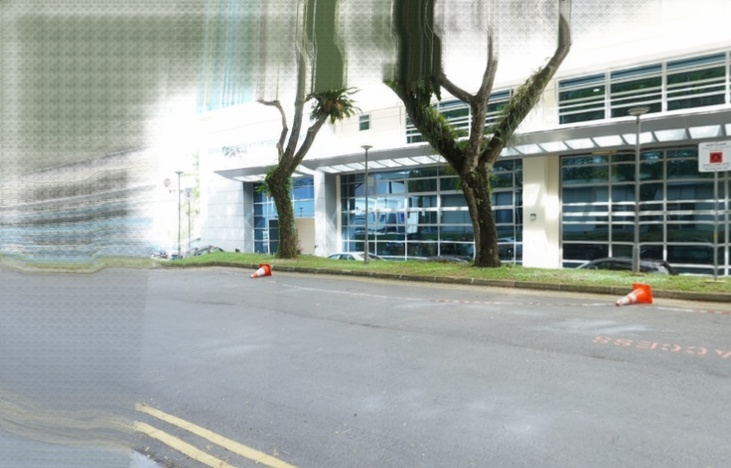} \\
Ground truth & MVSNeRF~\cite{chen2021mvsnerf} & MVSNeRF + ft & ENeRF~\cite{lin2022efficient} & ENeRF + ft \\
 & \includegraphics[width=0.2\textwidth]{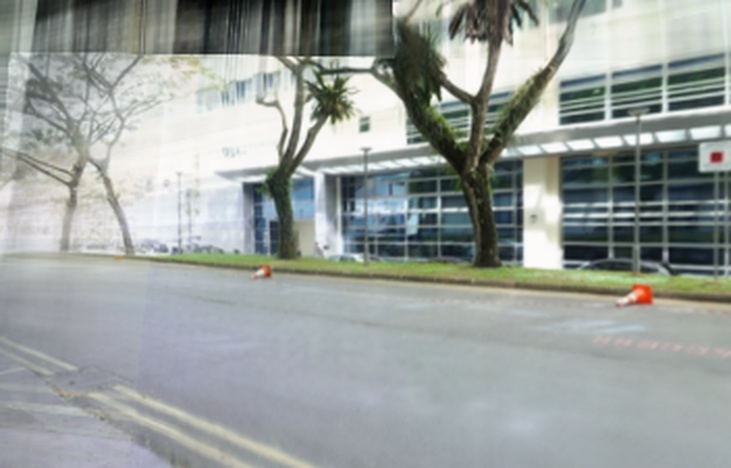} & \includegraphics[width=0.2\textwidth]{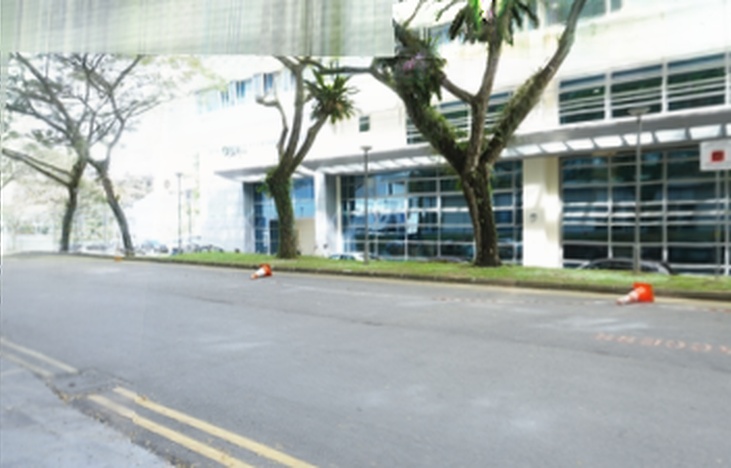} & \includegraphics[width=0.2\textwidth]{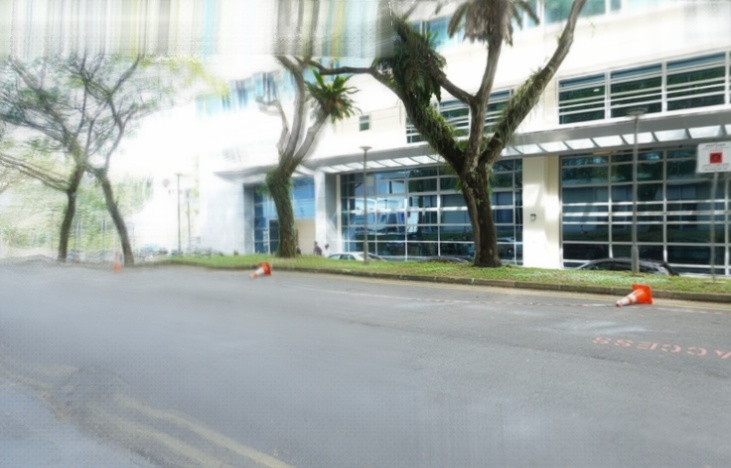} & \includegraphics[width=0.2\textwidth]{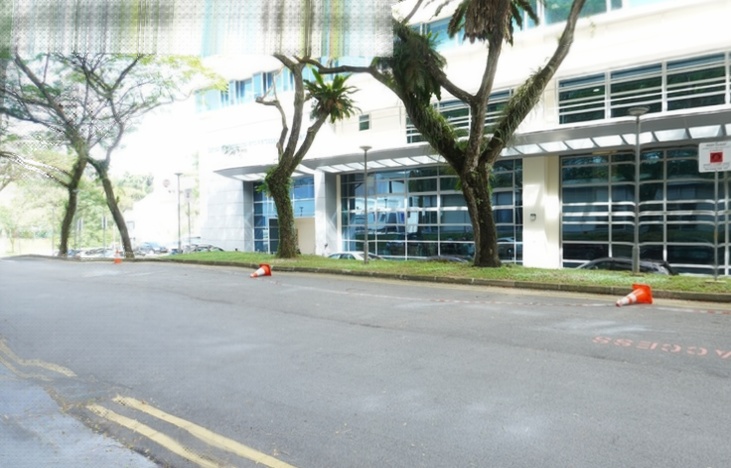}\\
 & MVSNeRF + \textbf{ours} & MVSNeRF + \textbf{ours} + ft & ENeRF + \textbf{ours} & ENeRF + \textbf{ours} + ft
\end{tabular}%
}
\caption{\textbf{Qualitative rendering quality improvements of integrating our method into MVS-based NeRF methods on the Free dataset.}}
\label{fig:qualitative_boostmvs_free}
\end{figure*}

Our method outperforms generalizable NeRF techniques like MVSNeRF and ENeRF on the Free dataset (Table~\ref{tab:quantitative_free}), enhancing rendering quality through our view selection and multiple cost volume combined rendering approach. Integrated with MVS-based NeRFs, our method achieves a PSNR improvement of 0.5-1.0 dB without requiring additional training. End-to-end fine-tuning on test scenes further enhances rendering quality, particularly in regions where a single cost volume falls short. This highlights the benefit of multiple-cost volume fusion. For detailed visual comparisons, please refer to Fig.~\ref{fig:qualitative_boostmvs_free},~\ref{fig:appendix_qualitative_free},~\ref{fig:appendix_qualitative_boostmvs_free}, and the supplementary material.

\begin{table}[t]
\centering
\small
\caption{\textbf{Quantitative comparisons with state-of-the-art methods on the ScanNet~\cite{dai2017scannet} dataset.}}
\label{tab:quantitative_scannet}
\resizebox{\columnwidth}{!}{%
\begin{tabular}{lccccc}
\toprule
Method & Setting & PSNR $\uparrow$ & SSIM $\uparrow$ & LPIPS $\downarrow$ & FPS $\uparrow$ \\
\midrule
SurfelNeRF~\cite{gao2023surfelnerf} & \multirow{5}{*}{\makecell{No per-scene\\optimization}} & 19.28 & 0.623 & 0.528 & 1.25 \\
MVSNeRF~\cite{chen2021mvsnerf} &  & 23.40 & 0.862 & 0.367 & 1.99 \\
MVSNeRF + Ours & & 23.66 & 0.872 & 0.365 & 1.41 \\
ENeRF~\cite{lin2022efficient} &  & \textbf{31.73} & 0.955 & \textbf{0.206} & \textbf{11.03} \\
ENeRF + Ours &  & 31.01 & \textbf{0.957} & 0.219 & 6.14 \\
\midrule
F2-NeRF~\cite{wang2023f2} & \multirow{7}{*}{\makecell{Per-scene\\optimization}} & 28.11 & 0.894 & 0.230 & 4.18 \\
$\text{SurfelNeRF}_\text{ft}$~\cite{gao2023surfelnerf} & & 20.04 & 0.653 & 0.504 & 1.25 \\
Zip-NeRF~\cite{barron2023zip} & & 32.24 & 0.917 & 0.214 & 0.74 \\
$\text{MVSNeRF}_\text{ft}$~\cite{chen2021mvsnerf} & & 24.69 & 0.872 & 0.316 & 1.99 \\
MVSNeRF + $\text{Ours}_\text{ft}$ &  & 24.63 & 0.880 & 0.320 & 1.41 \\
$\text{ENeRF}_\text{ft}$~\cite{lin2022efficient} &  & 32.70 & \textbf{0.960} & 0.174 & \textbf{11.03} \\ 
$\text{ENeRF + Ours}_\text{ft}$ & & \textbf{32.87} & 0.955 & \textbf{0.173} & 6.14 \\
\bottomrule
\end{tabular}%
}
\end{table}

\paragraph{ScanNet dataset.}
We conducted a comprehensive comparison of BoostMVSNeRFs with other state-of-the-art methods in no per-scene and per-scene optimization settings on the ScanNet dataset in Table~\ref{tab:quantitative_scannet}. 
BoostMVSNeRFs demonstrates superior performance with a PSNR of 31.73 dB in no per-scene optimization, outperforming SurfelNeRF due to its cost volume fusion and efficient view selection strategy. In per-scene optimization, BoostMVSNeRFs excels again with a PSNR of 32.87 dB, indicating its effectiveness in cost volume fusion and per-scene adaptation. We also compare our method with two generalizable NeRF methods, MVSNeRF and ENeRF, on the ScanNet dataset in Table~\ref{tab:quantitative_scannet} and Fig.~\ref{fig:qualitative_boostmvs_scannet}. 
Our method achieves better rendering quality over existing MVS-based NeRF methods in SSIM without per-scene optimization and in PSNR and LPIPS with per-scene fine-tuning.

Furthermore, our approach showed impressive results on large-scale scenes, outperforming SurfelNeRF in both direct inference and per-scene fine-tuning. 
Unlike SurfelNeRF, which suffered from artifacts due to its surfel-based rendering approach, our model's multiple cost volume fusion and efficient view information selection and aggregation led to high-quality and consistent renderings, as shown in Fig.~\ref{fig:qualitative}. This indicates our cost volume fusion's effectiveness in reconstructing large-scale scenes efficiently and accurately.

\subsection{Ablation Study}
\paragraph{Cost volume selection scheme.} In Sec. \ref{sec:CV_select}, we propose a greedy method to select the cost volumes that will approximately maximize the view coverage. To validate the effectiveness of our method, We conducted experiments comparing two other cost volume selection methods. These two methods are: (a) selecting  $K$ cost volumes that are closest to the render view pose, which is adopted by ENeRF~\cite{lin2022efficient} and (b) selecting corresponding cost volumes directly with the highest contribution of 2D visibility mask. 
In particular, method (b) is a degenerate version of method our proposed selection method (c), which is based on view coverage.
Table~\ref{tab:ablation_cv_select} shows that our greedy cost volume selection method performs better than the other two methods.

\begin{table}[t]
\centering
\small
\caption{
\textbf{Ablation of the cost volume selection.} We compare three different strategies for cost volume selection on all scenes of the Free~\cite{wang2023f2} dataset: (a) ENeRF's method, which is based on pose distance, (b) direct selection of cost volumes with maximum visibility, and (c) Our proposed greedy method, which maximizes the visibility coverage.
}
\label{tab:ablation_cv_select}
\begin{tabular}{lccc}
\toprule
Method & PSNR $\uparrow$ & SSIM $\uparrow$ & LPIPS $\downarrow$ \\
\midrule
(a) ENeRF~\cite{lin2022efficient} & 24.09 & 0.861 & 0.220 \\
(b) Maximize 2D visibility $\textbf{M}^\text{2D}_i$ & 24.19 & 0.861 & 0.218 \\
(c) Maximize view coverage $\textbf{P}_i$ & \textbf{24.21} & \textbf{0.862} & \textbf{0.218} \\
\bottomrule
\end{tabular}
\end{table}



\begin{figure}[t]
\centering
\footnotesize
\setlength{\tabcolsep}{1pt}
\renewcommand{\arraystretch}{1}
\resizebox{1.0\columnwidth}{!} 
{
\begin{tabular}{cccc}
\includegraphics[height=0.25\columnwidth]{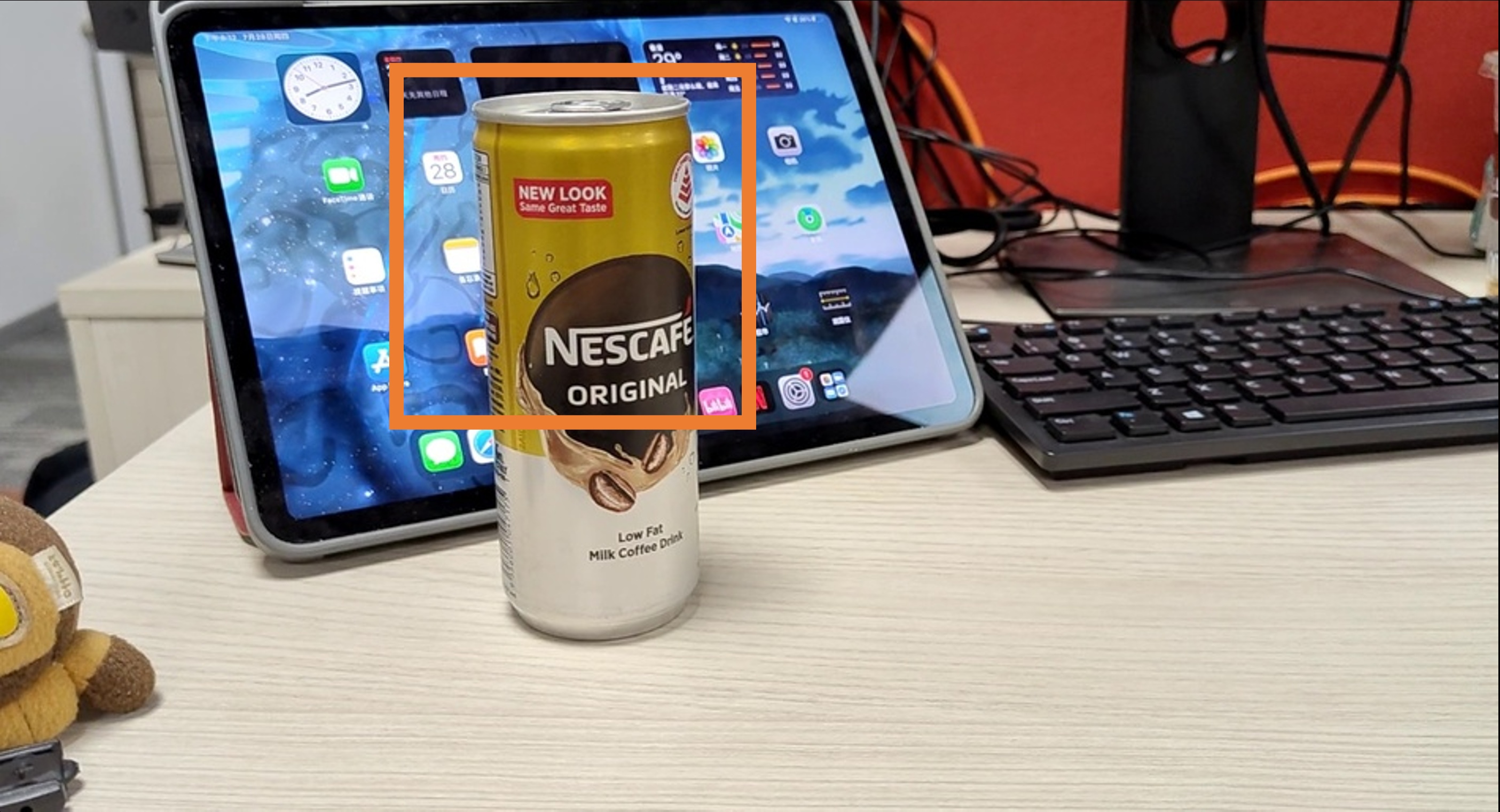} & 
\includegraphics[height=0.25\columnwidth]{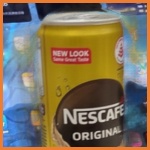} & 
\includegraphics[height=0.25\columnwidth]{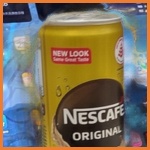} & 
\includegraphics[height=0.25\columnwidth]{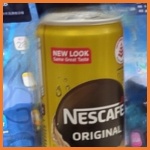} \\
& $\text{ENeRF}^\text{3-view}$ & $\text{ENeRF}^\text{6-view}$ & $\text{ENeRF}^\text{3-view}$ + $\text{Ours}$ \\
\includegraphics[height=0.25\columnwidth]{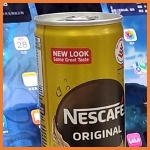} & \includegraphics[height=0.25\columnwidth]{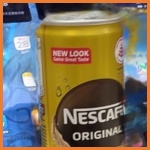} & 
\includegraphics[height=0.25\columnwidth]{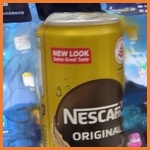} & 
\includegraphics[height=0.25\columnwidth]{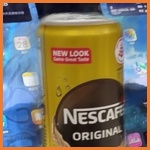} \\
Ground truth & $\text{ENeRF}^\text{3-view}_\text{ft}$ & $\text{ENeRF}^\text{6-view}_\text{ft}$ & $\text{ENeRF}^\text{3-view}$ + $\text{Ours}_\text{ft}$ \\
\end{tabular}%
}
\caption{\textbf{Visual effects of different ways of combining more input views.} Artifacts in disocclusion regions cannot be resolved by including more input views for a single cost volume. Our method could alleviate these artifacts by combining more cost volumes in rendering.}
\label{fig:visual_6views}
\end{figure}

\paragraph{Single cost volume with more input views vs. Combining multiple cost volumes} 
In our method, we select multiple cost volumes and combine them in volume rendering, while ENeRF only forms one cost volume. To examine our method's effectiveness, we train ENeRF (originally three input views) with more input views (6 in this ablation, in order to evenly compare with our proposed method). The results are shown in Table~\ref{tab:ablation_6_views} and Fig.~\ref{fig:visual_6views}. We can see an increase in the number of input views which requires time-consuming training to construct a single cost volume. However, the rendering quality improvements are subtle both with or without per-scene fine-tuning. In contrast, our cost volume selection method and combined rendering scheme improve the rendering quality by a large margin and could be further optimized with per-scene fine-tuning.

\begin{table}[t]
\centering
\caption{
\textbf{Different ways of combining more input views.} We compare training an MVS-based NeRF with a larger number of input views (6 input views here) and our proposed cost volume selection and combined rendering on all scenes of the Free~\cite{wang2023f2} dataset.
}
\label{tab:ablation_6_views}
\resizebox{\columnwidth}{!}{%
\begin{tabular}{lcccc}
\toprule
Method & Setting & PSNR $\uparrow$ & SSIM $\uparrow$ & LPIPS $\downarrow$ \\
\midrule
$\text{ENeRF}^\text{3-view}$~\cite{lin2022efficient} & \multirow{3}{*}{\makecell{No per-scene\\optimization}} & 23.24 & 0.844 & 0.225 \\
$\text{ENeRF}^\text{6-view}$~\cite{lin2022efficient} & & 23.53 & 0.770 & 0.231 \\
$\text{ENeRF}^\text{3-view}$ + $\text{Ours}$ &  & \textbf{24.21} & \textbf{0.862} & \textbf{0.218} \\
\midrule
$\text{ENeRF}^\text{3-view}_\text{ft}$~\cite{lin2022efficient} & \multirow{3}{*}{\makecell{Per-scene\\optimization}} & 25.19 & 0.880 & 0.180 \\
$\text{ENeRF}^\text{6-view}_\text{ft}$~\cite{lin2022efficient} & & 25.61 & 0.840 & 0.172 \\
$\text{ENeRF}^\text{3-view}$ + $\text{Ours}_\text{ft}$ &  & \textbf{26.14} & \textbf{0.894} & \textbf{0.171} \\
\bottomrule
\end{tabular}%
}
\end{table}

\begin{figure}[t]
\centering
\setlength{\tabcolsep}{1pt}
\renewcommand{\arraystretch}{1}
\resizebox{1.0\columnwidth}{!} 
{
\begin{tabular}{cc}
\includegraphics[width=0.5\columnwidth]{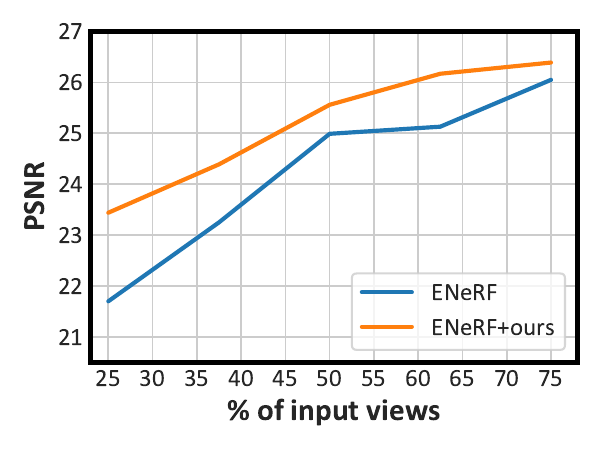} & \includegraphics[width=0.5\columnwidth]{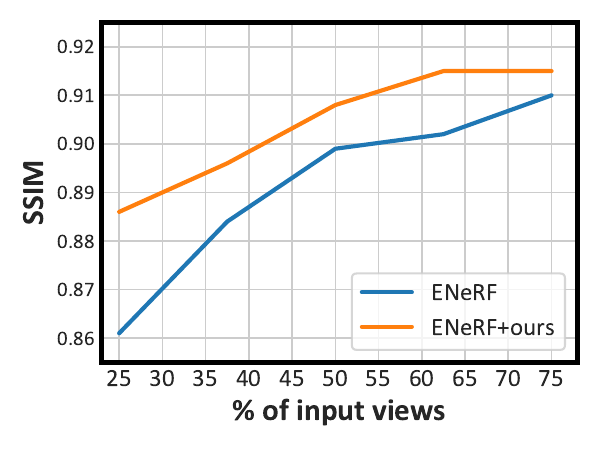}
\end{tabular}%
}
\caption{\textbf{Robustness with sparse input views.} 
With more sparse input views, the performance drop of our method is less severe than ENeRF, demonstrating the robustness of our method against sparse input views by combining multiple cost volumes in rendering.
}
\label{fig:sparse_input_views}
\end{figure}

\paragraph{Robustness with sparse input views.}
Our proposed combined rendering from multiple cost volumes addresses the challenges of reconstructing large-scale and unbounded scenes due to broader viewport coverage. Therefore, our method could be more robust to sparse input views as more and farther cost volumes are considered during rendering.
We conduct an experiment comparing performance across various degrees of sparse views to demonstrate the robustness of our method with sparse input views. Specifically, we uniformly sub-sample the training views and evaluate the rendering quality. The results show a more significant decline in both PSNR and SSIM for ENeRF compared to ours while input views become sparse, as indicated by the curve in figure~\ref{fig:sparse_input_views}.

\section{Conclusion}
%
In summary, our BoostMVSNeRFs enhances MVS-based NeRFs, tackling large-scale and unbounded scene rendering challenges. Utilizing 3D visibility scores for multi-cost volume integration, BoostMVSNeRFs synthesizes significantly better novel views, enhancing viewport coverage and minimizing typical single-cost volume artifacts. Compatible with current MVS-based NeRFs, BoostMVSNeRFs supports end-to-end training for scene-specific enhancement. Experimental results validate the efficacy of our method in boosting advanced MVS-based NeRFs, contributing to more scalable and high-quality view synthesis. Future work will focus on reducing MVS dependency and optimizing memory usage, furthering the field of neural rendering for virtual and augmented reality applications.

\begin{acks}
This research was funded by the National Science and Technology Council, Taiwan, under Grants NSTC 112-2222-E-A49-004-MY2 and 113-2628-E-A49-023-.
The authors are grateful to Google and NVIDIA for generous donations.
Yu-Lun Liu acknowledges the Yushan Young Fellow Program by the MOE in Taiwan.
The authors thank the anonymous reviewers for their valuable feedback.
\end{acks}

\bibliographystyle{ACM-Reference-Format}
\bibliography{sample-base}

\appendix

\begin{figure*}[t]
\centering
\small
\setlength{\tabcolsep}{1pt}
\renewcommand{\arraystretch}{1}
\resizebox{1.0\textwidth}{!} 
{
\begin{tabular}{c|cc|cccc}
& \multicolumn{2}{c|}{No per-scene optimization} & \multicolumn{4}{c}{Per-scene optimization} \\
\includegraphics[width=0.143\textwidth]{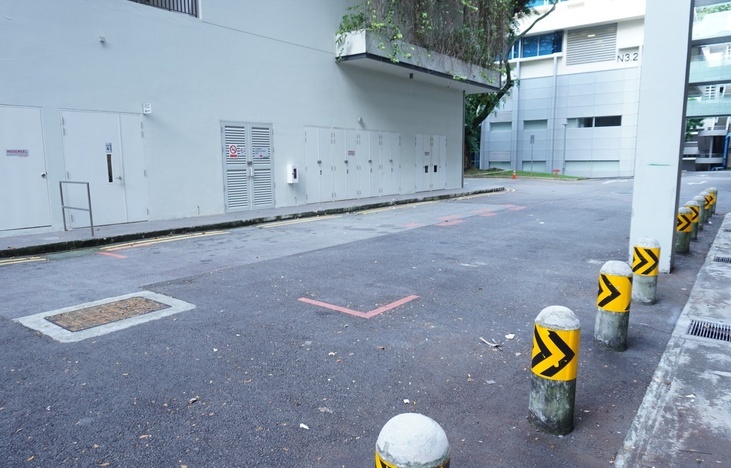} & \includegraphics[width=0.143\textwidth]{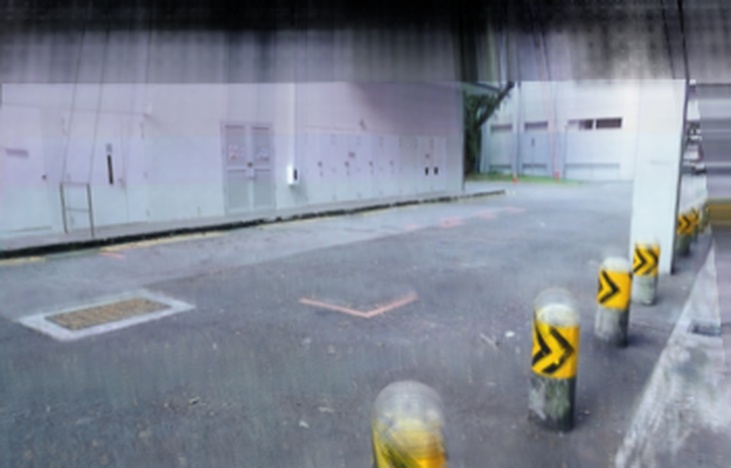} & \includegraphics[width=0.143\textwidth]{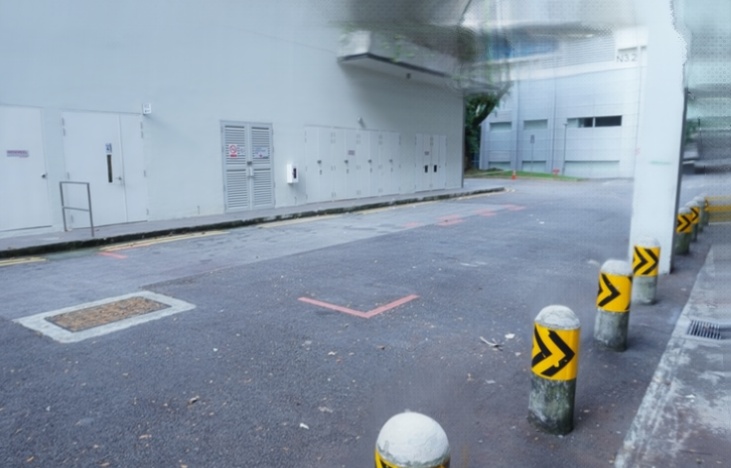} & \includegraphics[width=0.143\textwidth]{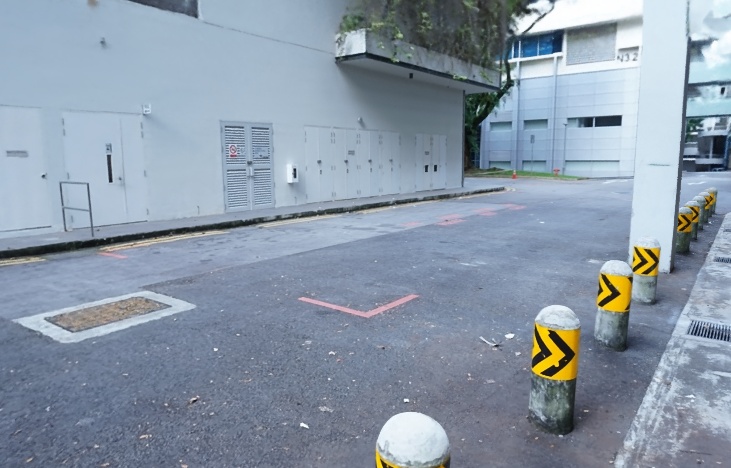} & \includegraphics[width=0.143\textwidth]{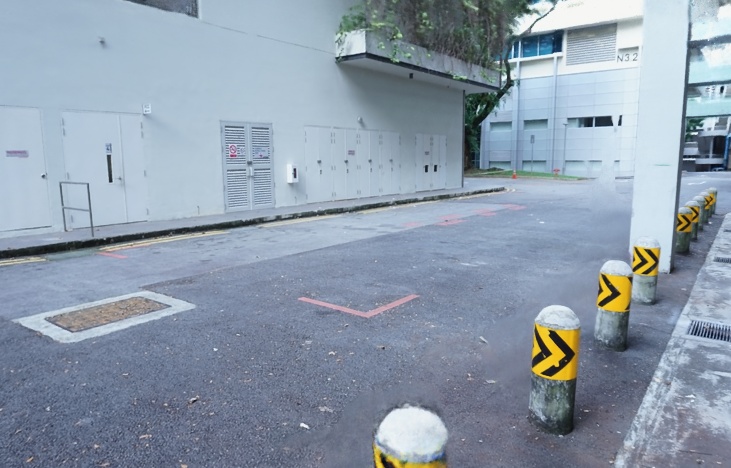} & \includegraphics[width=0.143\textwidth]{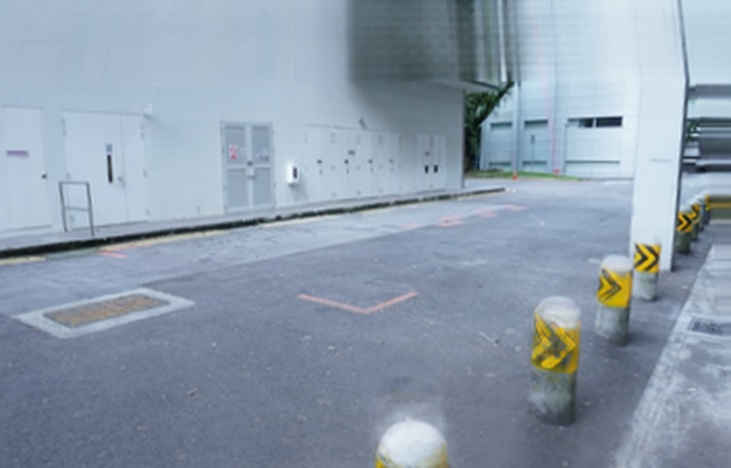} & \includegraphics[width=0.143\textwidth]{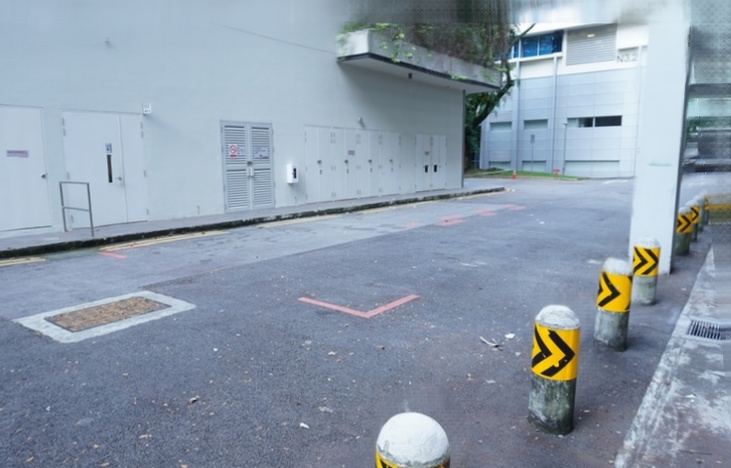} \\
\includegraphics[width=0.143\textwidth]{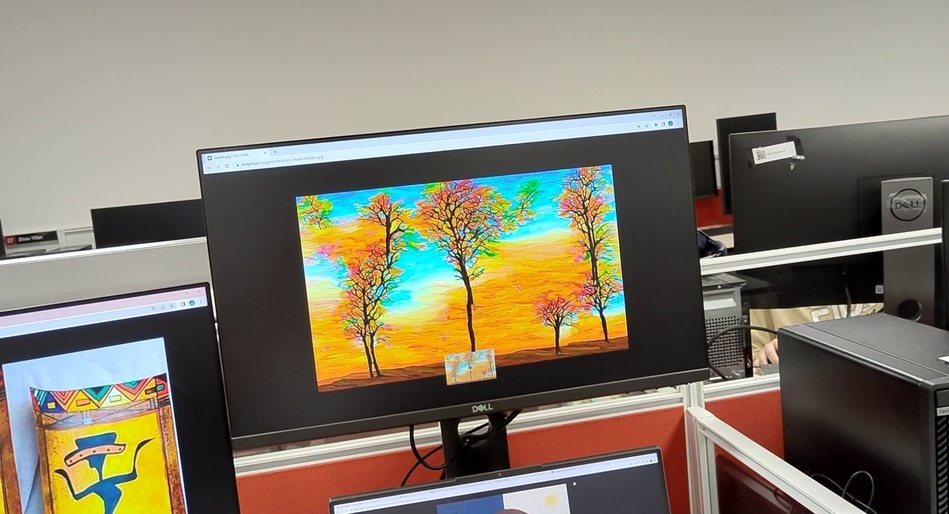} & \includegraphics[width=0.143\textwidth]{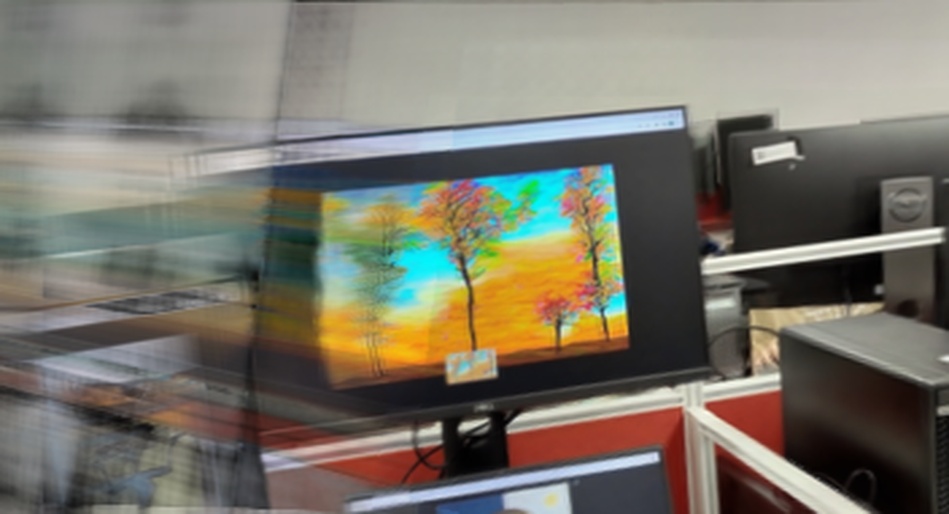} & \includegraphics[width=0.143\textwidth]{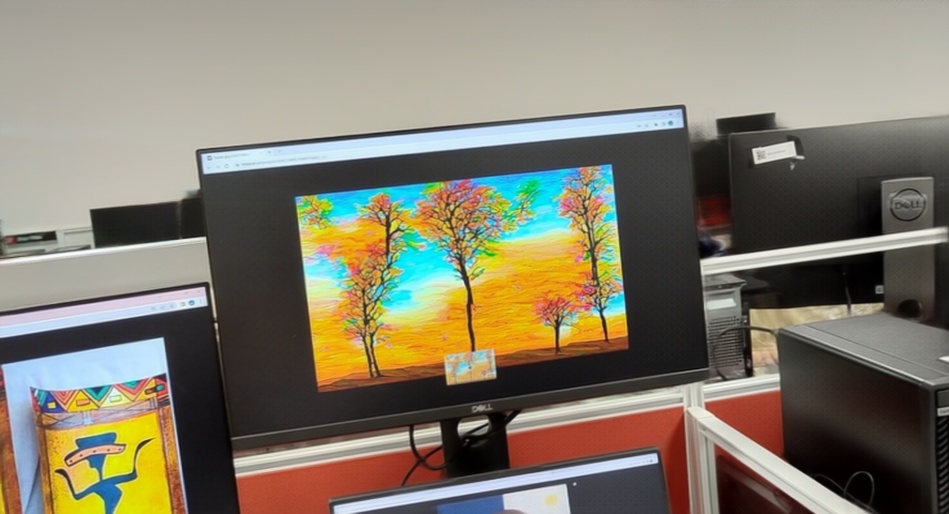} & \includegraphics[width=0.143\textwidth]{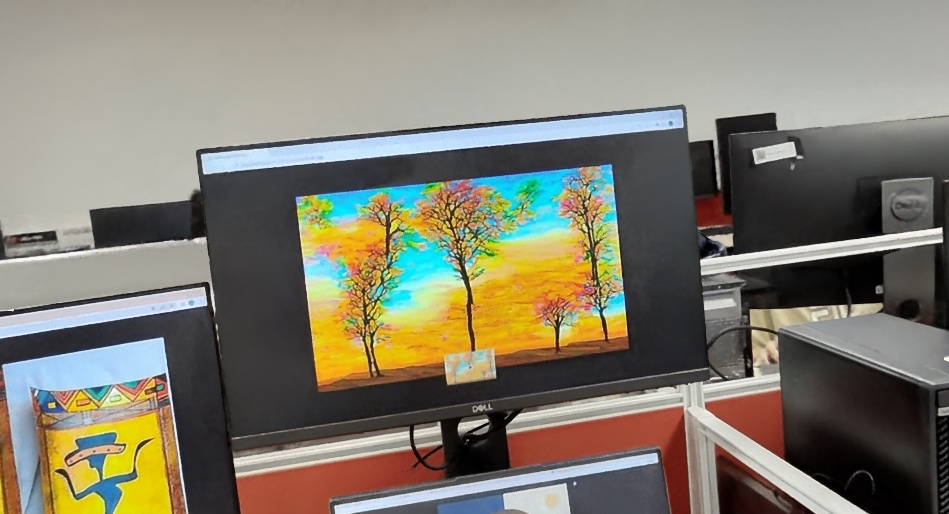} & \includegraphics[width=0.143\textwidth]{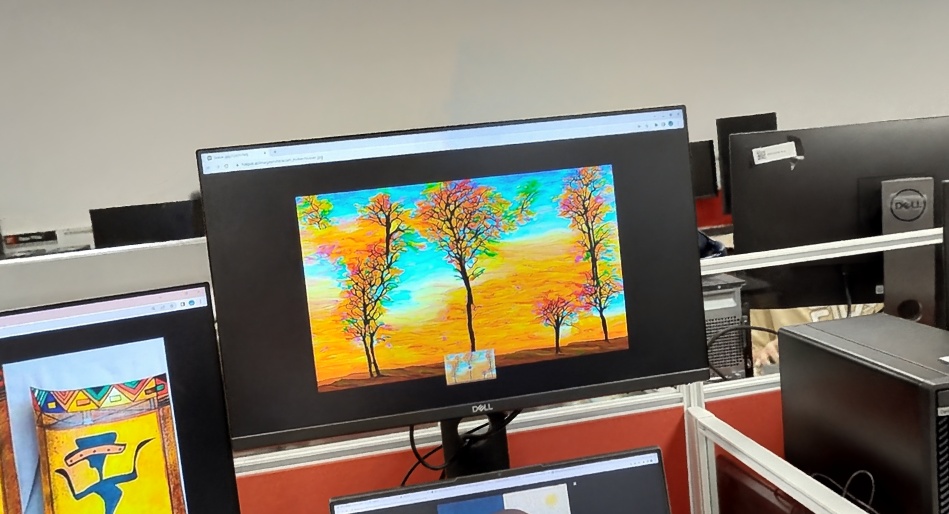} & \includegraphics[width=0.143\textwidth]{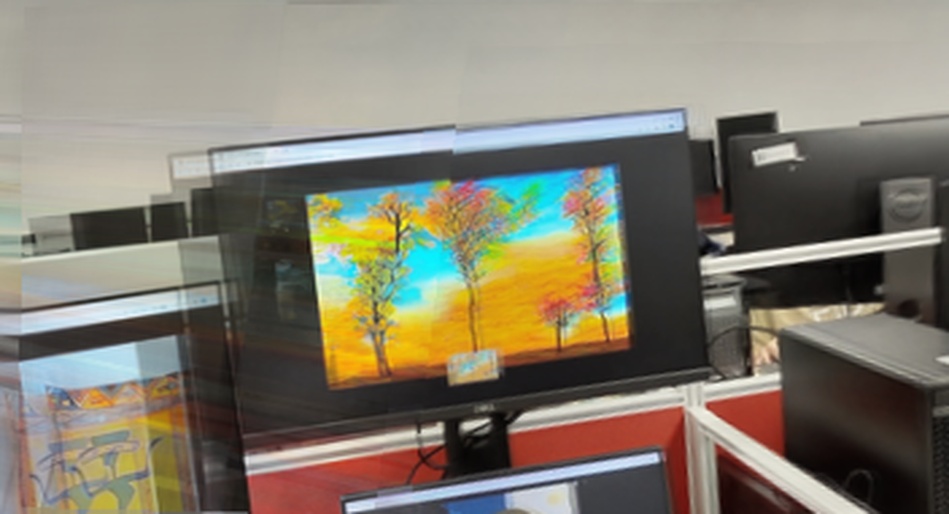} & \includegraphics[width=0.143\textwidth]{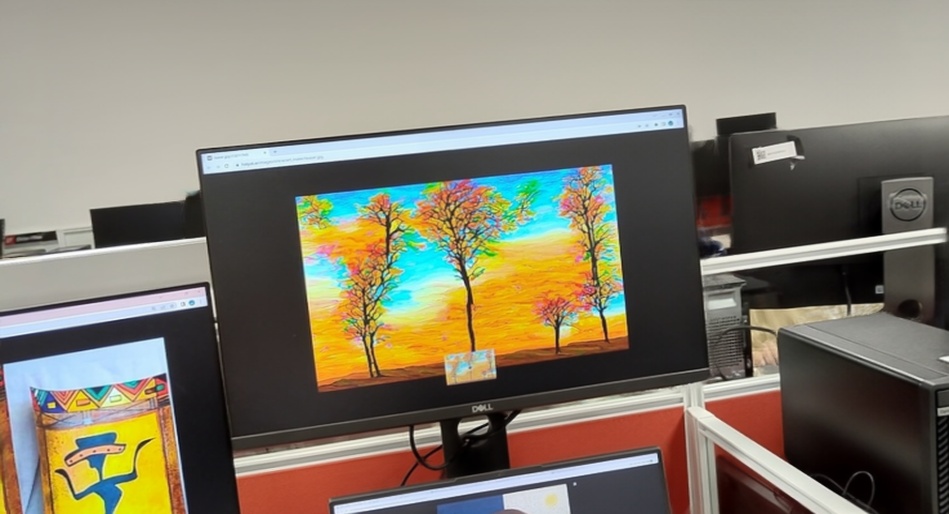} \\
\includegraphics[width=0.143\textwidth]{figures/results/road_176_0_gt.jpg} & \includegraphics[width=0.143\textwidth]{figures/results/road_176_0_mvsnerf.jpg} & \includegraphics[width=0.143\textwidth]{figures/results/road_176_0_enerf_ours.jpg} & \includegraphics[width=0.143\textwidth]{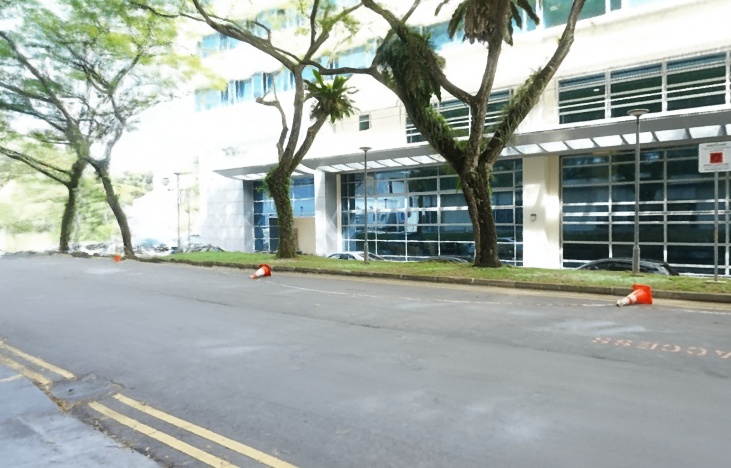} & \includegraphics[width=0.143\textwidth]{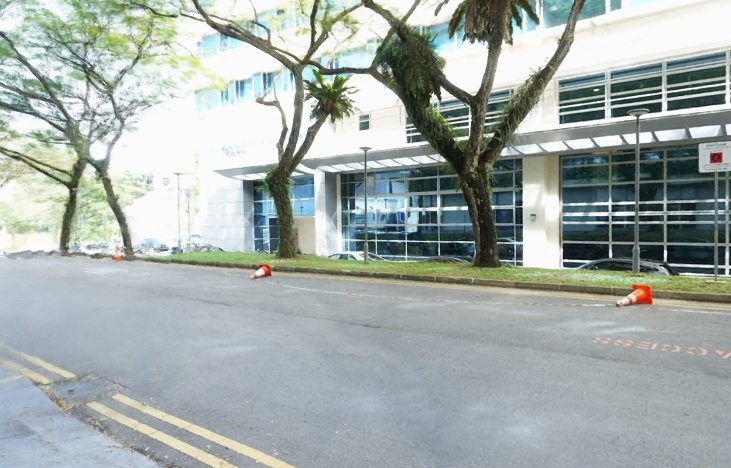} & \includegraphics[width=0.143\textwidth]{figures/results/road_176_0_mvsnerf_ft.jpg} & \includegraphics[width=0.143\textwidth]{figures/results/road_176_0_enerf_ours_ft.jpg} \\
\includegraphics[width=0.143\textwidth]{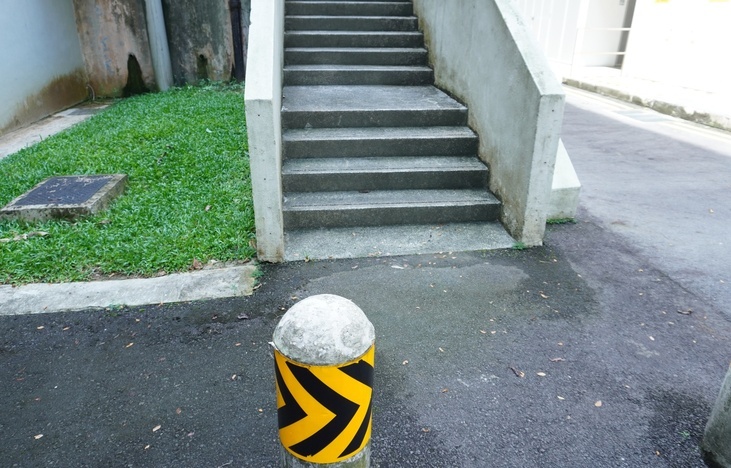} & \includegraphics[width=0.143\textwidth]{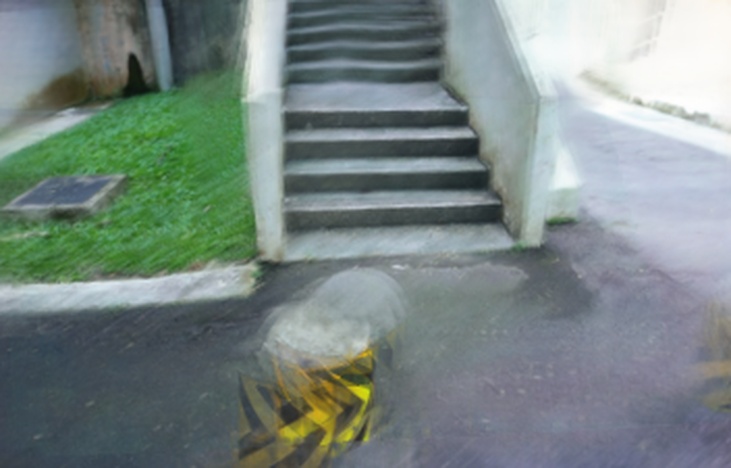} & \includegraphics[width=0.143\textwidth]{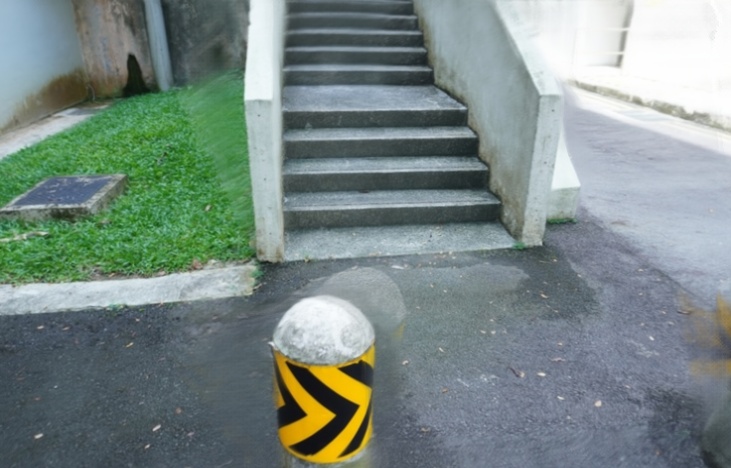} & \includegraphics[width=0.143\textwidth]{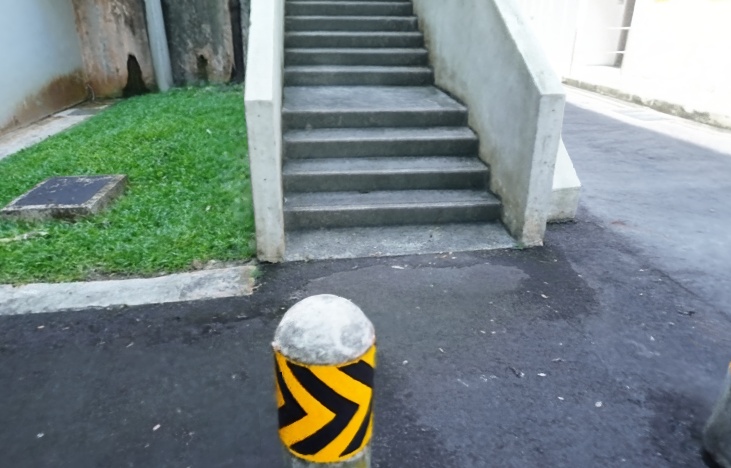} & \includegraphics[width=0.143\textwidth]{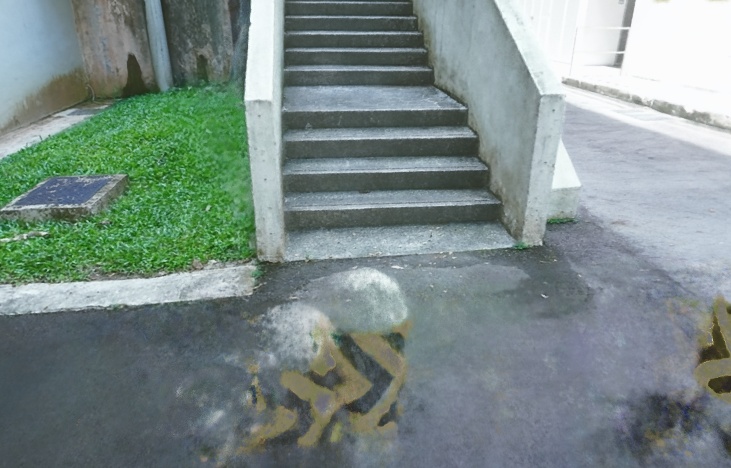} & \includegraphics[width=0.143\textwidth]{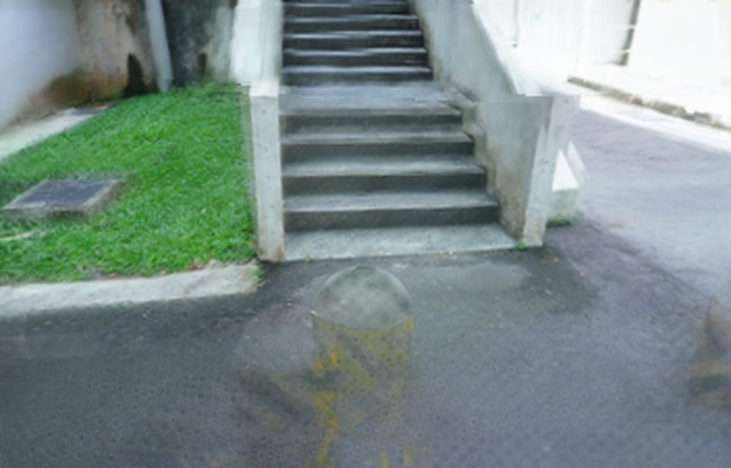} & \includegraphics[width=0.143\textwidth]{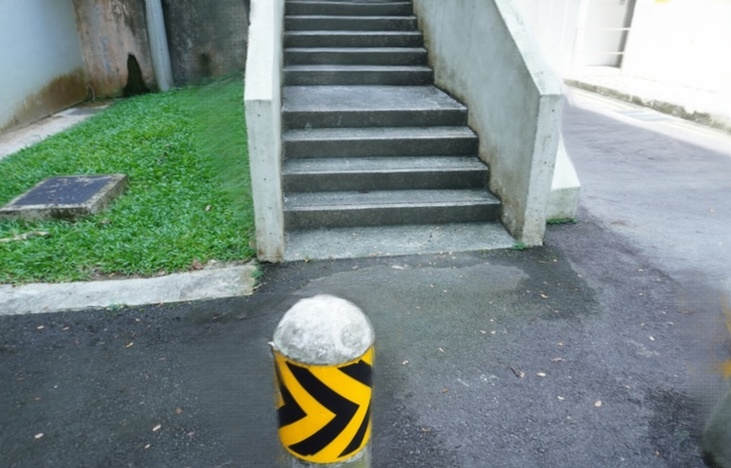} \\
\includegraphics[width=0.143\textwidth]{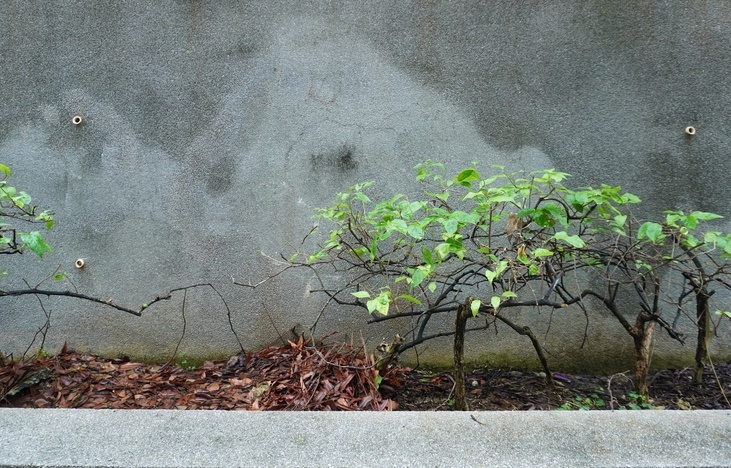} & \includegraphics[width=0.143\textwidth]{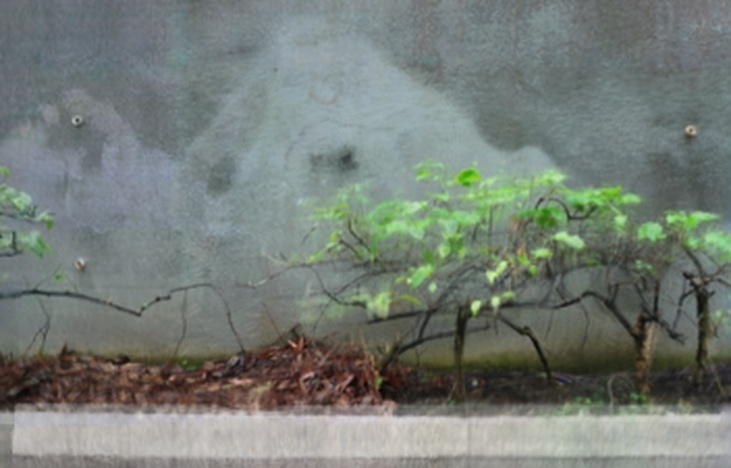} & \includegraphics[width=0.143\textwidth]{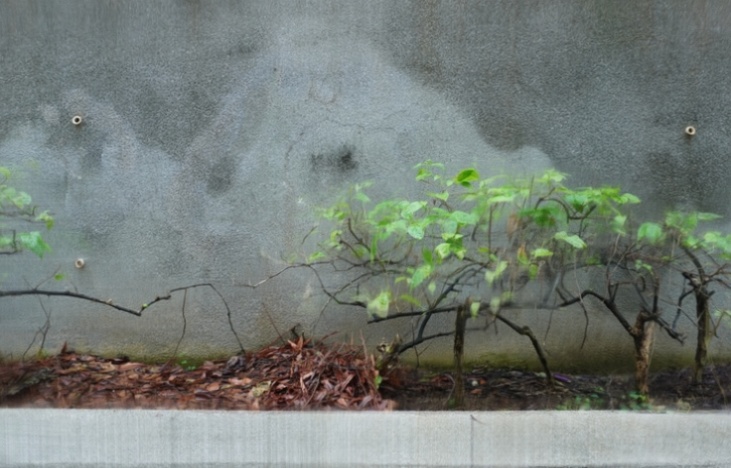} & \includegraphics[width=0.143\textwidth]{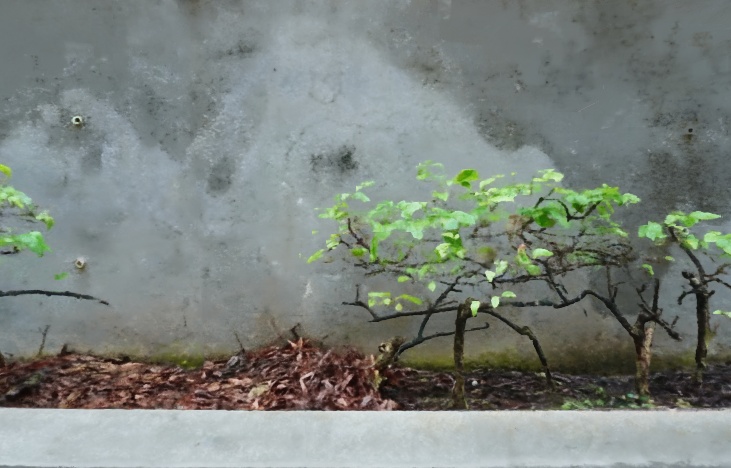} & \includegraphics[width=0.143\textwidth]{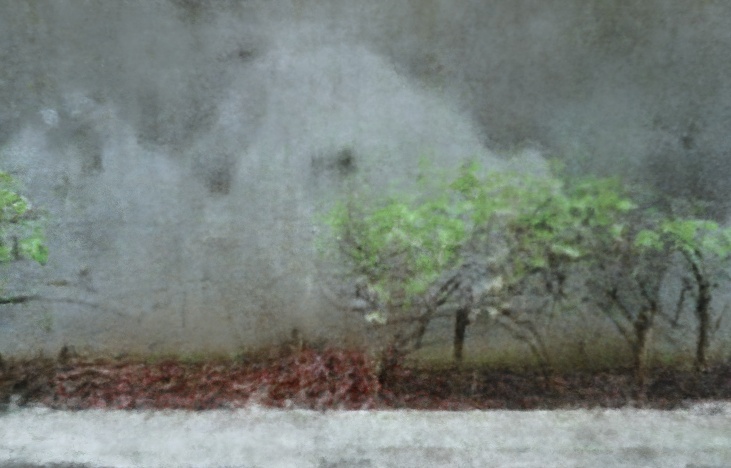} & \includegraphics[width=0.143\textwidth]{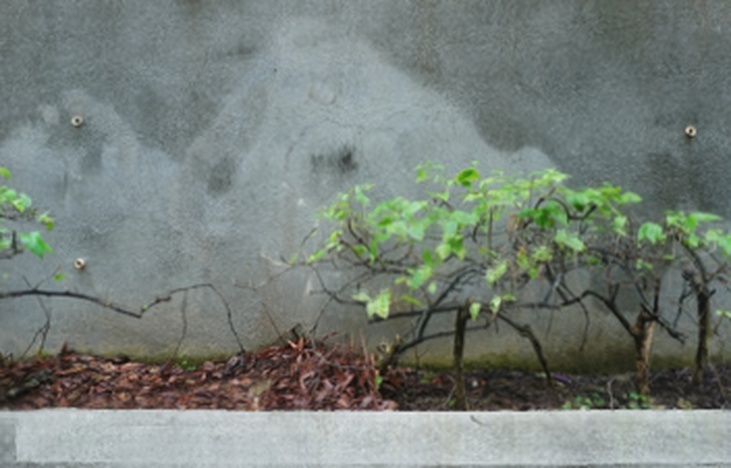} & \includegraphics[width=0.143\textwidth]{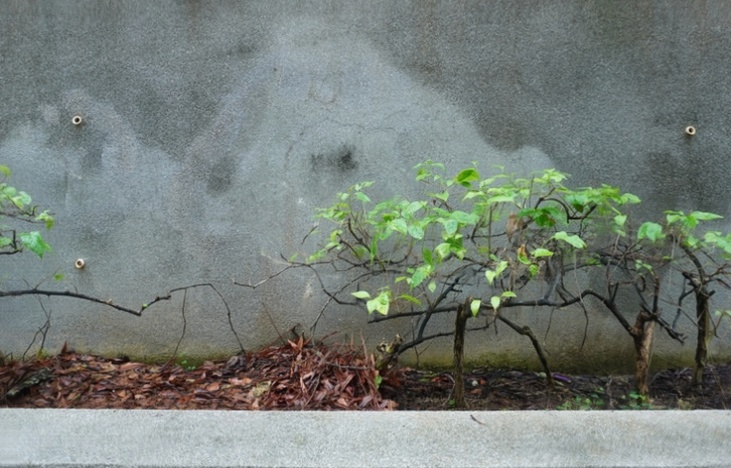} \\
Ground truth & MVSNeRF & Ours & F2-NeRF & Zip-NeRF & MVSNeRF + ft & Ours + ft \\
& \cite{chen2021mvsnerf} & & \cite{wang2023f2} & \cite{barron2023zip} & \cite{chen2021mvsnerf}\\
\end{tabular}%
}
\vspace{-2mm}
\caption{\textbf{Qualitative comparisons of rendering quality on the Free~\cite{wang2023f2} dataset.}}
\label{fig:appendix_qualitative_free}
\end{figure*}

\begin{figure*}[t]
\centering
\small
\setlength{\tabcolsep}{1pt}
\renewcommand{\arraystretch}{1}
\resizebox{1.0\textwidth}{!} 
{
\begin{tabular}{ccccc}
\includegraphics[width=0.2\textwidth]{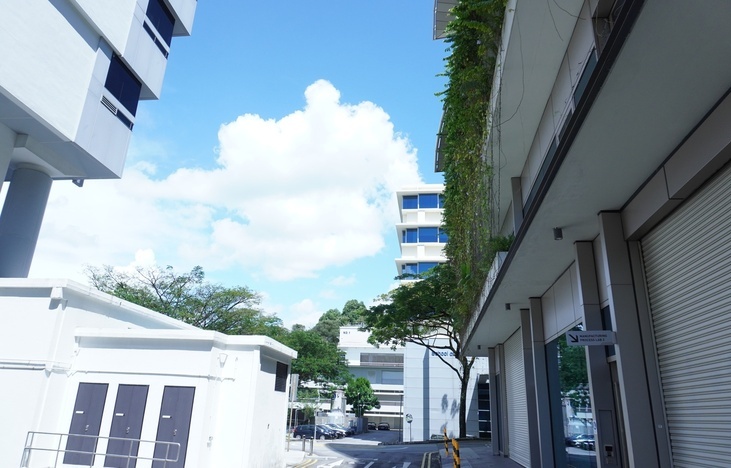} & \includegraphics[width=0.2\textwidth]{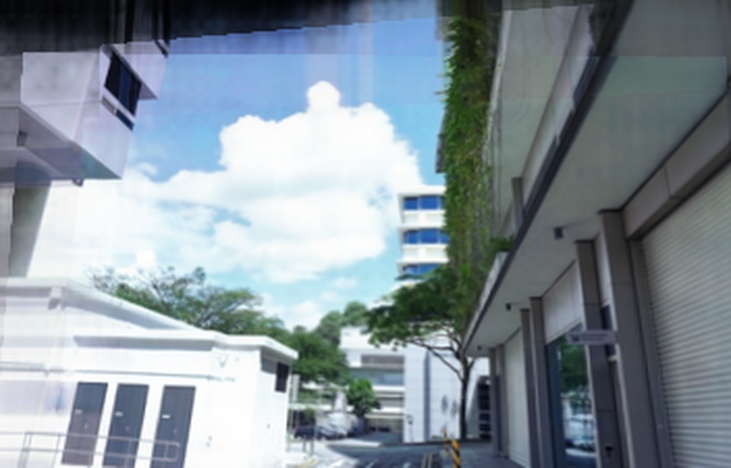} & \includegraphics[width=0.2\textwidth]{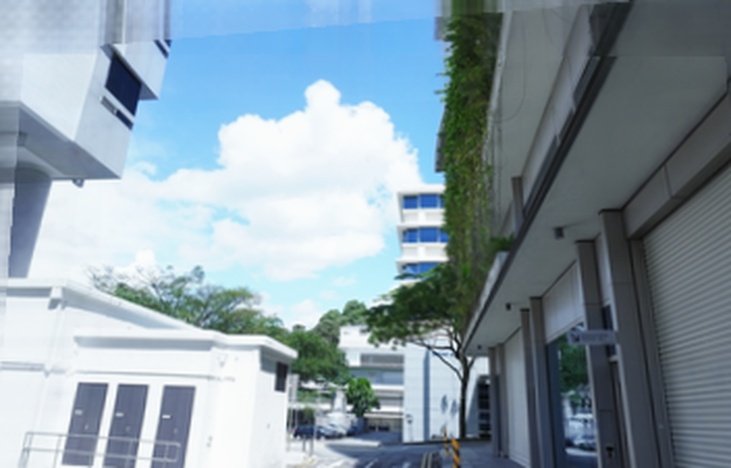} & \includegraphics[width=0.2\textwidth]{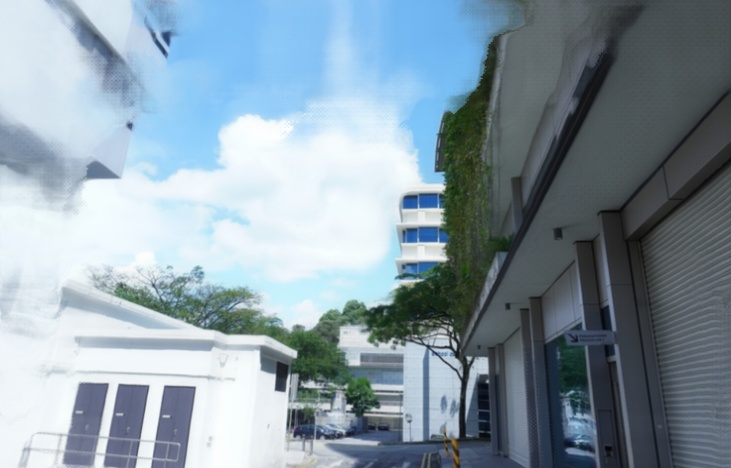} & \includegraphics[width=0.2\textwidth]{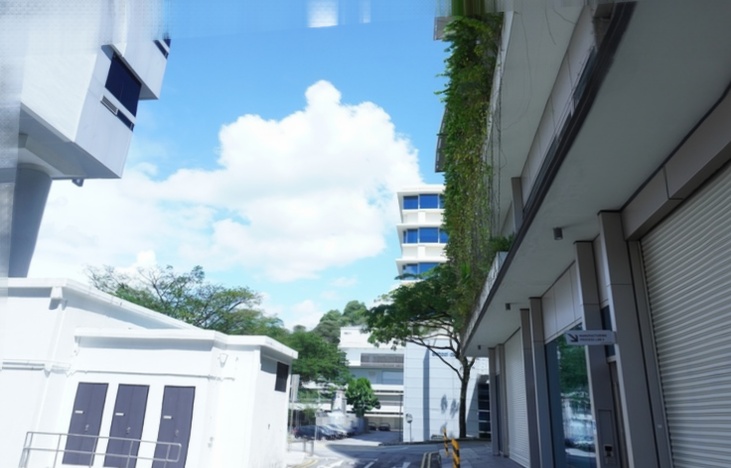} \\
Ground truth & MVSNeRF~\cite{chen2021mvsnerf} & MVSNeRF + ft & ENeRF~\cite{lin2022efficient} & ENeRF + ft \\
 & \includegraphics[width=0.2\textwidth]{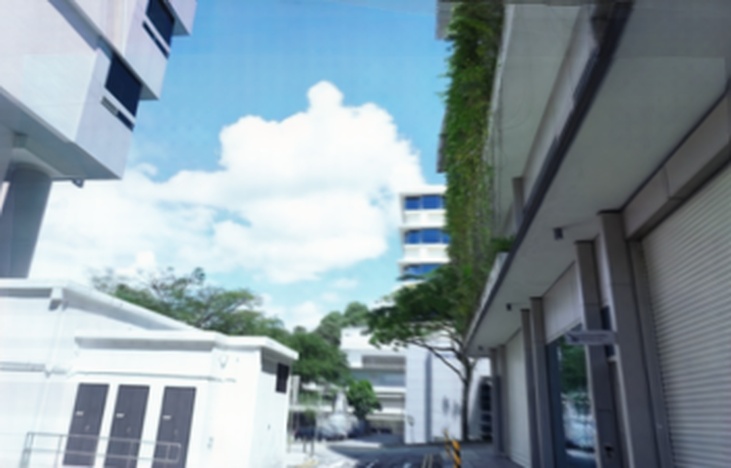} & \includegraphics[width=0.2\textwidth]{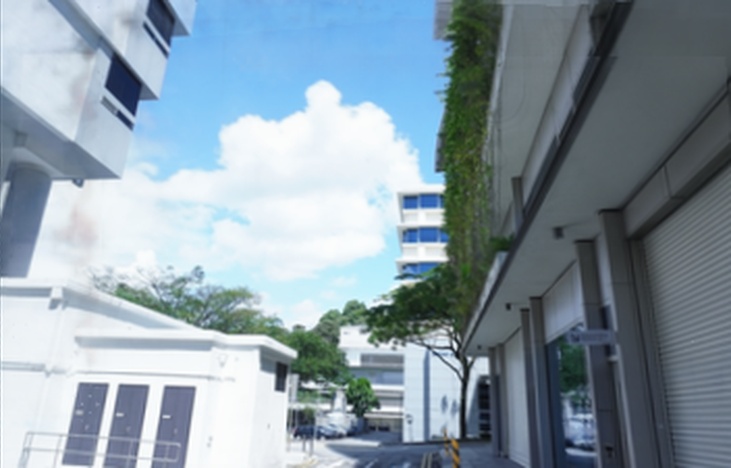} & \includegraphics[width=0.2\textwidth]{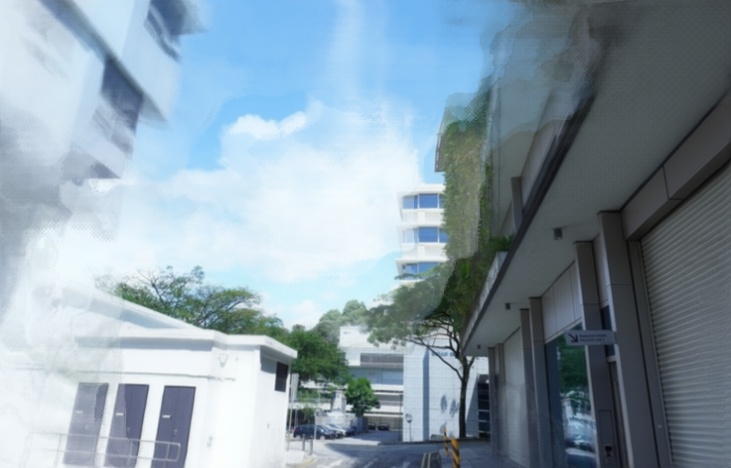} & \includegraphics[width=0.2\textwidth]{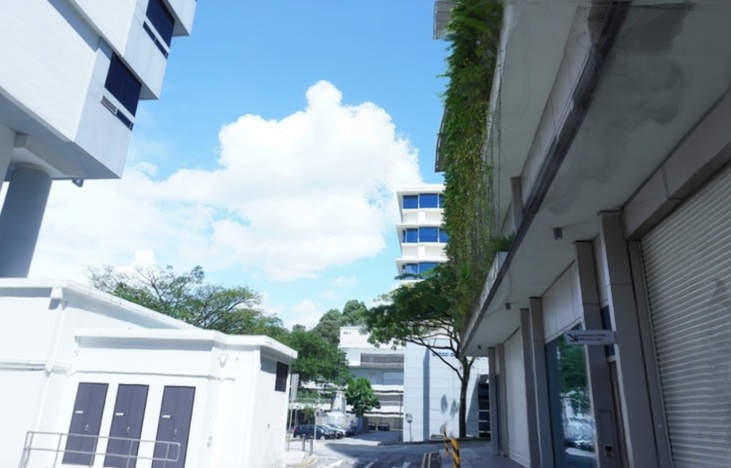}\\
 & MVSNeRF + \textbf{ours} & MVSNeRF + \textbf{ours} + ft & ENeRF + \textbf{ours} & ENeRF + \textbf{ours} + ft \\
\includegraphics[width=0.2\textwidth]{figures/results/pillar_72_0_gt.jpg} & \includegraphics[width=0.2\textwidth]{figures/results/pillar_72_0_mvsnerf.jpg} & \includegraphics[width=0.2\textwidth]{figures/results/pillar_72_0_mvsnerf_ft.jpg} & \includegraphics[width=0.2\textwidth]{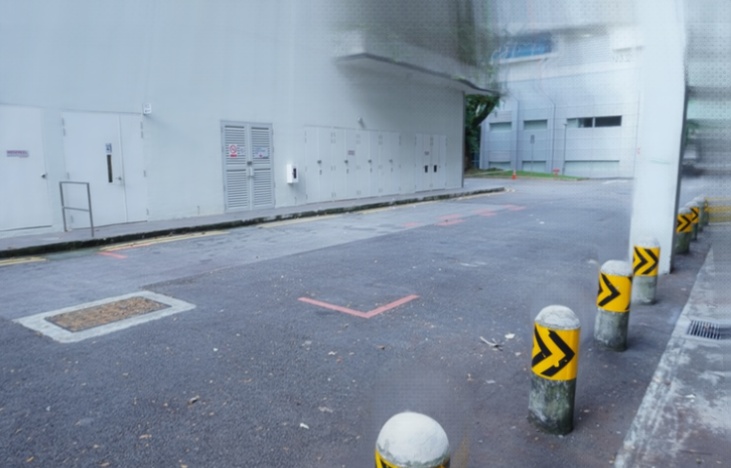} & \includegraphics[width=0.2\textwidth]{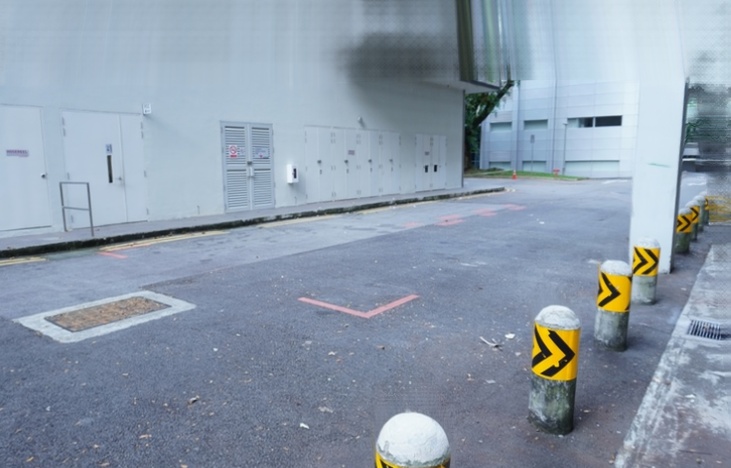} \\
Ground truth & MVSNeRF~\cite{chen2021mvsnerf} & MVSNeRF + ft & ENeRF~\cite{lin2022efficient} & ENeRF + ft \\
 & \includegraphics[width=0.2\textwidth]{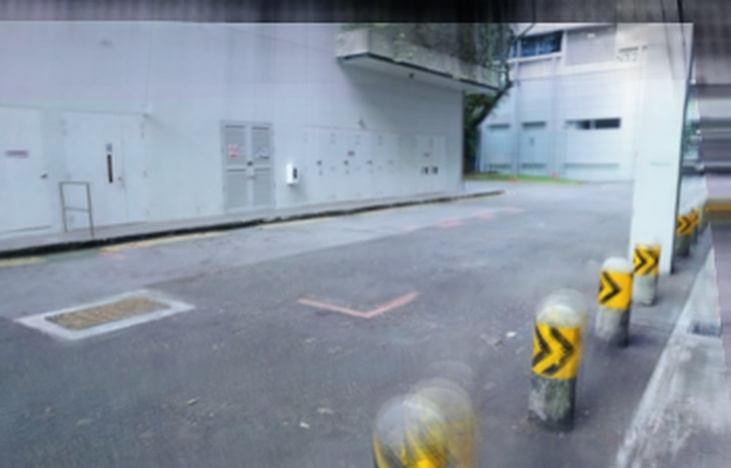} & \includegraphics[width=0.2\textwidth]{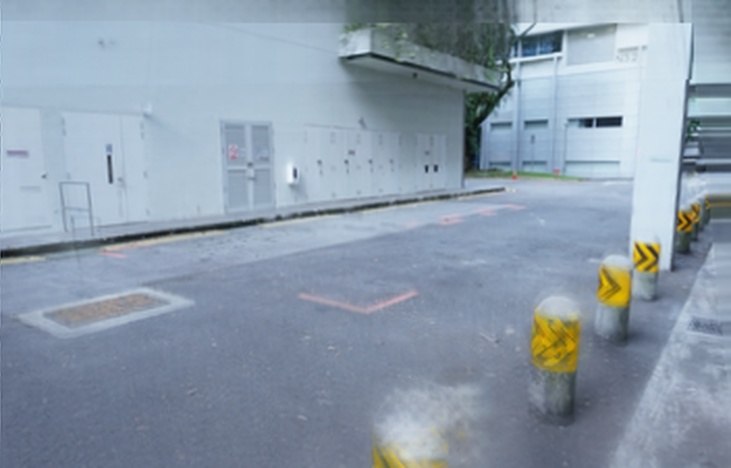} & \includegraphics[width=0.2\textwidth]{figures/results/pillar_72_0_enerf_ours.jpg} & \includegraphics[width=0.2\textwidth]{figures/results/pillar_72_0_enerf_ours_ft.jpg}\\
 & MVSNeRF + \textbf{ours} & MVSNeRF + \textbf{ours} + ft & ENeRF + \textbf{ours} & ENeRF + \textbf{ours} + ft
\end{tabular}%
}
\vspace{-2mm}
\caption{\textbf{Qualitative rendering quality improvements of integrating our method into MVS-based NeRF methods on the Free dataset.}}
\label{fig:appendix_qualitative_boostmvs_free}
\end{figure*}

\begin{figure*}[t]
\centering
\small
\setlength{\tabcolsep}{1pt}
\renewcommand{\arraystretch}{1}
\resizebox{1.0\textwidth}{!} 
{
\begin{tabular}{c|ccc|ccccc}
& \multicolumn{3}{c|}{No per-scene optimization} & \multicolumn{5}{c}{Per-scene optimization} \\
\includegraphics[width=0.13\textwidth]{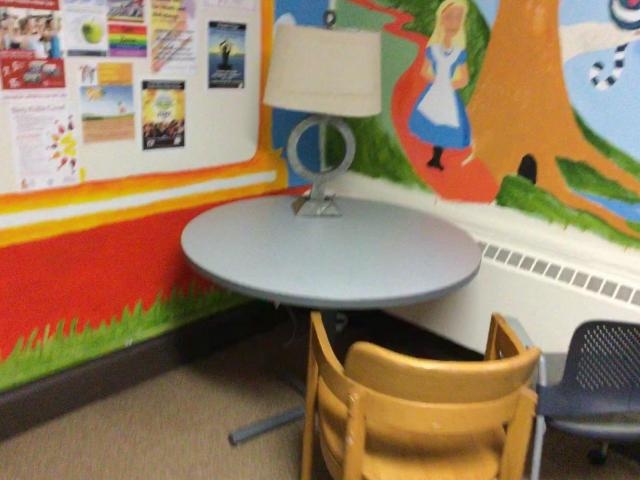} & 
\includegraphics[width=0.13\textwidth]{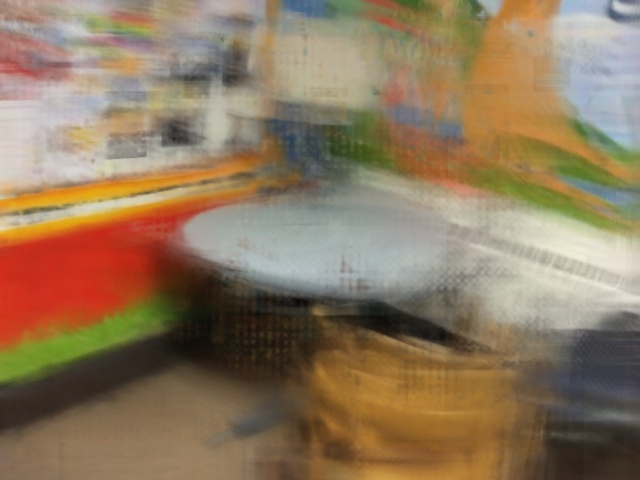} & 
\includegraphics[width=0.13\textwidth]{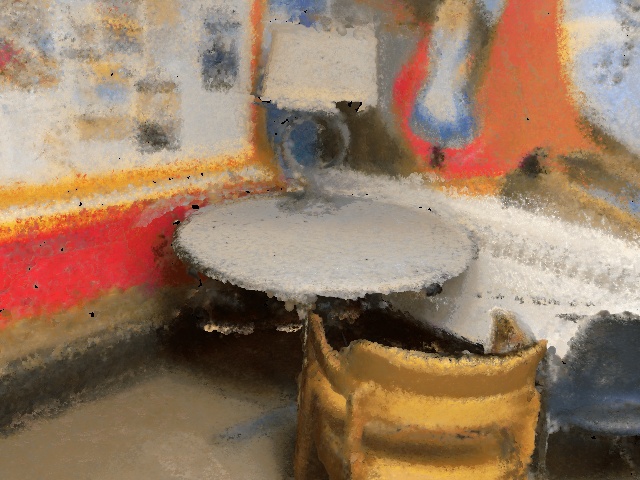} & 
\includegraphics[width=0.13\textwidth]{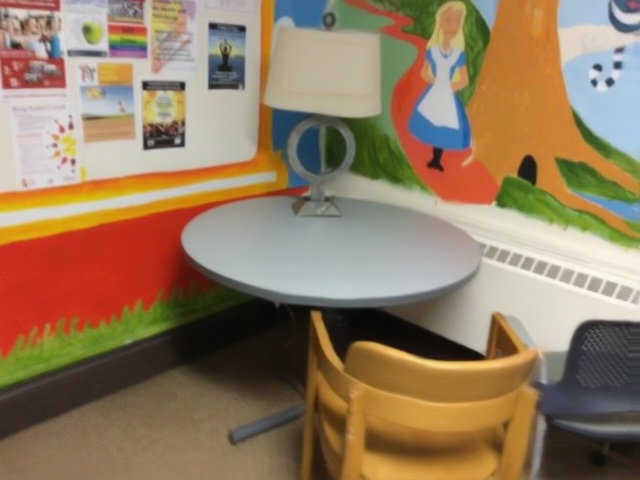} & 
\includegraphics[width=0.13\textwidth]
{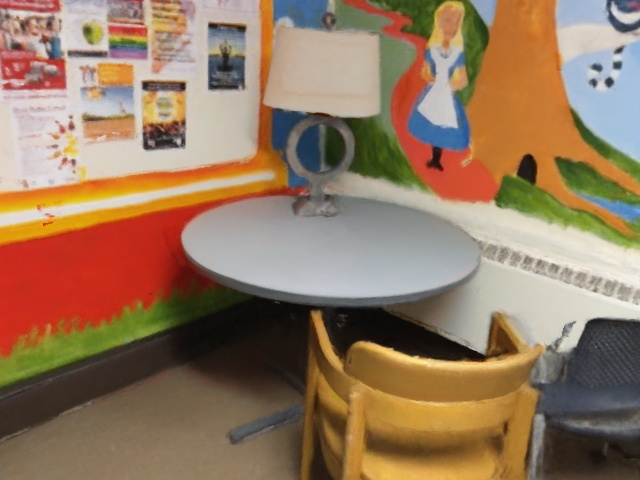} & 
\includegraphics[width=0.13\textwidth]{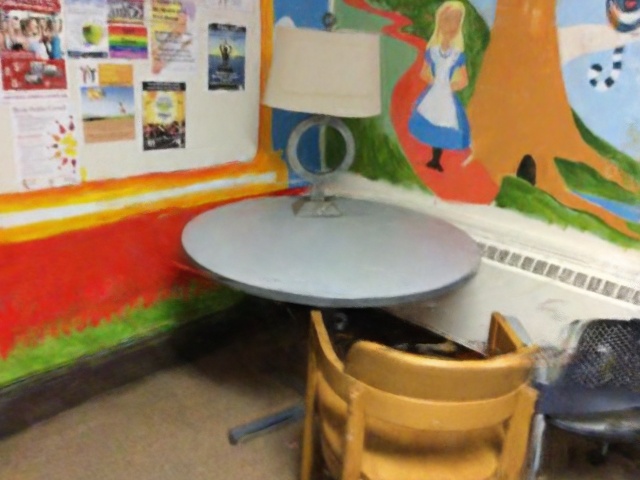} & 
\includegraphics[width=0.13\textwidth]{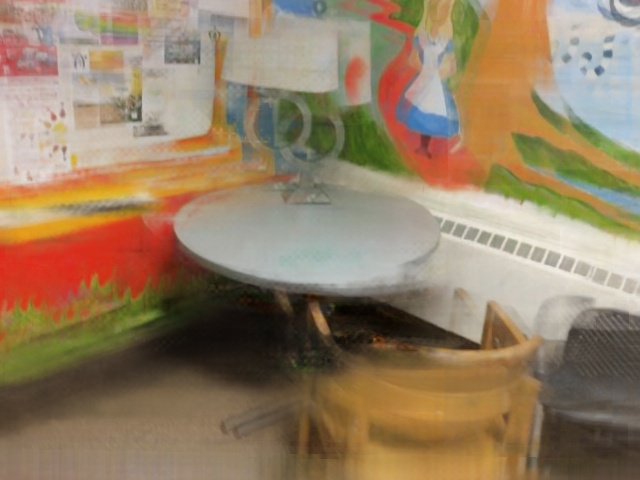} & 
\includegraphics[width=0.13\textwidth]{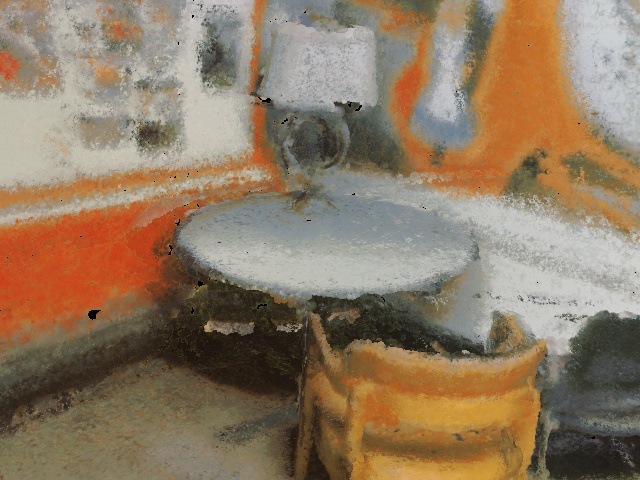} & 
\includegraphics[width=0.13\textwidth]{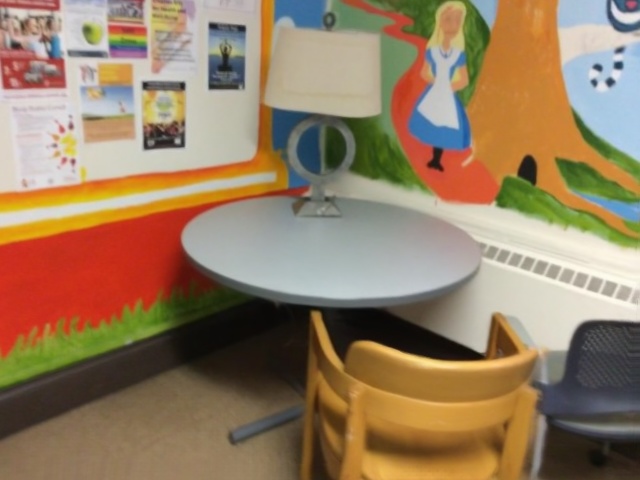} \\
\includegraphics[width=0.13\textwidth]{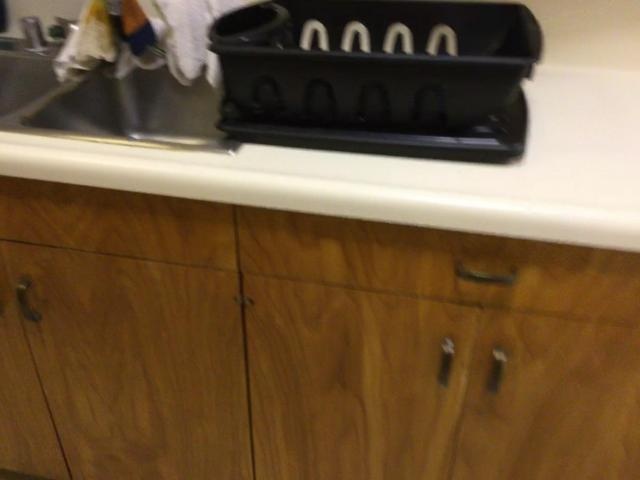} & 
\includegraphics[width=0.13\textwidth]{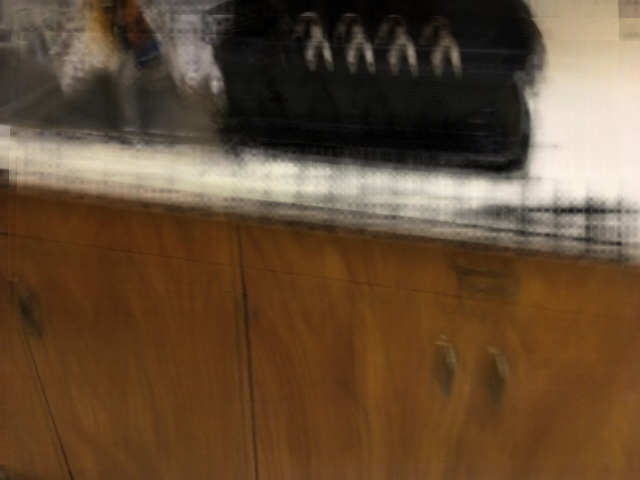} & 
\includegraphics[width=0.13\textwidth]{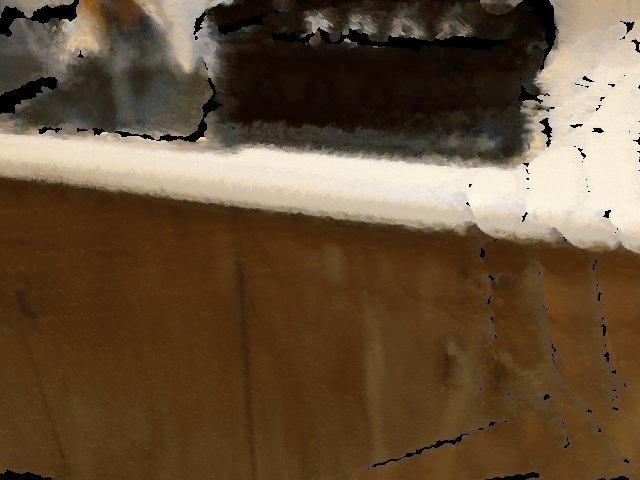} & 
\includegraphics[width=0.13\textwidth]{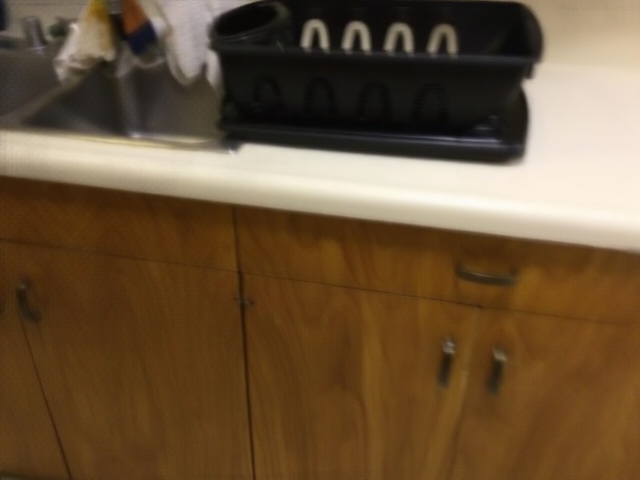} & 
\includegraphics[width=0.13\textwidth]{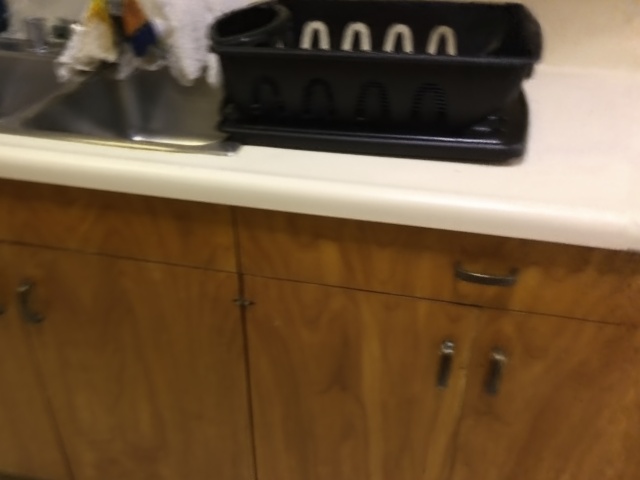} & 
\includegraphics[width=0.13\textwidth]{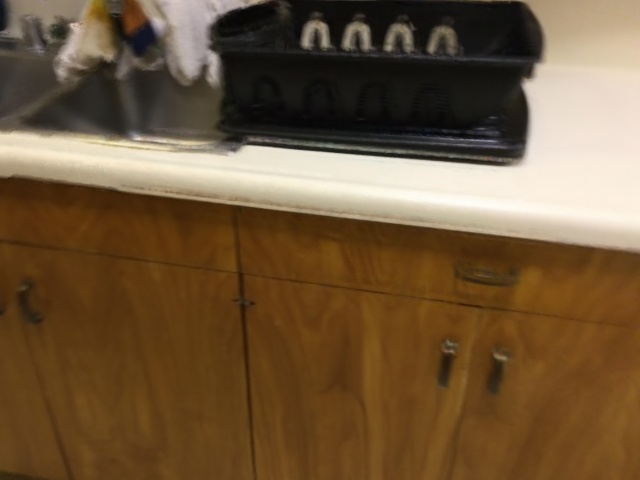} & 
\includegraphics[width=0.13\textwidth]{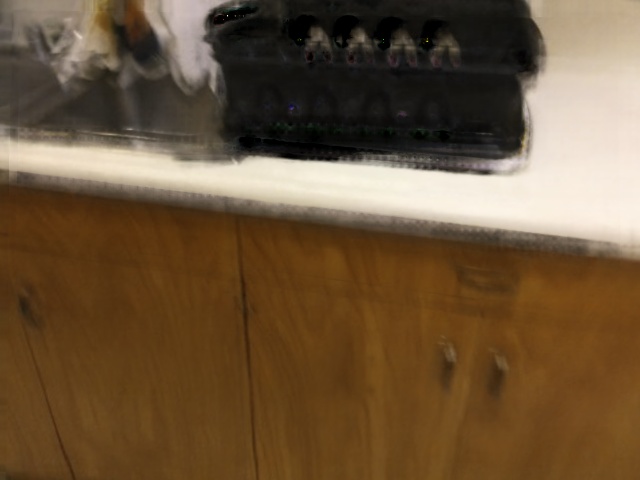} & 
\includegraphics[width=0.13\textwidth]{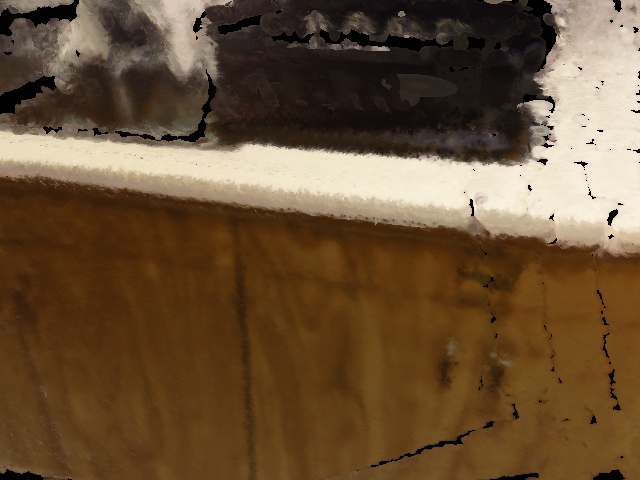} & 
\includegraphics[width=0.13\textwidth]{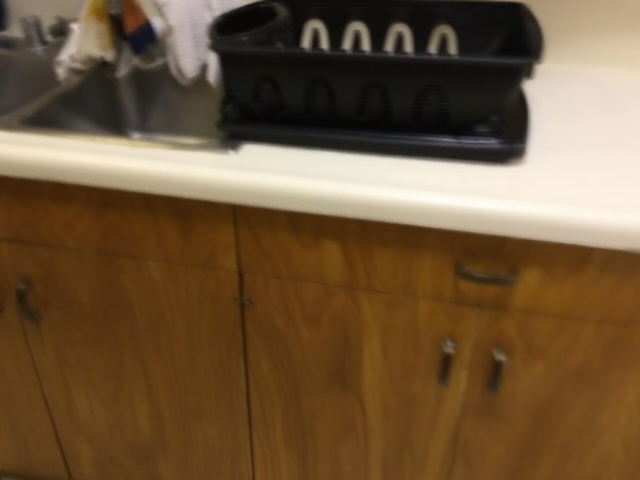} \\
\includegraphics[width=0.13\textwidth]{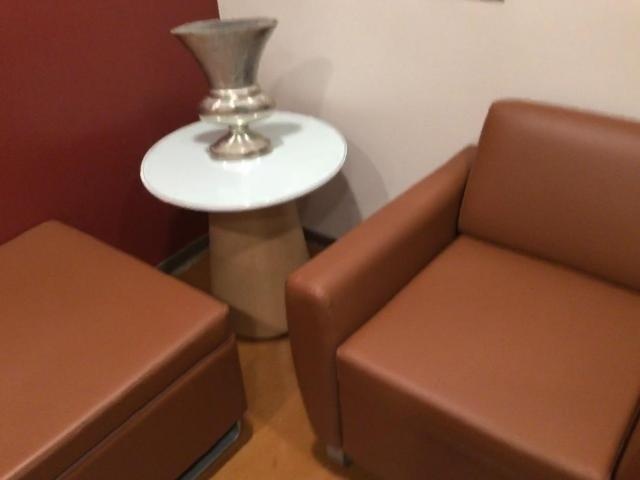} & 
\includegraphics[width=0.13\textwidth]{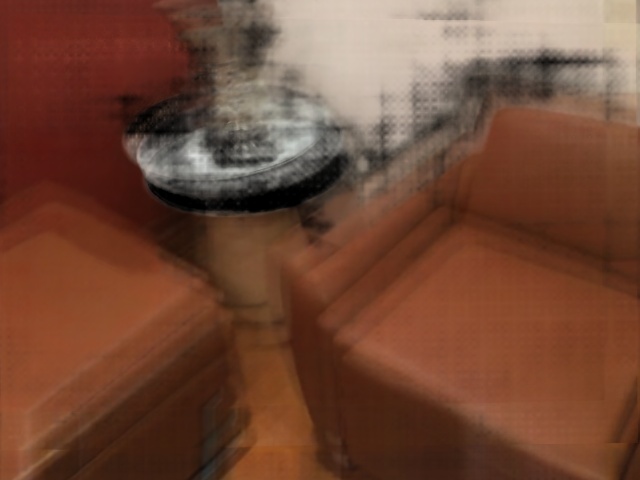} & 
\includegraphics[width=0.13\textwidth]{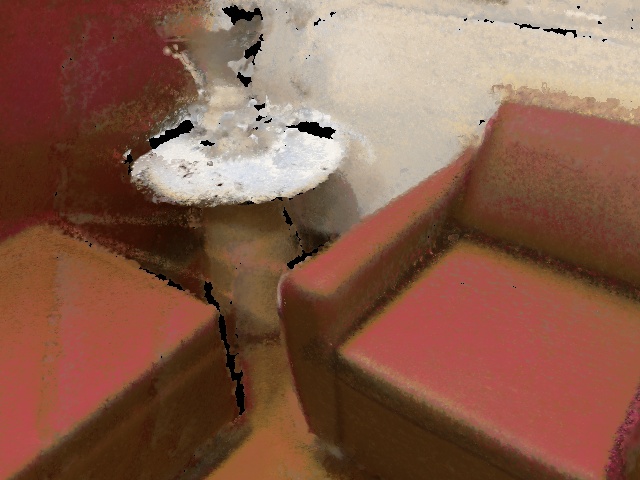} & 
\includegraphics[width=0.13\textwidth]{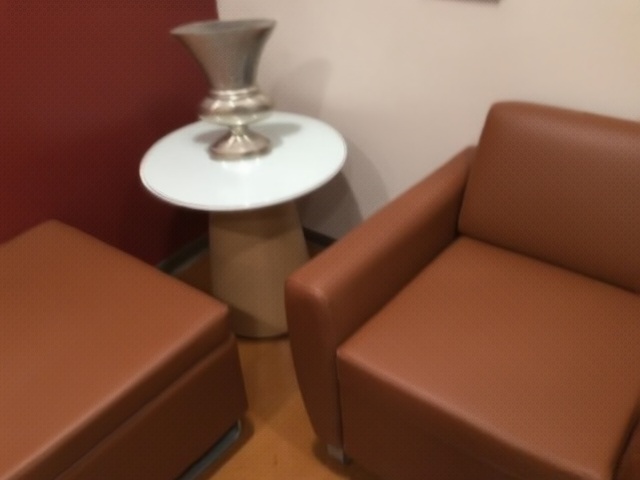} & 
\includegraphics[width=0.13\textwidth]{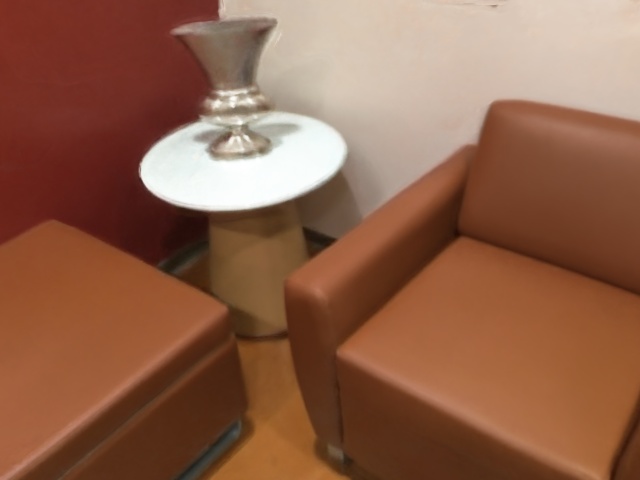} & 
\includegraphics[width=0.13\textwidth]{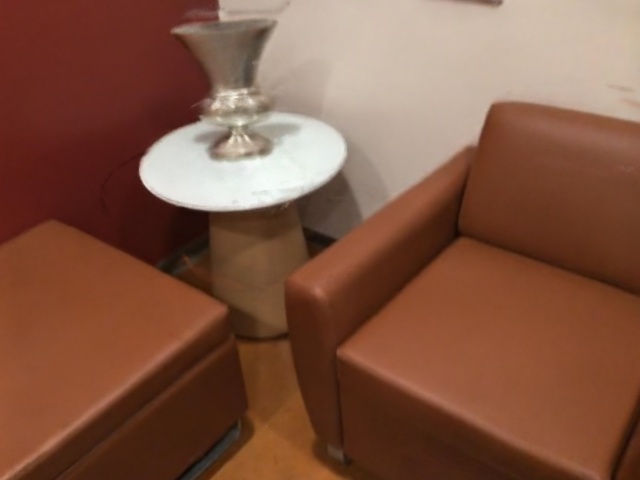} & 
\includegraphics[width=0.13\textwidth]{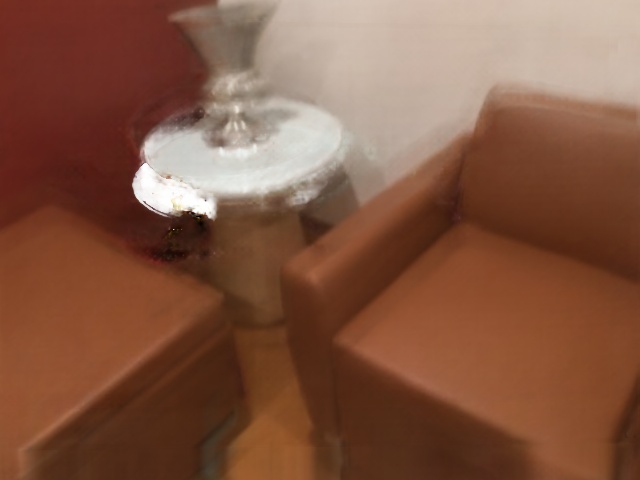} & 
\includegraphics[width=0.13\textwidth]{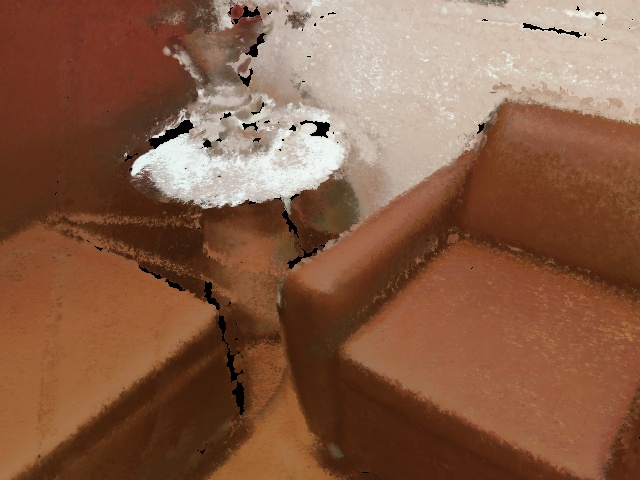} & 
\includegraphics[width=0.13\textwidth]{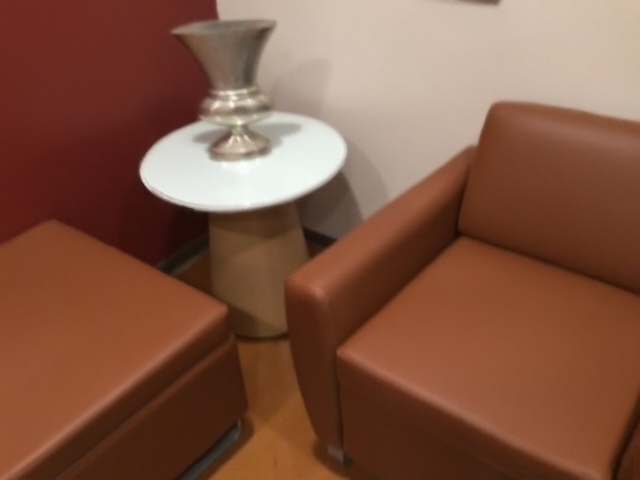} \\
\includegraphics[width=0.13\textwidth]{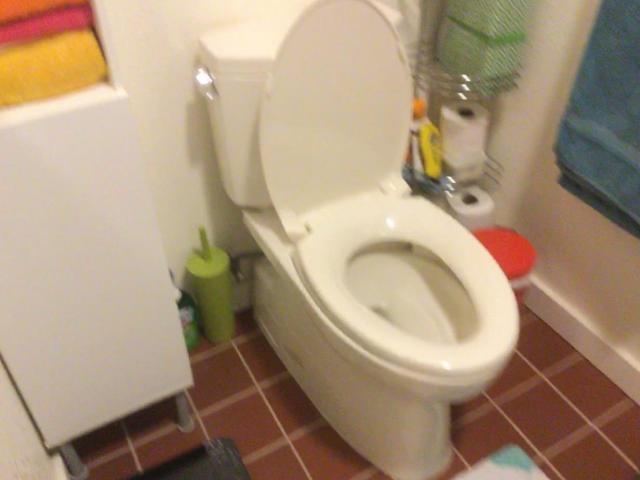} & 
\includegraphics[width=0.13\textwidth]{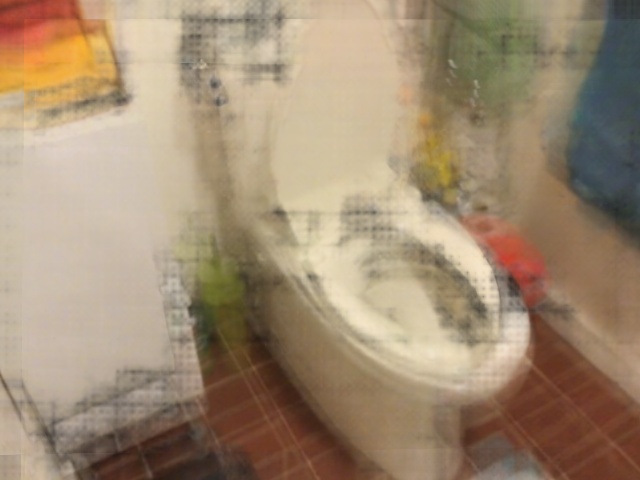} & 
\includegraphics[width=0.13\textwidth]{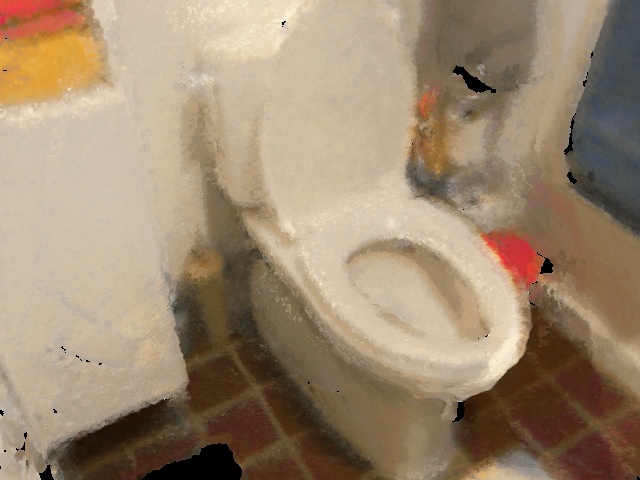} & 
\includegraphics[width=0.13\textwidth]{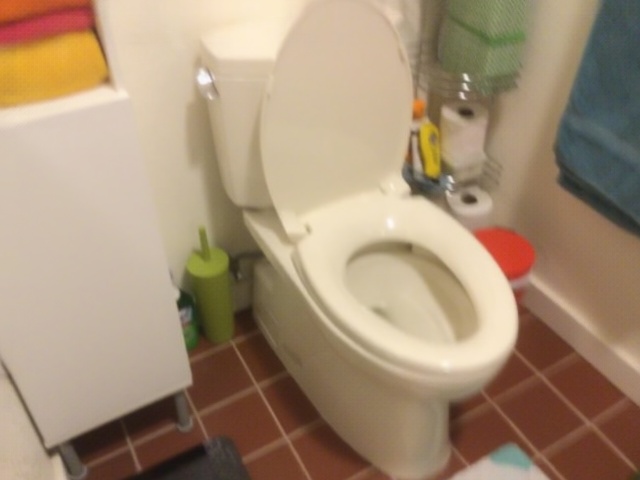} & 
\includegraphics[width=0.13\textwidth]{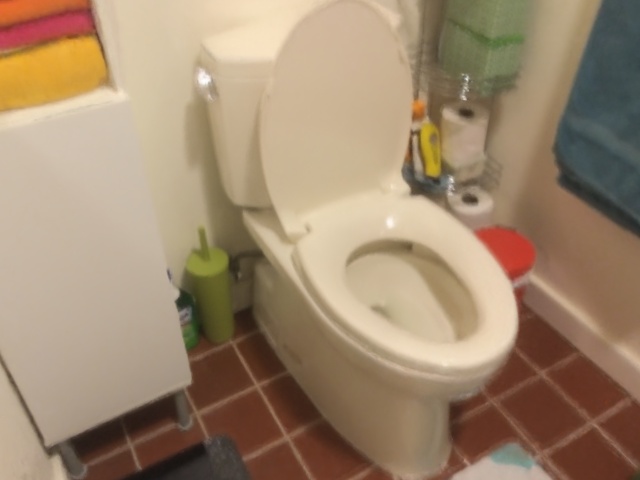} & 
\includegraphics[width=0.13\textwidth]{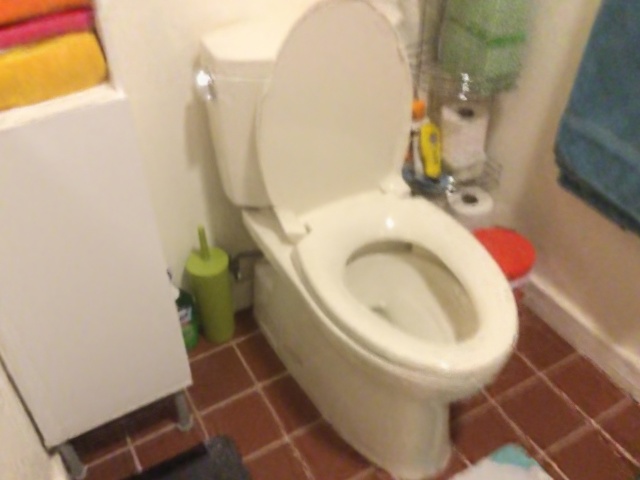} & 
\includegraphics[width=0.13\textwidth]{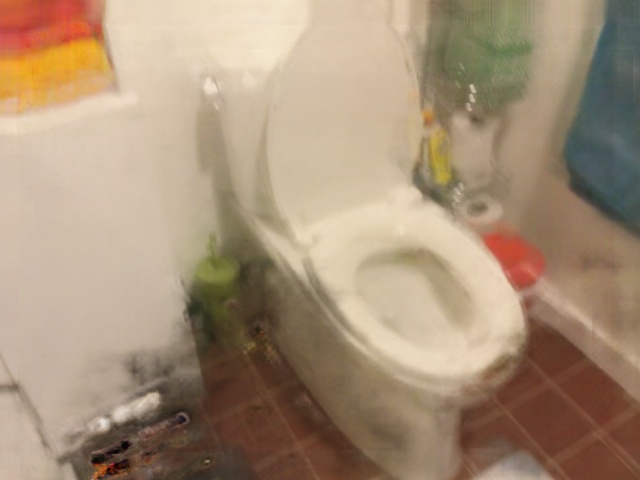} & 
\includegraphics[width=0.13\textwidth]{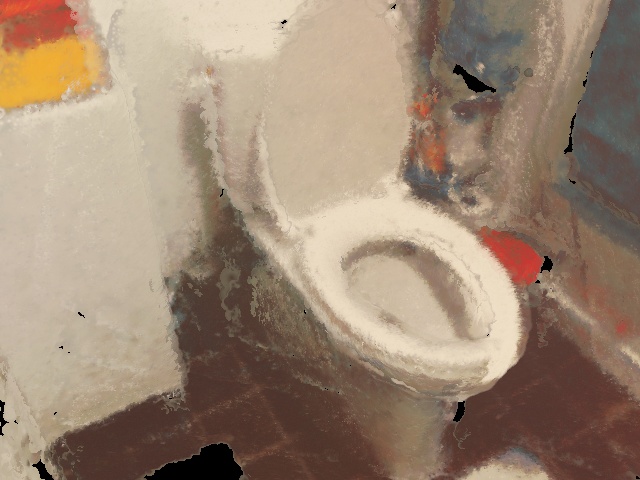} & 
\includegraphics[width=0.13\textwidth]{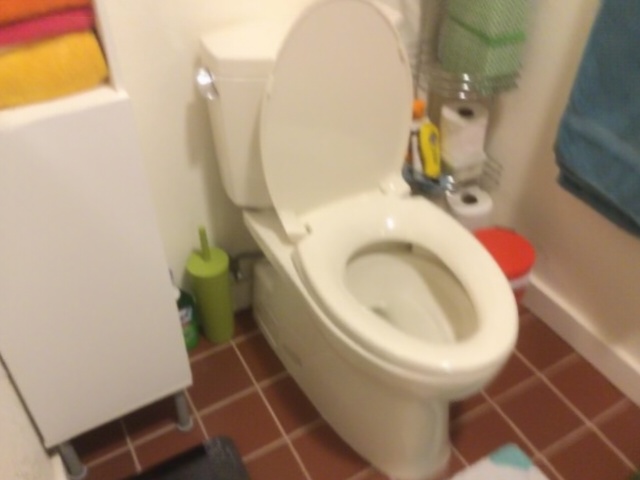} \\
Ground truth & MVSNeRF & SurfelNeRF & Ours & F2-NeRF & Zip-NeRF & MVSNeRF + ft & SurfelNeRF + ft & Ours + ft \\
& \cite{chen2021mvsnerf} & \cite{gao2023surfelnerf} & & \cite{wang2023f2} & \cite{barron2023zip} & \cite{chen2021mvsnerf} & \cite{gao2023surfelnerf}\\
\end{tabular}%
}
\caption{\textbf{Qualitative comparisons of rendering quality on the ScanNet~\cite{dai2017scannet} dataset.}}
\label{fig:appendix_qualitative_scannet}
\end{figure*}

\begin{figure*}[t]
\centering
\small
\setlength{\tabcolsep}{1pt}
\renewcommand{\arraystretch}{1}
\resizebox{1.0\textwidth}{!} 
{
\begin{tabular}{ccccc}
\includegraphics[width=0.2\textwidth]{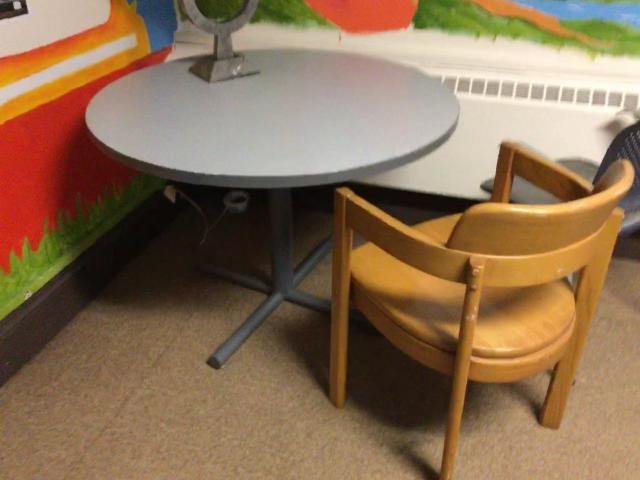} & \includegraphics[width=0.2\textwidth]{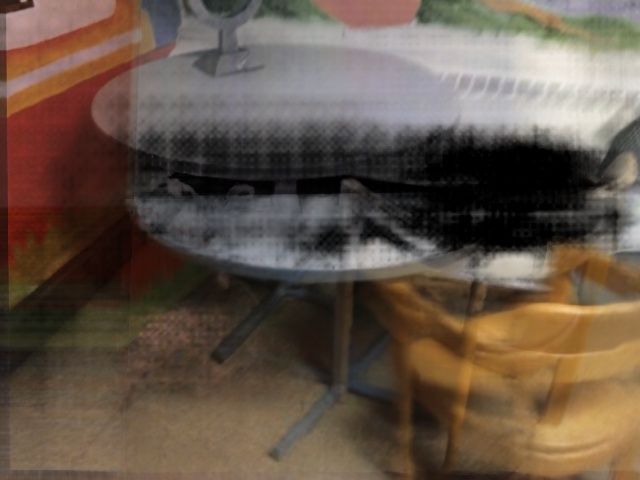} & \includegraphics[width=0.2\textwidth]{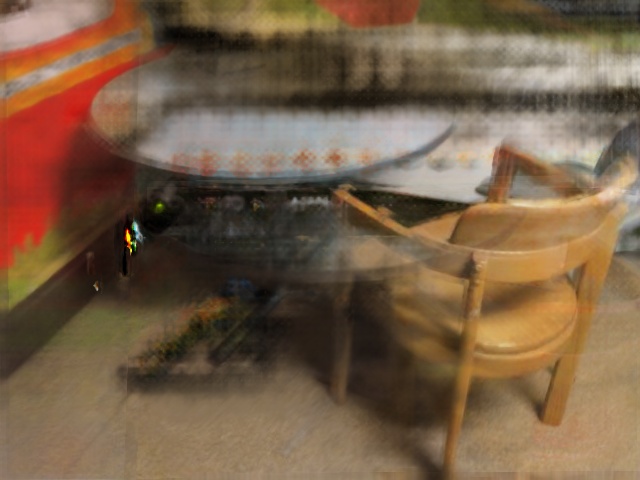} & \includegraphics[width=0.2\textwidth]{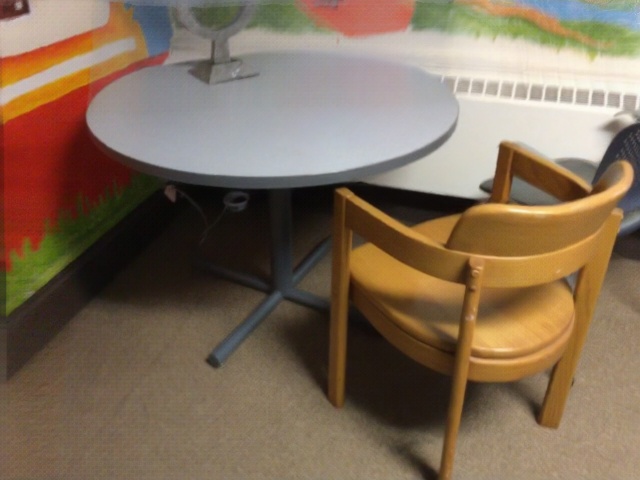} & \includegraphics[width=0.2\textwidth]{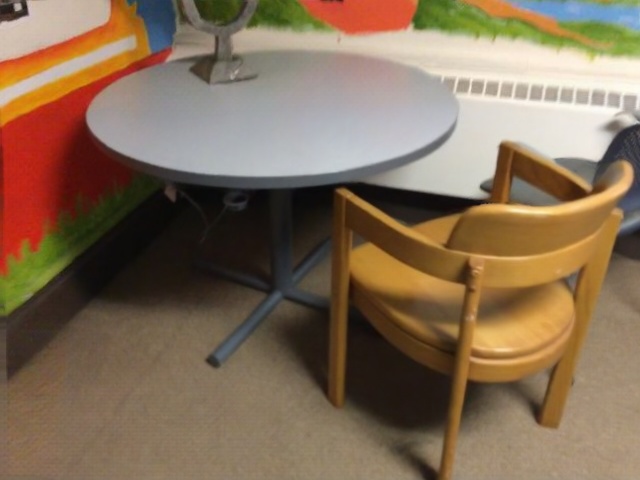} \\
Ground truth & MVSNeRF~\cite{chen2021mvsnerf} & MVSNeRF + ft & ENeRF~\cite{lin2022efficient} & ENeRF + ft \\
 & \includegraphics[width=0.2\textwidth]{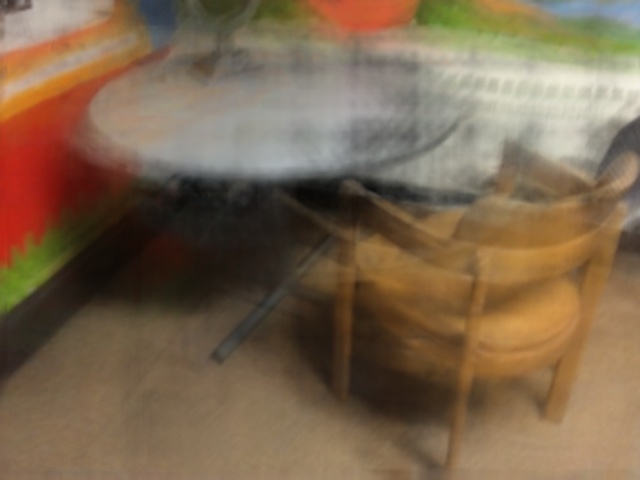} & \includegraphics[width=0.2\textwidth]{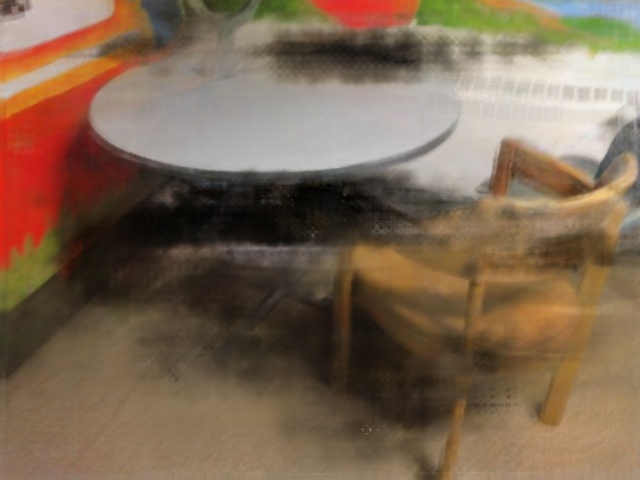} & \includegraphics[width=0.2\textwidth]{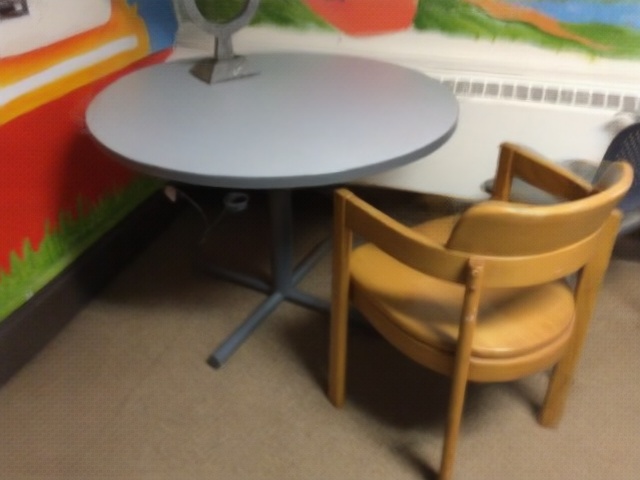} & \includegraphics[width=0.2\textwidth]{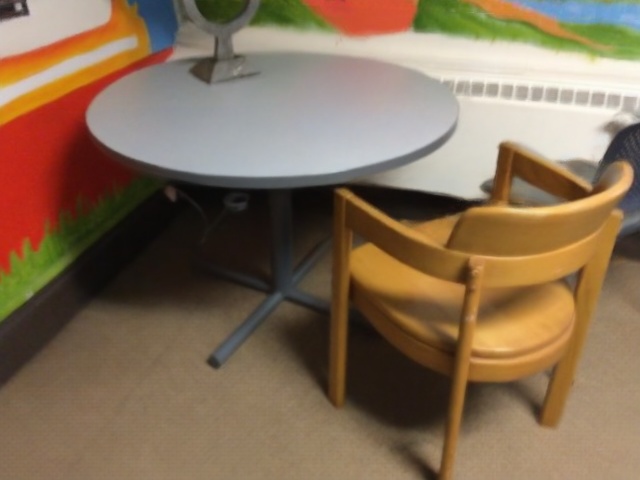}\\
 & MVSNeRF + \textbf{ours} & MVSNeRF + \textbf{ours} + ft & ENeRF + \textbf{ours} & ENeRF + \textbf{ours} + ft \\
\includegraphics[width=0.2\textwidth]{figures/results/scene0158_00_144_0_gt.jpg} & \includegraphics[width=0.2\textwidth]{figures/results/scene0158_00_144_0_mvsnerf.jpg} & \includegraphics[width=0.2\textwidth]{figures/results/scene0158_00_144_0_mvsnerf_ft.jpg} & \includegraphics[width=0.2\textwidth]{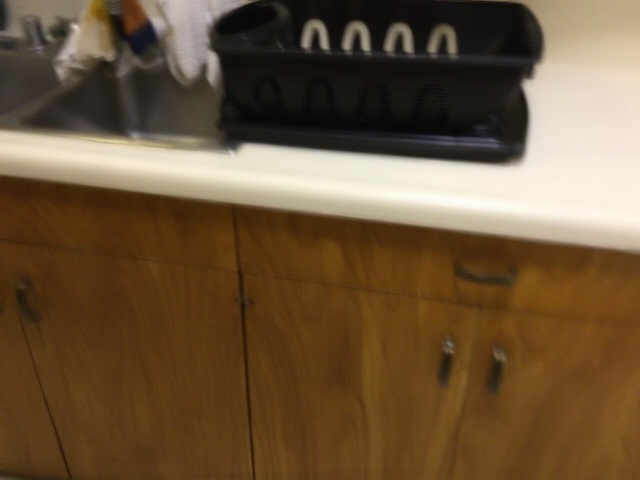} & \includegraphics[width=0.2\textwidth]{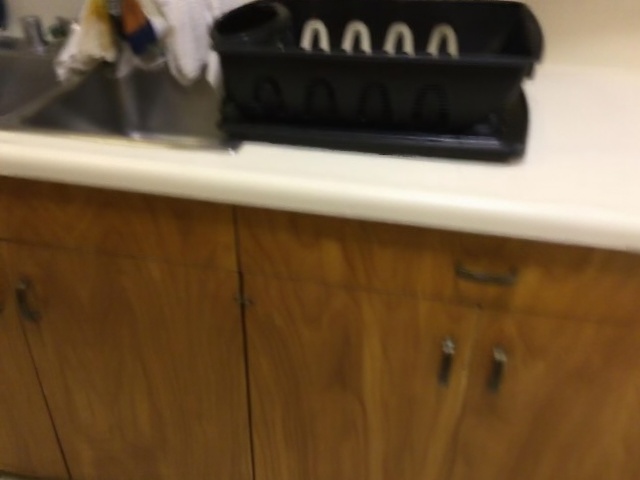} \\
Ground truth & MVSNeRF~\cite{chen2021mvsnerf} & MVSNeRF + ft & ENeRF~\cite{lin2022efficient} & ENeRF + ft \\
 & \includegraphics[width=0.2\textwidth]{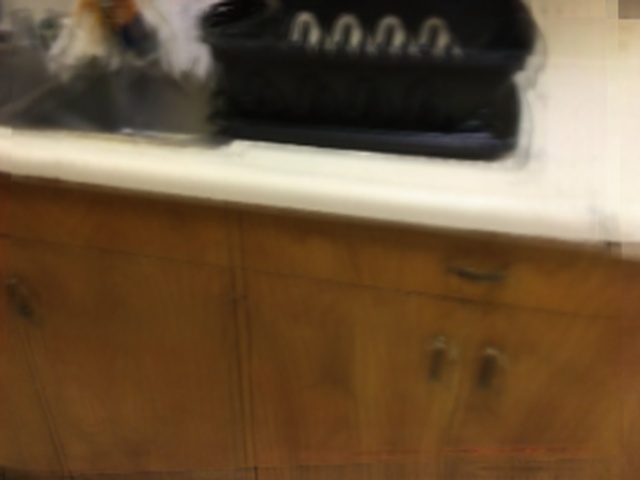} & \includegraphics[width=0.2\textwidth]{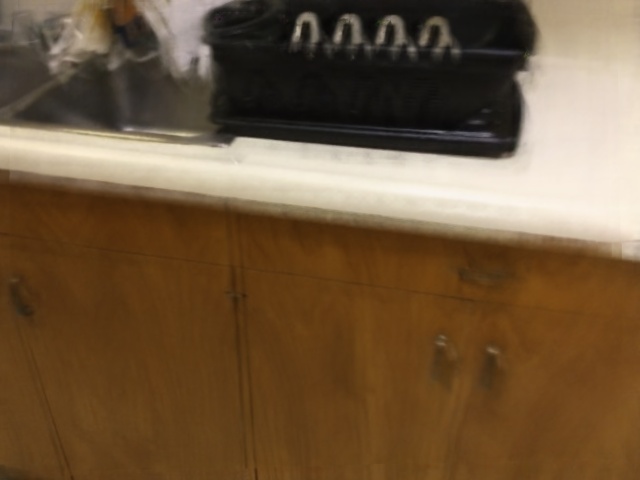} & \includegraphics[width=0.2\textwidth]{figures/results/scene0158_00_144_0_enerf_ours.jpg} & \includegraphics[width=0.2\textwidth]{figures/results/scene0158_00_144_0_enerf_ours_ft.jpg}\\
 & MVSNeRF + \textbf{ours} & MVSNeRF + \textbf{ours} + ft & ENeRF + \textbf{ours} & ENeRF + \textbf{ours} + ft
\end{tabular}%
}
\caption{\textbf{Qualitative rendering quality improvements of integrating our method into MVS-based NeRF methods on the ScanNet dataset.}}
\label{fig:qualitative_boostmvs_scannet}
\end{figure*}

\clearpage

\section{Appendix Overview}
This supplementary material presents additional results to complement the main paper.
First, we provide the detailed derivation of the combined volume rendering equation with multiple cost volumes in Sec.~\ref{sec:derivation}.
Then, we describe all the implementation details of BoostMVSNeRFs and baseline methods in Sec.~\ref{sec:implementation_details}.
Next, we show additional ablation studies, including the sensitivity analysis on the number of selected cost volumes and the effect of combining multiple cost volumes in 2D or in the 3D space in Sec.~\ref{sec:additional_ablation}.
Finally, we provide complete quantitative evaluations and additional qualitative comparisons in Sec.~\ref{sec:additional_exps} and Sec.~\ref{sec:additional_exps_visual}, respectively.
In addition to this document, we provide video results of our method and state-of-the-art methods and show the rendering quality comparison.
%

\section{Derivation of the Combined Volume Rendering with Multiple Cost Volumes} \label{sec:derivation}
Our proposed rendering differs from the traditional one by considering 3D visibility scores and combining multiple cost volumes.
Below, we explain the modifications we make.
First, let us only consider a single cost volume for simplicity. The pixel color output is given by:
\begin{equation} \label{eq:single_cv}
    C_{\text{single}}(\textbf{r})= \sum_{j=1}^{J} T_{\text{single}}(j) \alpha\left(\sigma_j \delta_j \right) m_j c_j, 
\end{equation}  
\begin{align}
    T_{\text{single}}(j) &= \prod_{s=1}^{j-1} \exp\left(-\sigma_s \delta_s \right) m_s \\
    &= \prod_{s=1}^{j-1} \exp\left(-\sigma_s \delta_s \right) \exp\left(\ln m_s \right) \\
    &= \prod_{s=1}^{j-1} \exp\left(-\left(\sigma_s \delta_s - \ln m_s \right)\right) \\
    &= \exp \left( - \sum_{s=1}^{j-1} \left( \sigma_s \delta_s - \ln m_s \right)\right).
\end{align}

To further consider multiple cost volumes and also utilize their corresponding 3D visibility scores, we modify Eq. \ref{eq:single_cv} to combine the result across multiple cost volumes. The final proposed volume rendering is given by:

\begin{equation} \label{eq:combine}
C(\textbf{r}) = \sum_{k=1}^K \sum_{j=1}^{J} T_{\text{combined}}(j) \alpha\left(\sigma^k_j \delta_j \right) M^k_j c^k_j, 
\end{equation}
\begin{equation}
T_{\text{combined}}(j) = \sum_{k=1}^K \exp \left( - \sum_{s=1}^{j-1} \left( \sigma_s^k \delta_s - \ln M_s^k \right)\right),
\end{equation}
\begin{equation}
M_j^k = \frac{m^k_j}{\sum_{k=1}^K m^k_j},
\end{equation}
where $K$ is the number of selected cost volumes,
and $M_j^k$ is the normalized 3D visibility score so that the summation of 3D visibility scores over selected cost volumes equals 1.



\section{Implementation Details} \label{sec:implementation_details}
Our method is compatible with MVS-based NeRF techniques, allowing us to employ pre-trained models such as MVSNeRF and ENeRF in our experiments. For per-scene optimization, we fine-tune our methods with 11,000 iterations, following the settings of ENeRF, with an initial learning rate of \( 5e^{-4} \) and an exponential scheduler. This fine-tuning process typically takes between 1 to 2 hours on an RTX 4090 GPU, depending on the image resolution.

For a fair comparison across methods, we evaluate them using a 731 x 468 image resolution in the Free dataset and a $640 \times 480$ image resolution in the ScanNet dataset. During fine-tuning, we use a resolution of $736 \times 480$ to match our model architecture and subsequently downsample the images to $731 \times 468$ for fair comparison.


In practical terms, both ENeRF and our models are configured with 2 samples per ray. When integrating MVSNeRF, we use 32 samples per ray to evaluate its generalizable rendering model; however, during fine-tuning, this is reduced to 8 samples per ray to address convergence issues. We employ 4 cost volumes (K=4) for combined rendering with multiple cost volumes.
. For each rendering view in both datasets, we consider its nearest 6 training views ($N=6$) and build 
\( C^N_3 \) cost volumes and select $K$ best cost volumes for our method.

\section{Additional Ablation Studies} \label{sec:additional_ablation}

\subsection{Sensitivity analysis on the number of selected cost volumes}
In our approach, we introduce a greedy method for cost volume selection for fusion, enabling the integration of multiple cost volumes. To optimize the most effective cost volume fusion strategy, we conduct experiments by considering $K$ cost volumes during volume rendering. We explore scenarios where $K$ equals 1, 2, 3, 4, and 5, and present the corresponding results in Table~\ref{tab:ablation_K}.

\begin{table}[t]
\centering
\caption{
\textbf{Sensitivity analysis on the number of selected cost
volumes for combination.}
We compare the rendering quality of different numbers of selected cost volumes for combined rendering on all scenes of the Free~\cite{wang2023f2} dataset. The rendering quality improves with more cost volumes selected in combined rendering.
}
\label{tab:ablation_K}
\begin{tabular}{lcccc}
\toprule
Method & Setting & PSNR $\uparrow$ & SSIM $\uparrow$ & LPIPS $\downarrow$ \\
\midrule
$K=1$ & \multirow{5}{*}{\makecell{No per-scene\\optimization}} & 24.05 & 0.865 & 0.205 \\
$K=2$ & & 24.55 & 0.872 & \textbf{0.202} \\
$K=3$ & & 24.68 & \textbf{0.875} & \textbf{0.202} \\
$K=4$ & & 24.75 & \textbf{0.875} & \textbf{0.202} \\ 
$K=5$ & & \textbf{24.79} & \textbf{0.875} & 0.203 \\
\bottomrule
\end{tabular}
\end{table}

\subsection{Continuous-valued 3D visibility scores.} 
Our method represents 3D visibility scores with continuous values.
We conduct experiments comparing our method with binary-valued 3D visibility scores, \ie, each score can only be either 0 or 1. The binary-valued 3D visibility scores $m_j'$ are calculated by $m_j' = 1 - \prod_{i=1}^{V} (1 - \mathbb{1}_i(p))$, where $m_j'$ is the binary-valued 3D visibility scores.

\begin{table}[t]
\centering
\caption{
\textbf{Effect of continuous or binary 3D visibility scores.}
We compare representing the 3D visibility score using binary values with our continuous values on all scenes of the Free~\cite{wang2023f2} dataset. Representing 3D visibility scores with continuous values performs better than binary values.}
\label{tab:ablation_discrete_mask}
\resizebox{\columnwidth}{!}{%
\begin{tabular}{lcccc}
\toprule
Method & Setting & PSNR $\uparrow$ & SSIM $\uparrow$ & LPIPS $\downarrow$ \\
\midrule
Discrete mask & \multirow{2}{*}{\makecell{No per-scene\\optimization}} & 24.05 & 0.855 & 0.227 \\
Continuous mask (Ours) & & \textbf{24.21} & \textbf{0.862} & \textbf{0.218} \\
\bottomrule
\end{tabular}%
}
\end{table}

\begin{table}[t]
\centering
\small
\caption{
\textbf{
Ablation on different combined methods with multiple cost volumes.} We compare the rendering quality between different combined methods on all scenes of the ScanNet dataset. Combined rendering in 3D performs slightly better than in the 2D space.}
\label{tab:ablation_early_late_fusion}
\resizebox{\columnwidth}{!}{%
\begin{tabular}{lccc}
\toprule
Method & PSNR $\uparrow$ & SSIM $\uparrow$ & LPIPS $\downarrow$ \\
\midrule
3D visibility scores as multipliers on densities & 17.60 & 0.767 & 0.549 \\
Rendered images and 2D visibility masks & 24.20 & \textbf{0.868} & 0.360 \\
Sampled 3D points and 3D visibility scores (Ours) & \textbf{24.22} & \textbf{0.868} & \textbf{0.361} \\
\bottomrule
\end{tabular}%
}
\end{table}

\subsection{Different combined methods with multiple cost volumes}
In our method, we use calculated 3D visibility scores as weights to combine multiple cost volumes during rendering.
A straightforward variant is to multiply the 3D visibility scores onto queried density values in volume rendering.
Yet another way is to blend rendering results from different cost volumes in the 2D image domain instead of 3D sampled points. 
Table~\ref{tab:ablation_early_late_fusion} demonstrates that our combination method using 3D sampled points and 3D visibility scores achieves the best rendering quality.

\section{Complete Quantitative Evaluations} \label{sec:additional_exps}

\subsection{Free dataset}
We show all 7 scenes of the quantitative comparisons on the Free~\cite{wang2023f2} dataset in Table~\ref{tab:complete_quantitative_free}.

\begin{table*}[t]
\centering
\caption{\textbf{Complete quantitative comparisons with state-of-the-art methods on the Free~\cite{wang2023f2} dataset.}}
\label{tab:complete_quantitative_free}
\begin{tabular}{lcccccccc}
\toprule
Method & Setting & Hydrant & Lab & Pillar & Road & Sky & Stair & Grass\\
\midrule
 & & \multicolumn{7}{c}{PSNR $\uparrow$} \\
\midrule
MVSNeRF~\cite{chen2021mvsnerf} & \multirow{4}{*}{\makecell{No per-scene\\optimization}} & 19.63 & 20.11 & 19.58 & 20.84 & 18.58 & 21.98 & 19.69 \\
MVSNeRF + Ours &  & 20.21 & 20.00 & 20.02 & 21.77 & 19.50 & 21.64 & 20.47\\
ENeRF~\cite{lin2022efficient} &  & 21.85 & 23.48 & 24.3 & 24.73 & 22.27 & 25.52 & 20.53 \\
ENeRF + Ours &  & \textbf{22.73} & \textbf{23.94} & \textbf{26.06} & \textbf{25.88} & \textbf{23.44} & \textbf{25.78} & \textbf{21.64} \\
\midrule
F2-NeRF~\cite{wang2023f2} & \multirow{6}{*}{\makecell{Per-scene\\optimization}} & 	23.75 & 24.34 & 28.05 & 26.03 & 25.10 & 28.14 & 23.44\\
Zip-NeRF~\cite{barron2023zip} & & \textbf{25.43} & \textbf{27.94} & 25.30 & \textbf{28.83} & \textbf{27.12} & 28.21 & 18.46\\
$\text{MVSNeRF}_\text{ft}$~\cite{chen2021mvsnerf} &  & 19.33 & 18.90 & 21.22 & 21.88 & 19.42 & 21.62 & 21.08 \\
MVSNeRF + $\text{Ours}_\text{ft}$ &  & 20.81 & 19.05 & 21.75 & 24.31 & 20.04 & 22.18 & 23.02 \\
$\text{ENeRF}_\text{ft}$~\cite{lin2022efficient} &  & 23.35 & 24.87 & 27.48 & 26.43 & 23.65 & 27.43 & 23.15 \\
ENeRF + $\text{Ours}_\text{ft}$ &  & 24.28 & 25.83 & \textbf{28.50} & 27.64 & 24.31 & \textbf{28.28} & \textbf{24.12} \\
\midrule
 & & \multicolumn{7}{c}{SSIM $\uparrow$} \\
\midrule
MVSNeRF~\cite{chen2021mvsnerf} & \multirow{4}{*}{\makecell{No per-scene\\optimization}} & 0.689 & 0.757 & 0.698 & 0.755 & 0.744 & 0.770 & 0.631 \\
MVSNeRF + Ours &  & 0.691 & 0.745 & 0.706 & 0.762 & 0.748 & 0.763 & 0.636 \\
ENeRF~\cite{lin2022efficient} &  & 0.812 & 0.881 & 0.854 & 0.868 & 0.873 & 0.891 & 0.729 \\
ENeRF + Ours &  & \textbf{0.837} & \textbf{0.889} & \textbf{0.888} & \textbf{0.889} & \textbf{0.881} & \textbf{0.897} & \textbf{0.755}	\\
\midrule
F2-NeRF~\cite{wang2023f2} & \multirow{6}{*}{\makecell{Per-scene\\optimization}} & 	0.743 & 0.825 & 0.794 & 0.802 & 0.856 & 0.835 & 0.581\\
Zip-NeRF~\cite{barron2023zip} & & 0.818 & 0.902 & 0.748 & 0.880 & 0.889 & 0.855 & 0.313\\
$\text{MVSNeRF}_\text{ft}$~\cite{chen2021mvsnerf} &  & 0.645 & 0.697 & 0.690 & 0.746 & 0.718 & 0.747 & 0.646 \\
MVSNeRF + $\text{Ours}_\text{ft}$ &  & 0.717 & 0.693 & 0.747 & 0.839 & 0.715 & 0.808 & 0.793\\
$\text{ENeRF}_\text{ft}$~\cite{lin2022efficient} &  & 0.847 & 0.908 & 0.905 &	0.898 & \textbf{0.892} & 0.917 & 0.791\\
ENeRF + $\text{Ours}_\text{ft}$ &  & \textbf{0.872} & \textbf{0.919} & \textbf{0.916} & \textbf{0.917} & 0.891 & \textbf{0.927} & \textbf{0.813} \\
\midrule
 & & \multicolumn{7}{c}{LPIPS $\downarrow$} \\
\midrule
MVSNeRF~\cite{chen2021mvsnerf} & \multirow{4}{*}{\makecell{No per-scene\\optimization}} & 0.458 & 0.389 & 0.532 & 0.474 & 0.438 & 0.462 & 0.528\\
MVSNeRF + Ours &  &  0.458 & 0.397 & 0.543 & 0.465 & 0.431 & 0.471 & 0.528\\
ENeRF~\cite{lin2022efficient} &  & 	0.232 & 0.195 & 0.216 & 0.218 & \textbf{0.218} & \textbf{0.178} & 0.317\\
ENeRF + Ours &  & \textbf{0.227} & \textbf{0.194} & \textbf{0.195} & \textbf{0.194} & 0.223 & 0.186 & \textbf{0.307}\\
\midrule
F2-NeRF~\cite{wang2023f2} & \multirow{6}{*}{\makecell{Per-scene\\optimization}} & 	0.283 & 0.262 & 0.233 & 0.270 & 0.237 & 0.215 & 0.448\\
Zip-NeRF~\cite{barron2023zip} & & 0.185 & 0.163 & 0.235 & 0.156 & \textbf{0.166} & 0.167 & 0.613\\
$\text{MVSNeRF}_\text{ft}$~\cite{chen2021mvsnerf} &  & 	0.434 & 0.383 & 0.474 & 0.421 & 0.391 & 0.418 & 0.451\\
MVSNeRF + $\text{Ours}_\text{ft}$ &  & 	0.277 & 0.259 & 0.344 & 0.210 & 0.263 & 0.268 & \textbf{0.233}\\
$\text{ENeRF}_\text{ft}$~\cite{lin2022efficient} &  & 0.190 & 0.158 & 0.160 & 0.175 & 0.179 & 0.142 & 0.258\\
ENeRF + $\text{Ours}_\text{ft}$ &  & \textbf{0.177} & \textbf{0.148} & \textbf{0.148} & \textbf{0.146} & 0.205 & \textbf{0.134} & 0.241\\
\bottomrule
\end{tabular}%
\end{table*}

\subsection{ScanNet dataset}
We show all 8 scenes of the quantitative comparisons on the ScanNet~\cite{dai2017scannet} dataset following the train/test split defined in NeRFusion, NerfingMVS, and SurfelNeRF in Table~\ref{tab:complete_quantitative_scannet_original_split}.

\begin{table*}[t]
\centering
\caption{\textbf{Complete quantitative comparisons with state-of-the-art methods on the ScanNet~\cite{dai2017scannet} dataset, following the train/test split on NeRFusion, NerfingMVS, and SurfelNeRF.}}
\label{tab:complete_quantitative_scannet_original_split}
\begin{tabular}{lccccccccc}
\toprule
Method & Setting & 0000 & 0079 & 0158 & 0316 & 0521 & 0553 & 0616 & 0653 \\
\midrule
 & & \multicolumn{8}{c}{PSNR $\uparrow$} \\
\midrule
MVSNeRF~\cite{chen2021mvsnerf} & \multirow{5}{*}{\makecell{No per-scene\\optimization}} & 23.56 & 28.98 & 25.96 & 19.48 & 20.69 & 27.99 & 16.19 & 24.38 \\
MVSNeRF + Ours &  & 24.02 & 29.10 & 25.24 & 18.73 & 20.59 & 28.82 & 18.02 & 24.75 \\
SurfelNeRF & & 16.57 & 20.33 & 20.43 & 20.42 & 20.89 & 19.88 & 18.58 & 17.12\\
ENeRF~\cite{lin2022efficient} &  & \textbf{29.33} & \textbf{33.49} & \textbf{34.03} & \textbf{32.72} & \textbf{32.12} & \textbf{34.61} & 24.41 & 33.11 \\
ENeRF + Ours &  & 28.89	& 33.27	& 32.31	& 31.14	& 30.59	& 32.96	& \textbf{25.65} & \textbf{33.27} \\
\midrule
F2-NeRF~\cite{wang2023f2} & \multirow{7}{*}{\makecell{Per-scene\\optimization}} & 28.03 & 30.28 & 31.75 & 27.21 & 24.84 & 31.55 & 23.24 & 28.02 \\
Zip-NeRF~\cite{barron2023zip} & & 31.56 & 32.53 & 34.86 & \textbf{34.42} & 31.85 & 34.57 & 24.80 & 33.34 \\
$\text{MVSNeRF}_\text{ft}$~\cite{chen2021mvsnerf} &  & 25.24 & 28.02 & 25.72 & 24.64 & 22.35 & 30.12 & 15.76 & 25.69 \\
MVSNeRF + $\text{Ours}_\text{ft}$ &  & 23.47 & 28.42 & 24.62 & 24.71 & 22.49 & 28.91 & 16.86 & 26.39 \\
$\text{SurfelNeRF}_\text{ft}$ & & 19.85	& 20.84	& 21.21	& 20.82	& 20.61	& 21.95	& 17.47	& 17.55 \\
$\text{ENeRF}_\text{ft}$~\cite{lin2022efficient} &  & 31.52	& \textbf{33.39}	& 35.82	& 33.33	& \textbf{32.32}	& \textbf{35.67} & 25.36 & 34.19 \\
ENeRF + $\text{Ours}_\text{ft}$ &  & \textbf{31.86} & 33.26 & \textbf{36.09} & 33.43 & 32.29 & 35.18 & \textbf{26.25} & \textbf{34.57} \\
\midrule
 & & \multicolumn{8}{c}{SSIM $\uparrow$} \\
\midrule
MVSNeRF~\cite{chen2021mvsnerf} & \multirow{5}{*}{\makecell{No per-scene\\optimization}} &  0.831 & 0.909 & 0.905 & 0.873 & 0.833 & 0.936 & 0.713 & 0.899\\
MVSNeRF + Ours &  & 0.838 & 0.911 & 0.904 & 0.880 & 0.847 & 0.940 & 0.757 & 0.899 \\
SurfelNeRF & & 0.444 & 0.644 & 0.712 & 0.699 & 0.688 & 0.660 & 0.508 & 0.631\\
ENeRF~\cite{lin2022efficient} &  & 0.938 & 0.952 & \textbf{0.974} & \textbf{0.977} & \textbf{0.958} & \textbf{0.978} & 0.890	& 0.974 \\
ENeRF + Ours &  & \textbf{0.943}	& \textbf{0.953}	& 0.973	& 0.975	& 0.955	& 0.976	& \textbf{0.902}	& \textbf{0.975} \\
\midrule
F2-NeRF~\cite{wang2023f2} & \multirow{7}{*}{\makecell{Per-scene\\optimization}} & 0.865 & 0.871 & 0.935 & 0.939 & 0.894 & 0.942 & 0.788 & 0.919 \\
Zip-NeRF~\cite{barron2023zip} & & 0.908 & 0.893 & 0.953 & 0.961 & 0.920 & 0.956 & 0.794 & 0.951 \\
$\text{MVSNeRF}_\text{ft}$~\cite{chen2021mvsnerf} &  & 0.856 & 0.894 & 0.901 & 0.917 & 0.857 & 0.948 & 0.700 & 0.899 \\
MVSNeRF + $\text{Ours}_\text{ft}$ &  & 0.841 & 0.902 & 0.893 & 0.917 & 0.859 & 0.941 & 0.727 & 0.907 \\
$\text{SurfelNeRF}_\text{ft}$ & & 0.547	& 0.664	& 0.728	& 0.714	& 0.659	& 0.750	& 0.482	& 0.677 \\
$\text{ENeRF}_\text{ft}$~\cite{lin2022efficient} &  & \textbf{0.949}	& \textbf{0.955}	& \textbf{0.979}	& \textbf{0.982}	& \textbf{0.958} & \textbf{0.981} & 0.896 & \textbf{0.976} \\
ENeRF + $\text{Ours}_\text{ft}$ &  & 0.943 & 0.953 & 0.973 & 0.975 & 0.955 & 0.976 & \textbf{0.902} & 0.975 \\
\midrule
 & & \multicolumn{8}{c}{LPIPS $\downarrow$} \\
\midrule
MVSNeRF~\cite{chen2021mvsnerf} & \multirow{5}{*}{\makecell{No per-scene\\optimization}} & 0.362	& 0.301	& 0.305	& 0.370 & 0.493 & 0.225 & 0.577 & 0.306\\
MVSNeRF + Ours &  & 0.366 & 0.333 & 0.327 & 0.362 & 0.465 & 0.240 & 0.551 & 0.274 \\
SurfelNeRF & & 0.575	& 0.541	& 0.499	& 0.477	& 0.508	& 0.514	& 0.573	& 0.534 \\
ENeRF~\cite{lin2022efficient} &  & 0.195	& \textbf{0.211}	& \textbf{0.193}	& \textbf{0.219}	& \textbf{0.219}	& \textbf{0.171}	& \textbf{0.281} & 	\textbf{0.161} \\
ENeRF + Ours &  & \textbf{0.194}	& 0.216	& 0.208	& 0.244	& 0.254	& 0.188	& 0.282	& 0.169 \\
\midrule
F2-NeRF~\cite{wang2023f2} & \multirow{7}{*}{\makecell{Per-scene\\optimization}} & 0.230 & 0.255 & 0.195 & 0.214 & 0.257 & 0.166 & 0.350 & 0.177 \\
Zip-NeRF~\cite{barron2023zip} & & 0.190 & 0.243 & 0.181 & 0.191 & 0.271 & 0.156 & 0.325 & 0.158 \\
$\text{MVSNeRF}_\text{ft}$~\cite{chen2021mvsnerf} &  & 0.288 & 0.274 & 0.287 & 0.276 & 0.426 & 0.183 & 0.568 & 0.229 \\
MVSNeRF + $\text{Ours}_\text{ft}$ &  & 0.284 & 0.270	& 0.285	& 0.280	& 0.426	& 0.189	& 0.544	& 0.223 \\
$\text{SurfelNeRF}_\text{ft}$ & & 0.532	& 0.502	& 0.424	& 0.487	& 0.532 & 0.458	& 0.595	& 0.498\\
$\text{ENeRF}_\text{ft}$~\cite{lin2022efficient} &  & 0.162	& \textbf{0.189}	& 0.144	& \textbf{0.154}	& \textbf{0.209}	& \textbf{0.128} & 0.271 & 0.133 \\
ENeRF + $\text{Ours}_\text{ft}$ &  & \textbf{0.160} & 0.191 & \textbf{0.142} & \textbf{0.154} & 0.212 & 0.130 & \textbf{0.266} & \textbf{0.132}\\
\bottomrule
\end{tabular}%
\end{table*}

\section{Additional Qualitative Comparisons} \label{sec:additional_exps_visual}

\subsection{Free dataset}
\begin{figure*}[t]
\centering
\small
\setlength{\tabcolsep}{1pt}
\renewcommand{\arraystretch}{1}
\resizebox{1.0\textwidth}{!} 
{
\begin{tabular}{c:cc:cccc}
\includegraphics[width=0.143\textwidth]{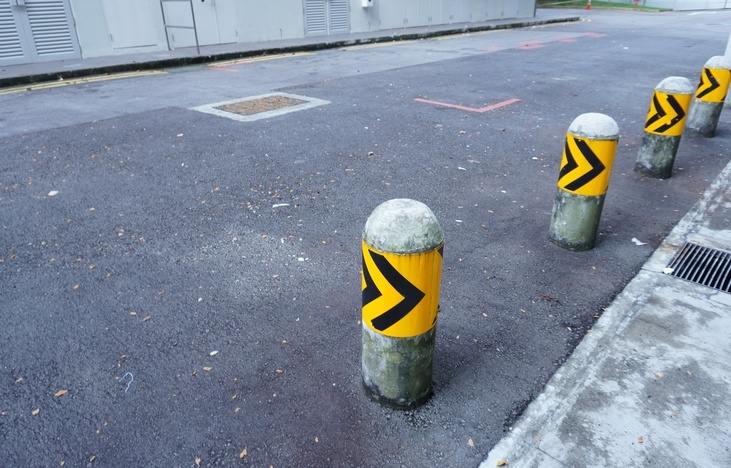} & 
\includegraphics[width=0.143\textwidth]{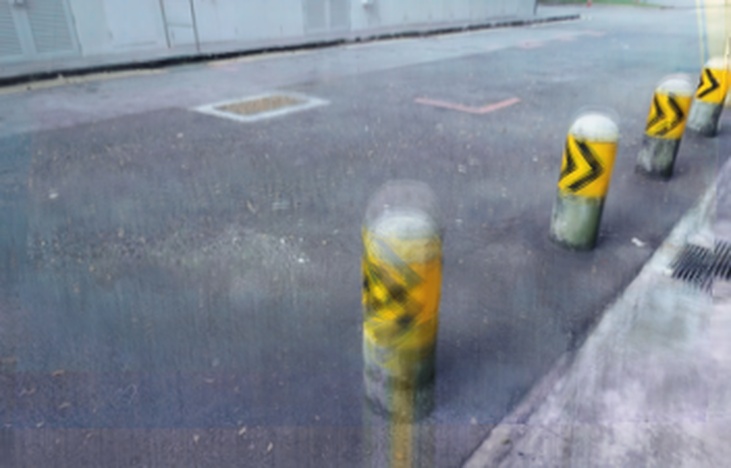} & 
\includegraphics[width=0.143\textwidth]{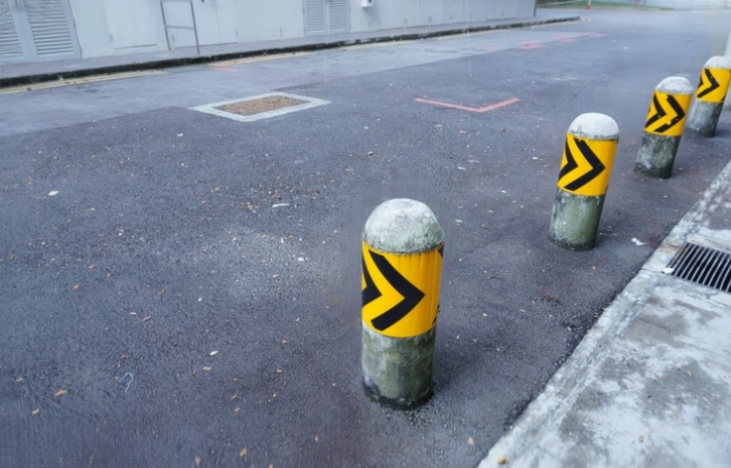} & 
\includegraphics[width=0.143\textwidth]{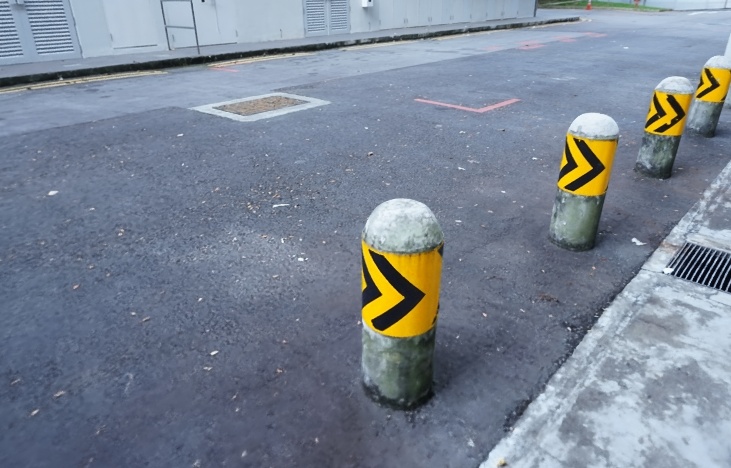} & 
\includegraphics[width=0.143\textwidth]{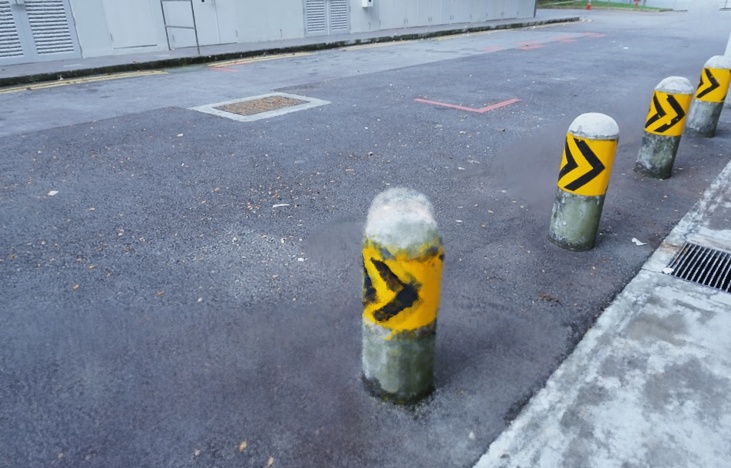} & 
\includegraphics[width=0.143\textwidth]{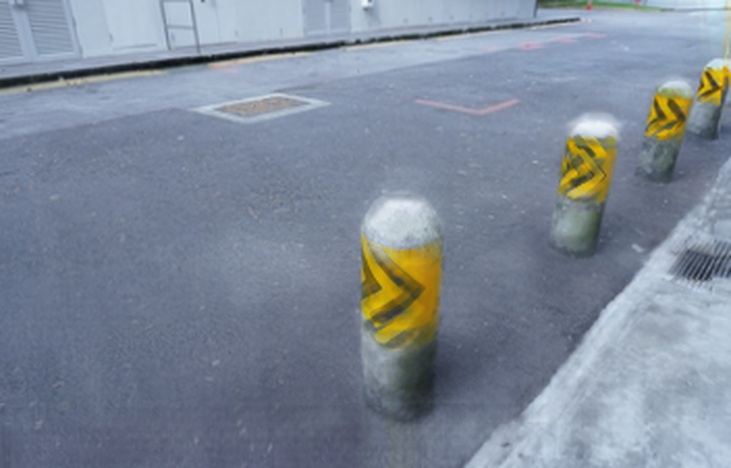} & 
\includegraphics[width=0.143\textwidth]{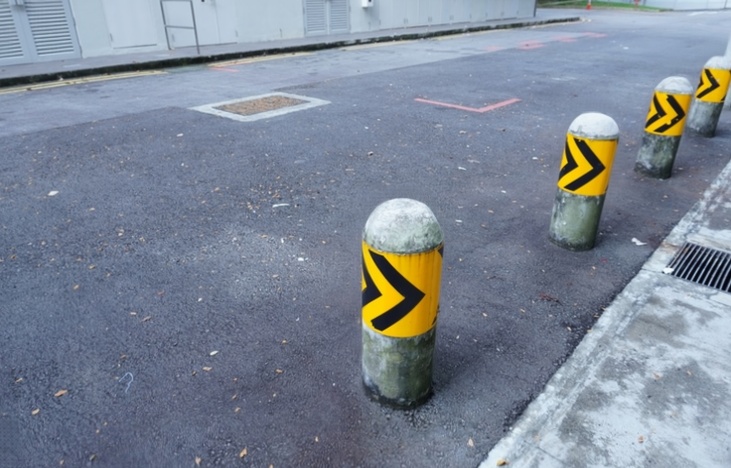} \\
\includegraphics[width=0.143\textwidth]{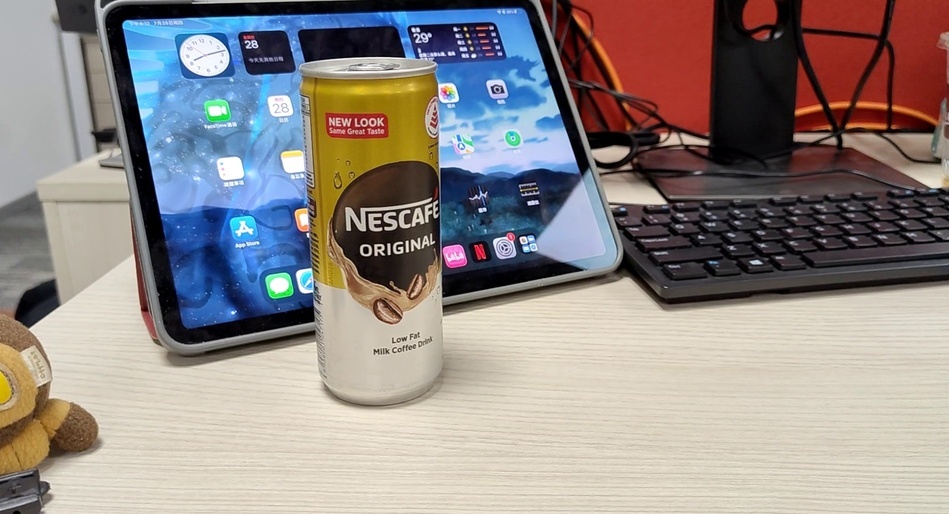} & 
\includegraphics[width=0.143\textwidth]{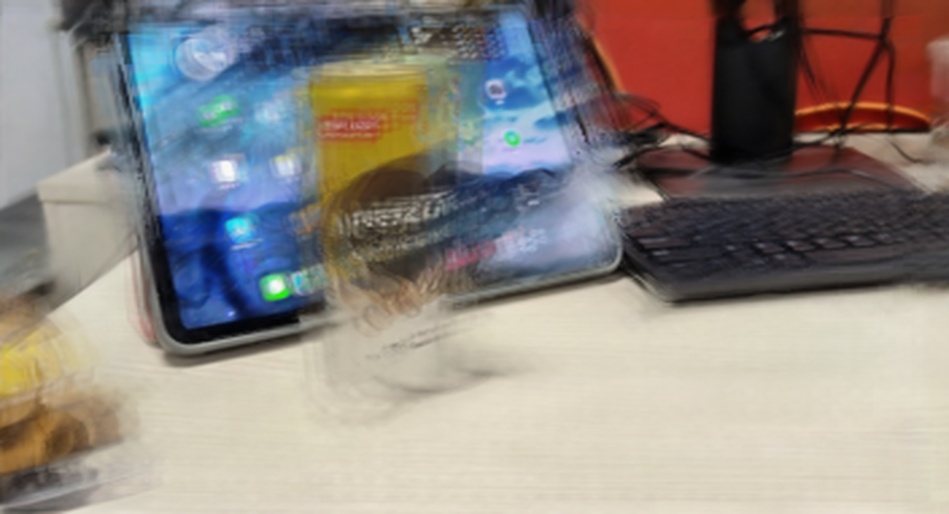} & 
\includegraphics[width=0.143\textwidth]{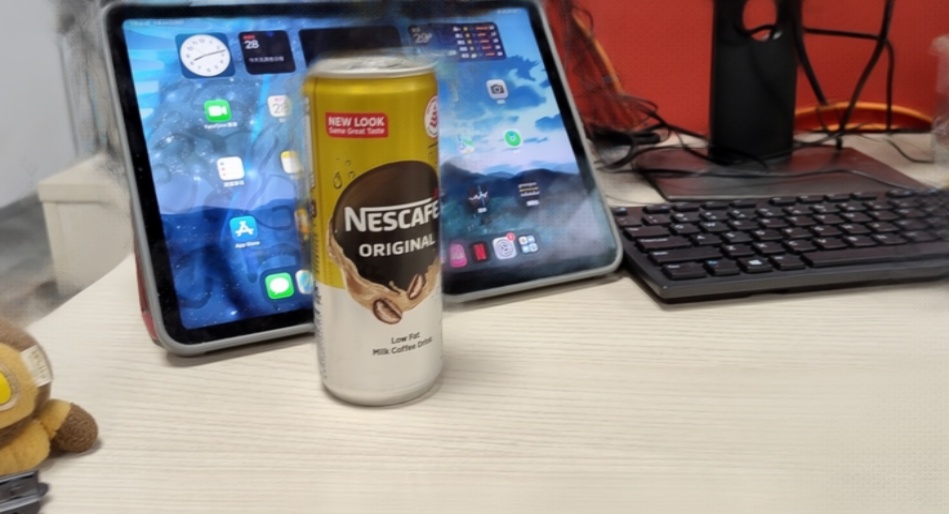} & 
\includegraphics[width=0.143\textwidth]{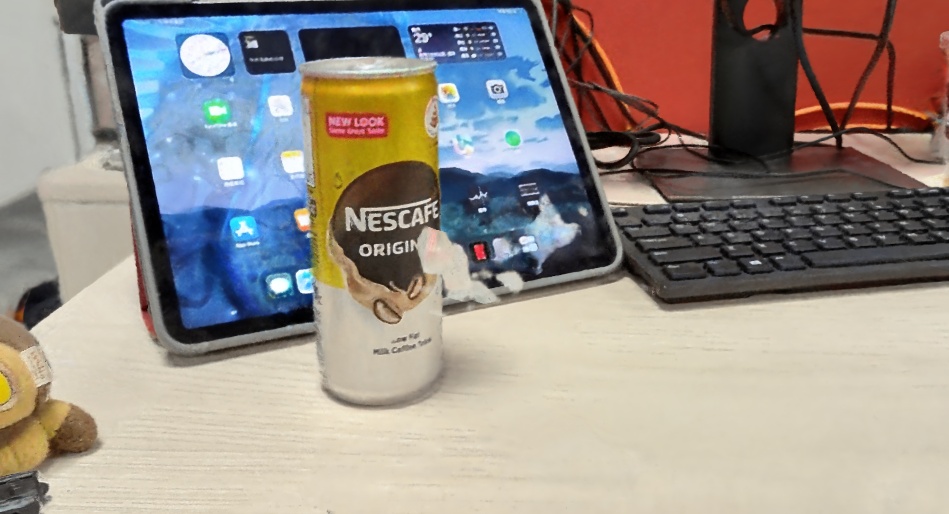} & 
\includegraphics[width=0.143\textwidth]{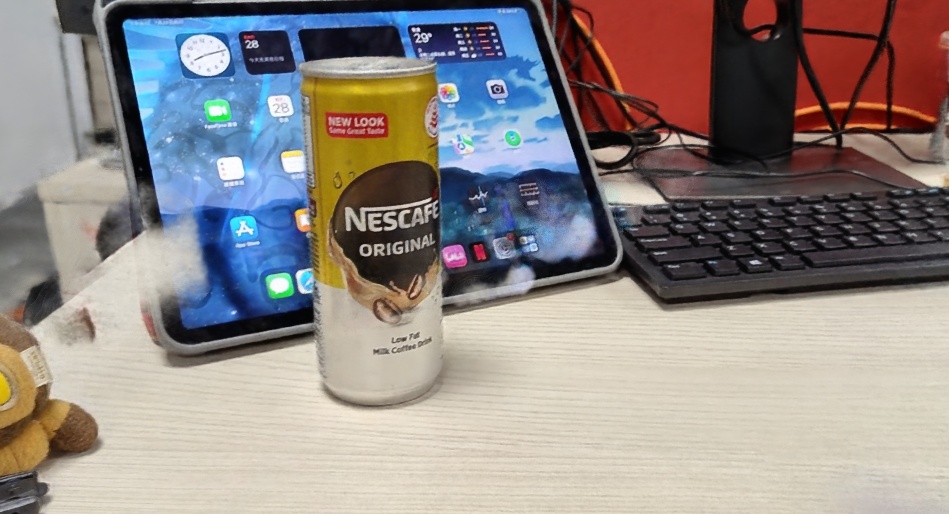} & 
\includegraphics[width=0.143\textwidth]{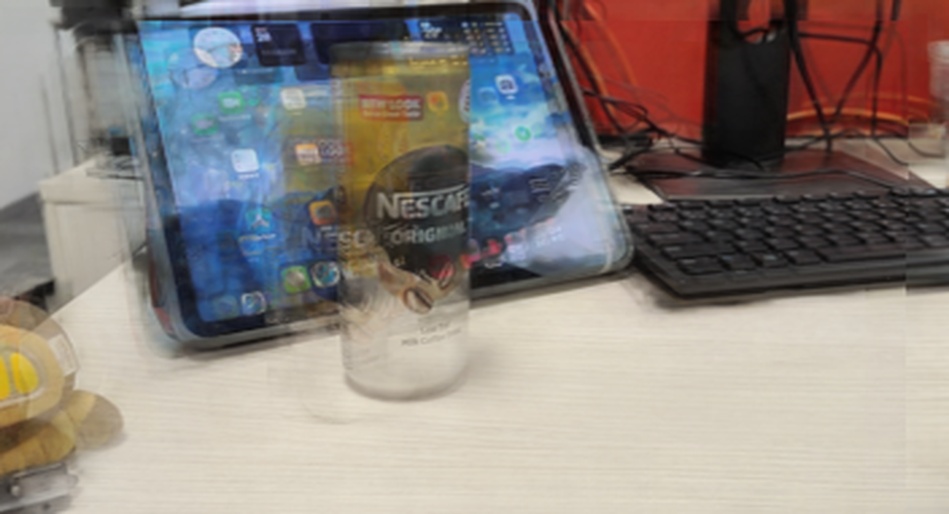} & 
\includegraphics[width=0.143\textwidth]{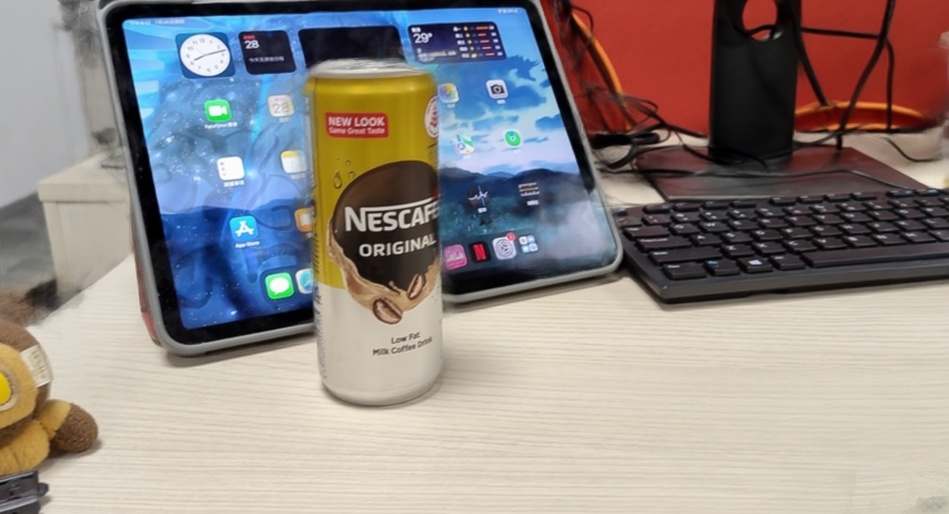} \\
\includegraphics[width=0.143\textwidth]{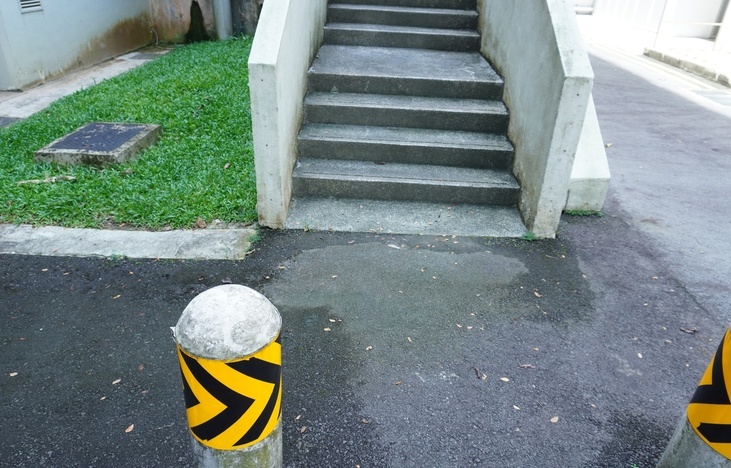} & 
\includegraphics[width=0.143\textwidth]{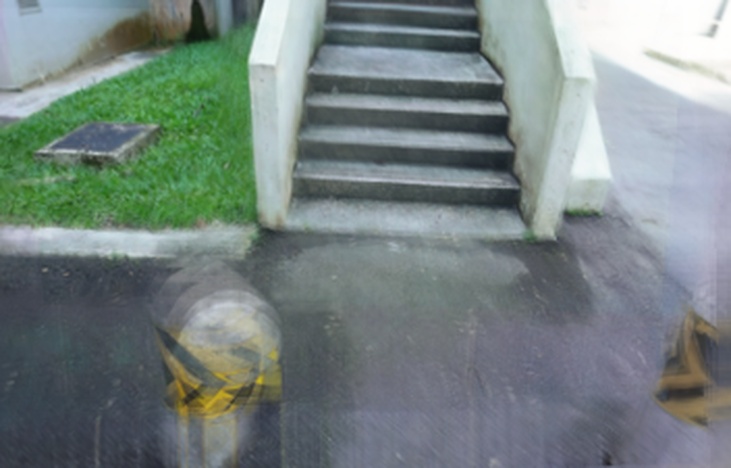} & 
\includegraphics[width=0.143\textwidth]{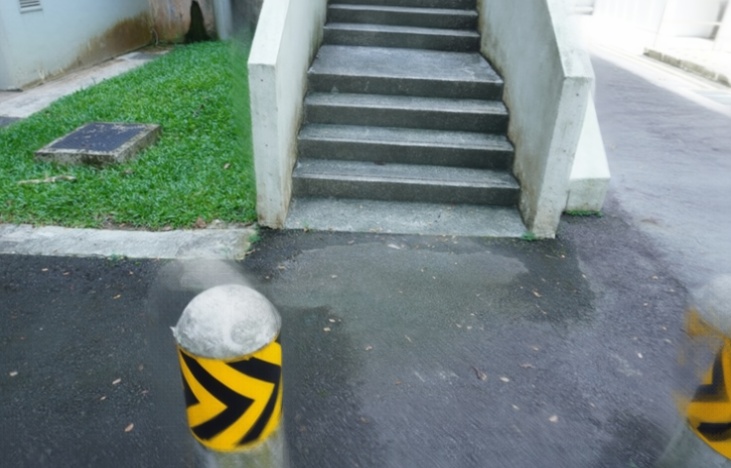} & 
\includegraphics[width=0.143\textwidth]{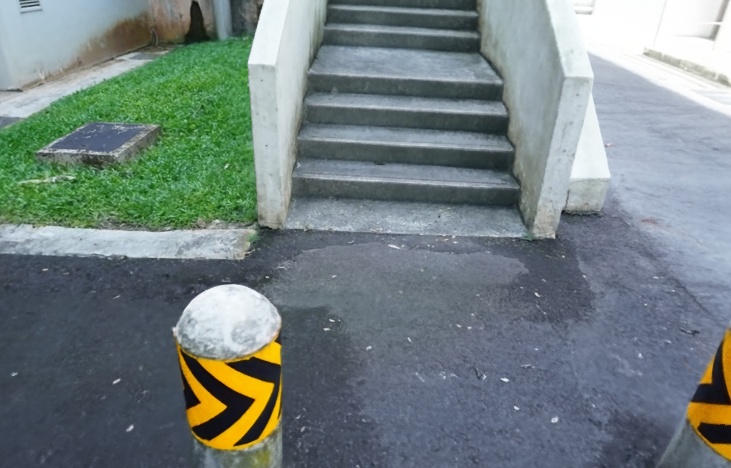} & 
\includegraphics[width=0.143\textwidth]{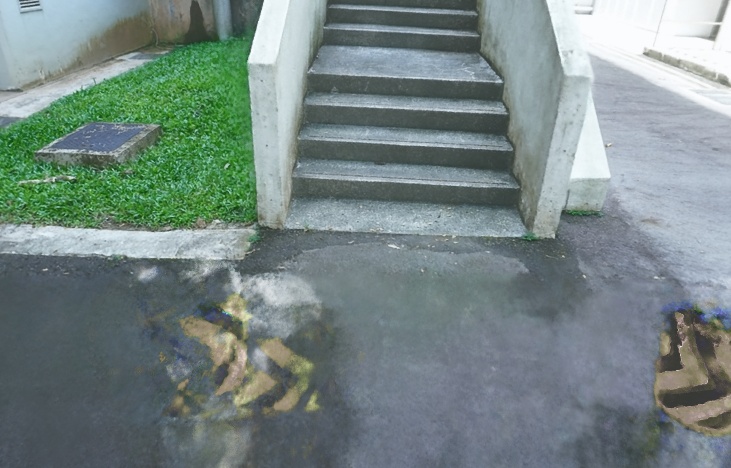} & 
\includegraphics[width=0.143\textwidth]{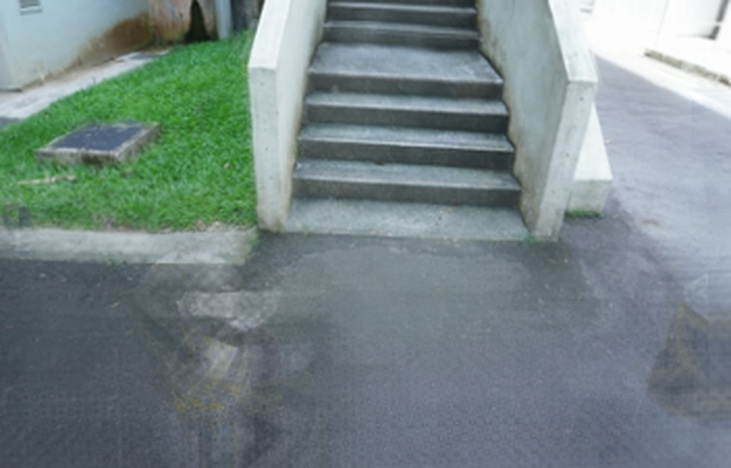} & 
\includegraphics[width=0.143\textwidth]{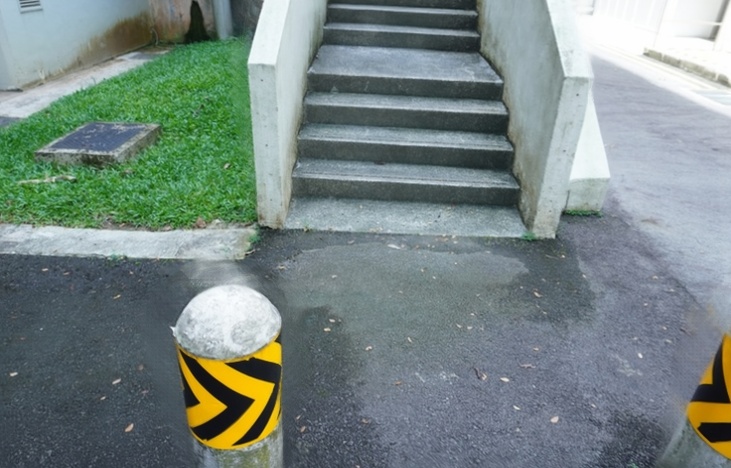} \\
\includegraphics[width=0.143\textwidth]{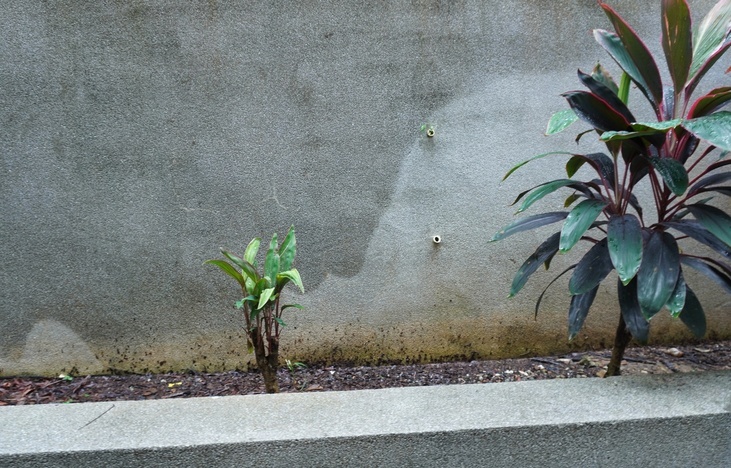} & 
\includegraphics[width=0.143\textwidth]{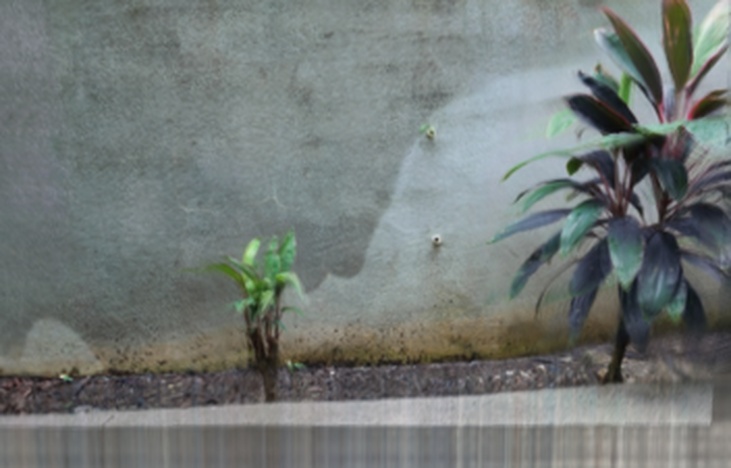} & 
\includegraphics[width=0.143\textwidth]{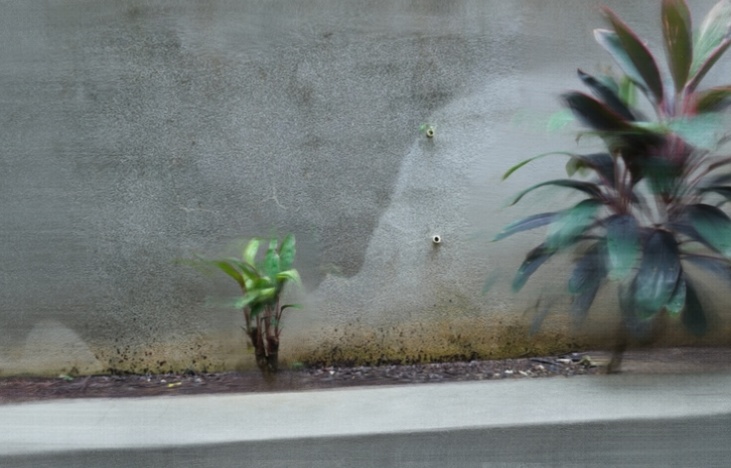} & 
\includegraphics[width=0.143\textwidth]{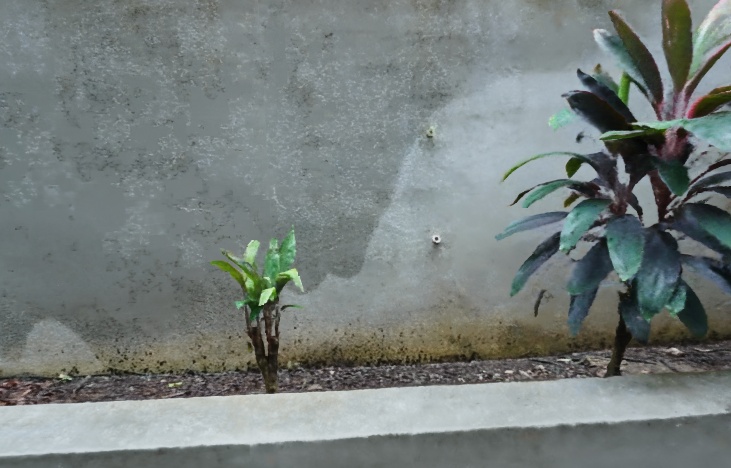} & 
\includegraphics[width=0.143\textwidth]{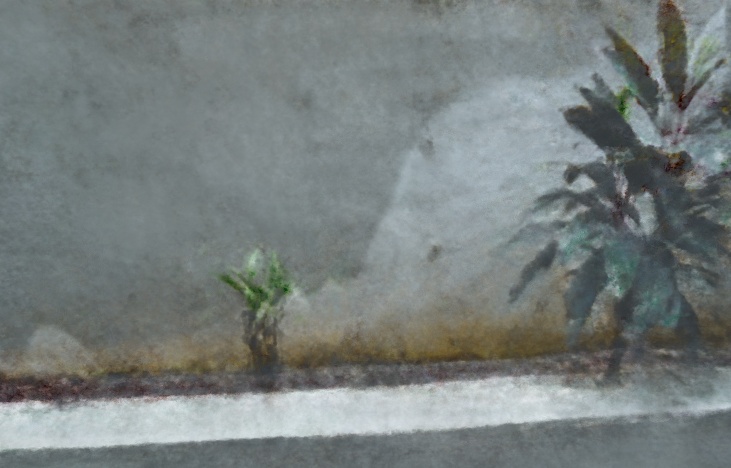} & 
\includegraphics[width=0.143\textwidth]{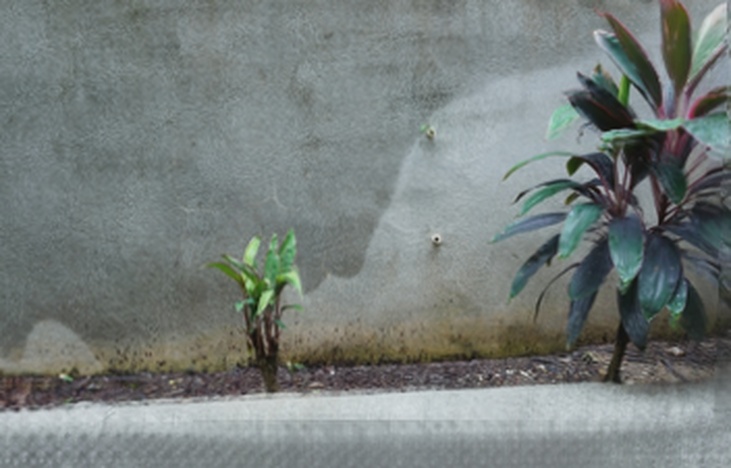} & 
\includegraphics[width=0.143\textwidth]{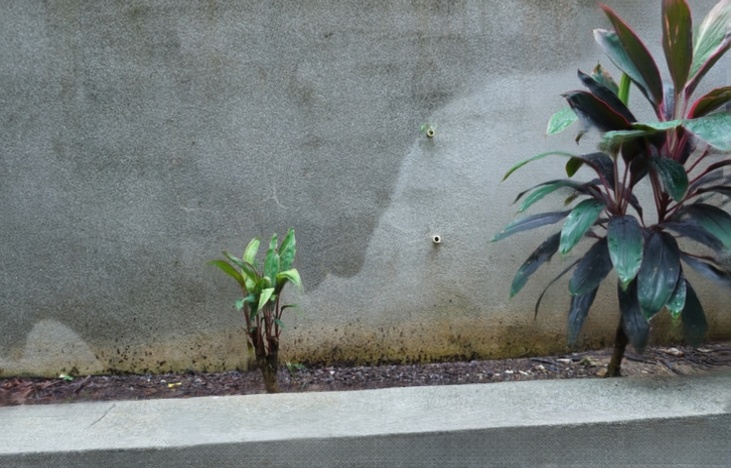} \\
\includegraphics[width=0.143\textwidth]{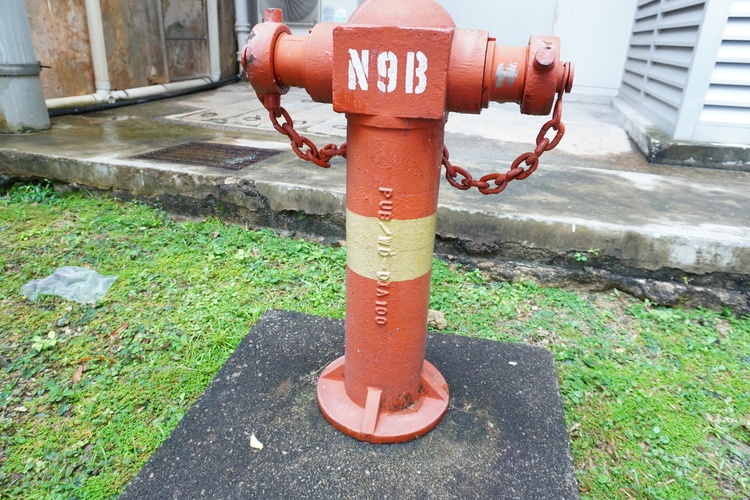} & 
\includegraphics[width=0.143\textwidth]{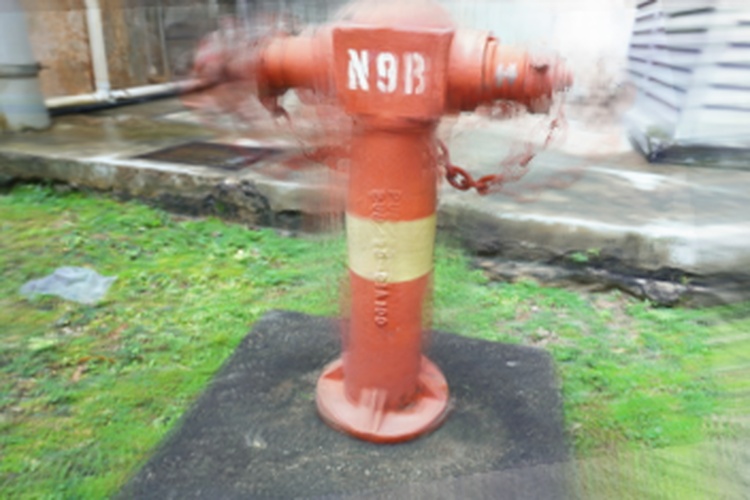} & 
\includegraphics[width=0.143\textwidth]{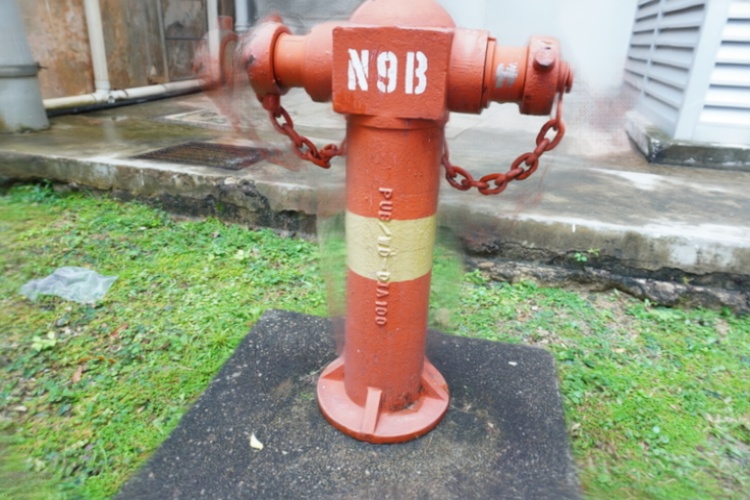} & 
\includegraphics[width=0.143\textwidth]{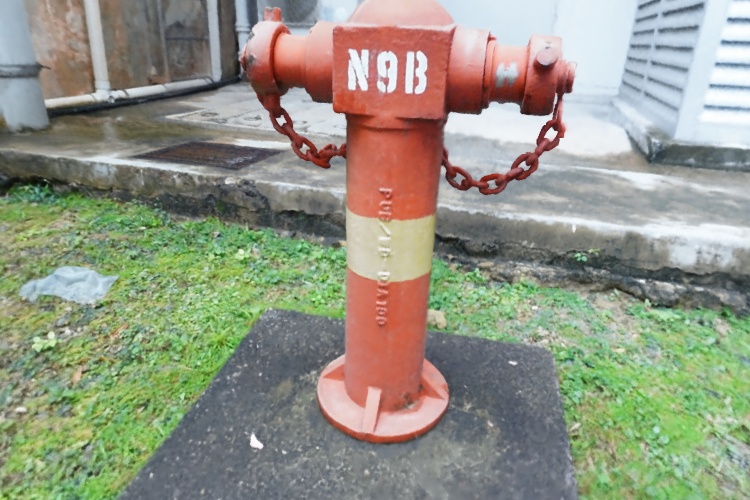} & 
\includegraphics[width=0.143\textwidth]{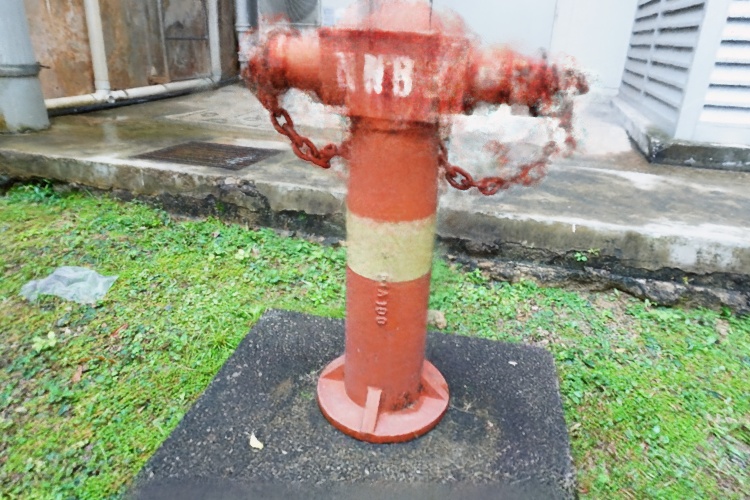} & 
\includegraphics[width=0.143\textwidth]{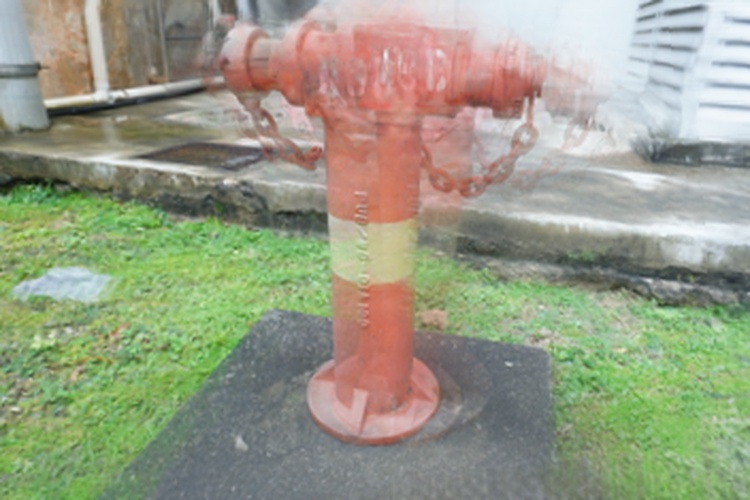} & 
\includegraphics[width=0.143\textwidth]{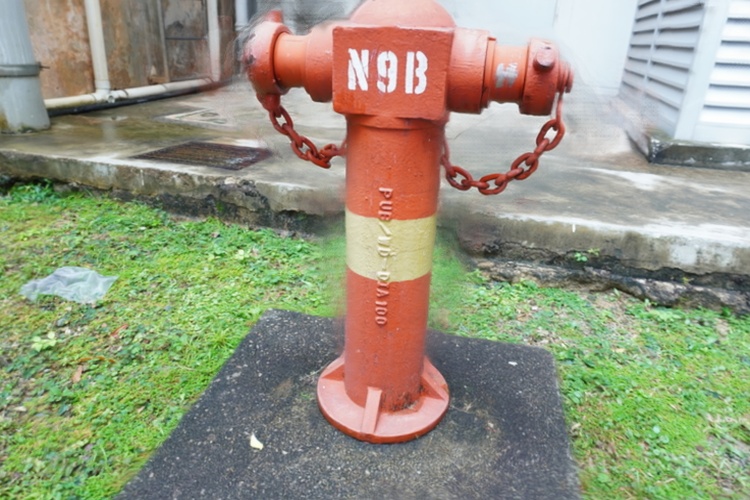} \\
\includegraphics[width=0.143\textwidth]{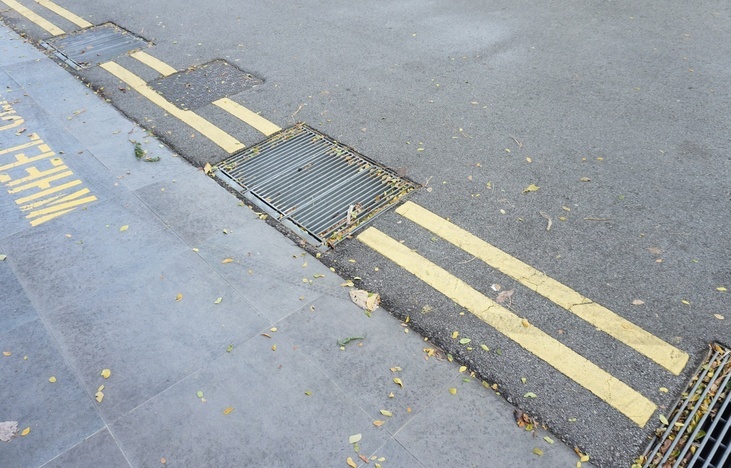} & 
\includegraphics[width=0.143\textwidth]{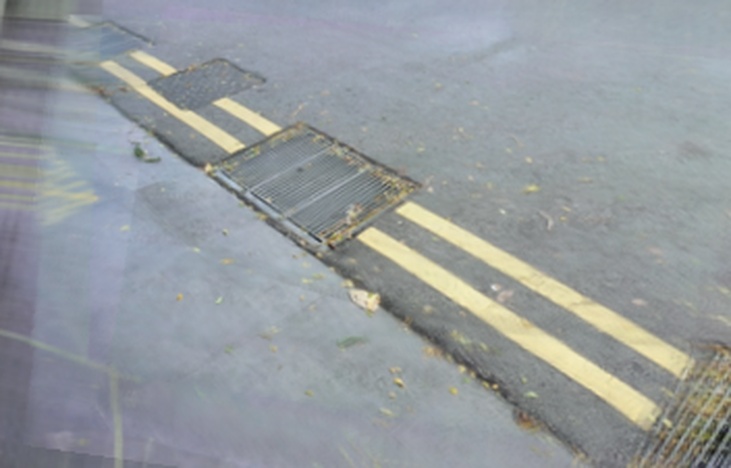} & 
\includegraphics[width=0.143\textwidth]{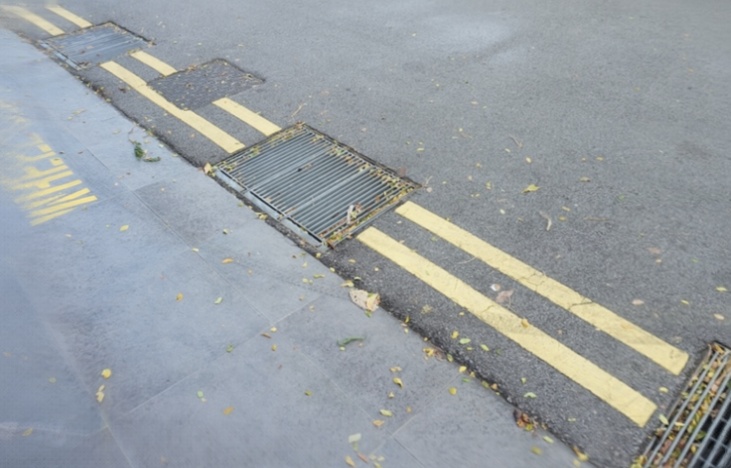} & 
\includegraphics[width=0.143\textwidth]{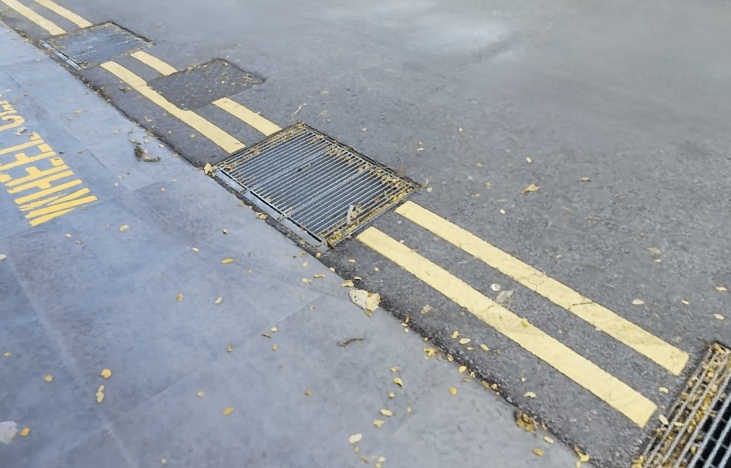} & 
\includegraphics[width=0.143\textwidth]{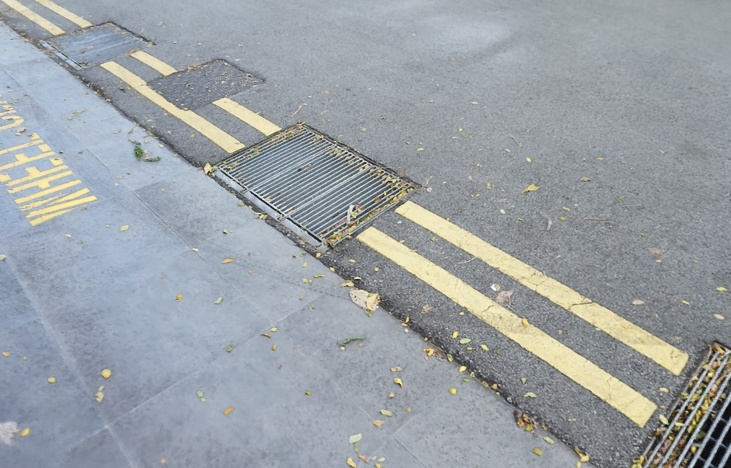} & 
\includegraphics[width=0.143\textwidth]{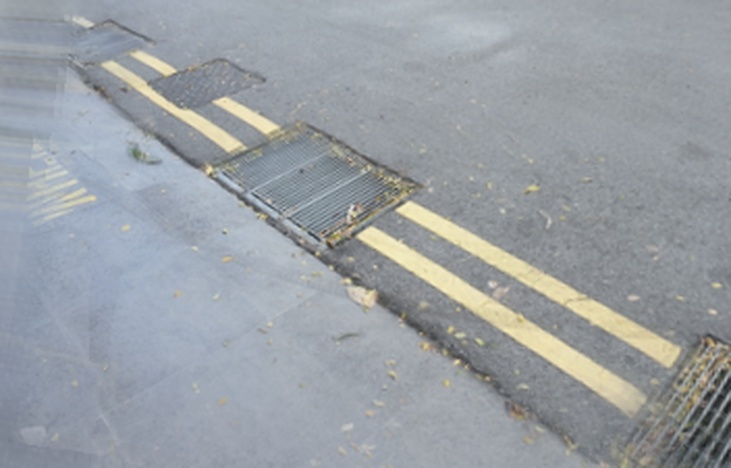} & 
\includegraphics[width=0.143\textwidth]{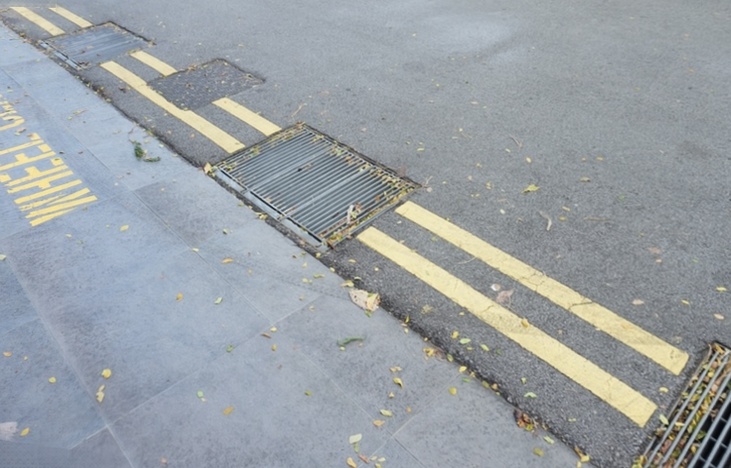} \\
\includegraphics[width=0.143\textwidth]{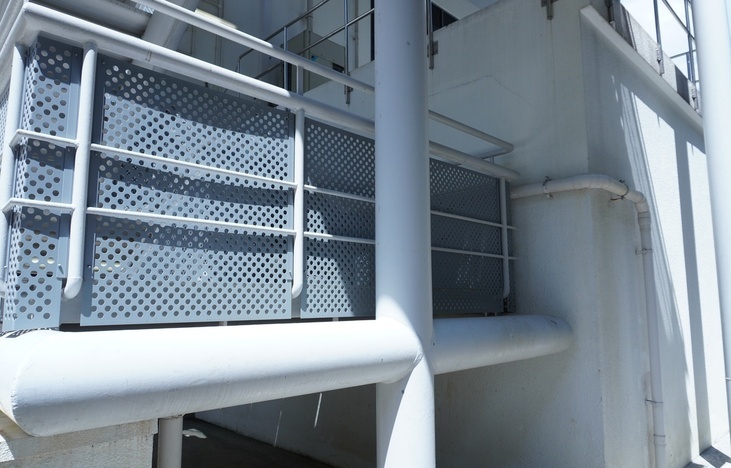} & 
\includegraphics[width=0.143\textwidth]{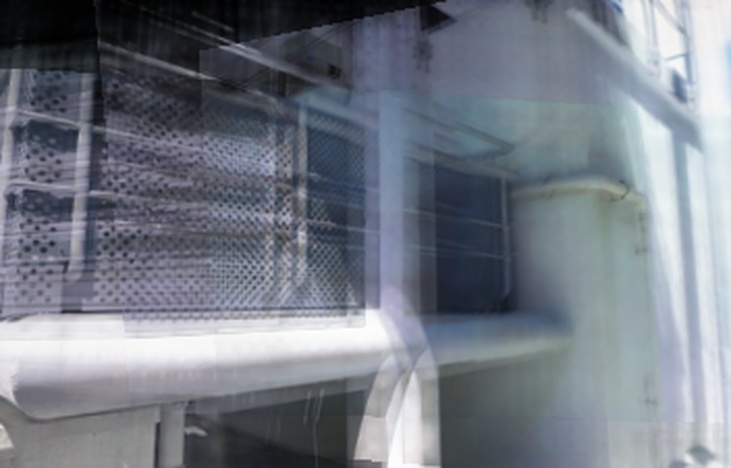} & 
\includegraphics[width=0.143\textwidth]{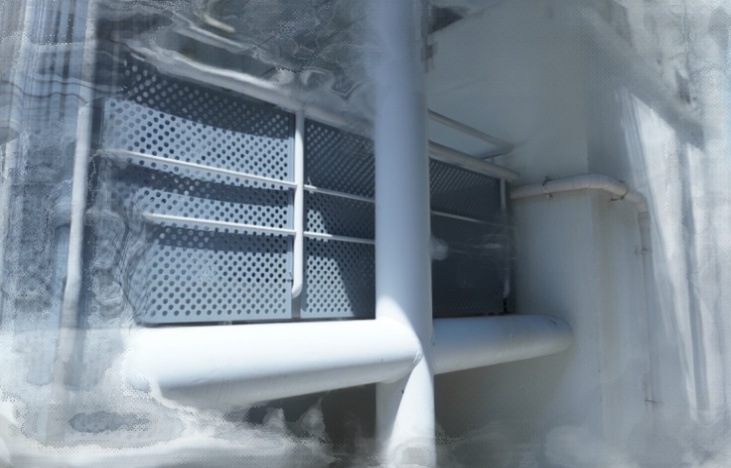} & 
\includegraphics[width=0.143\textwidth]{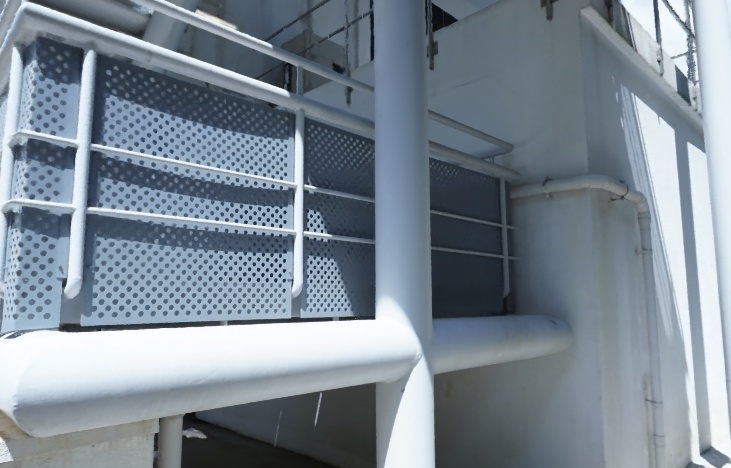} & 
\includegraphics[width=0.143\textwidth]{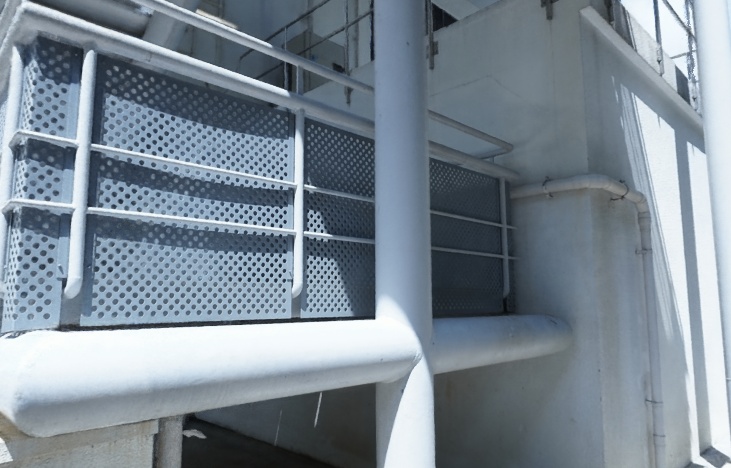} & 
\includegraphics[width=0.143\textwidth]{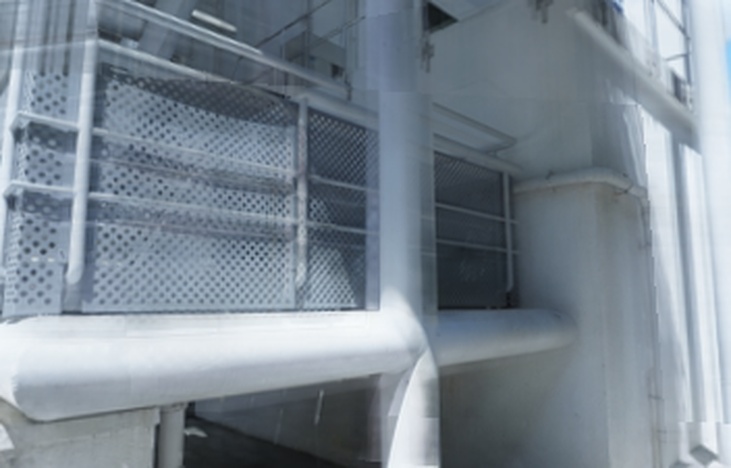} & 
\includegraphics[width=0.143\textwidth]{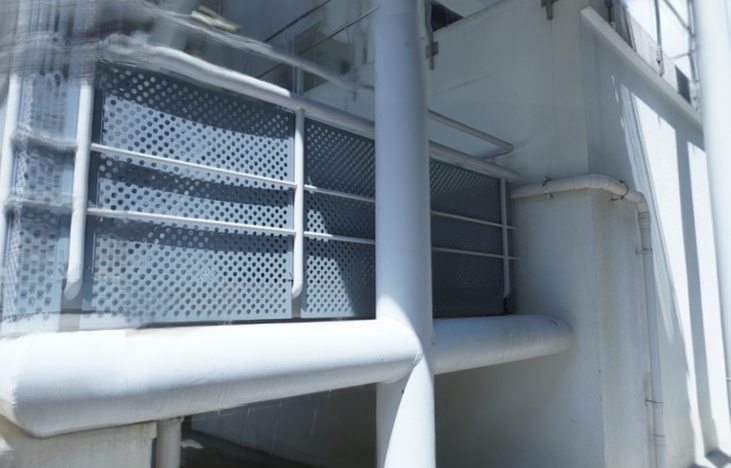} \\
Ground truth & MVSNeRF & Ours & F2-NeRF & Zip-NeRF & MVSNeRF + ft & Ours + ft \\
& \cite{chen2021mvsnerf} & & \cite{wang2023f2} & \cite{barron2023zip} & \cite{chen2021mvsnerf}\\
\end{tabular}%
}
\caption{\textbf{Additional qualitative comparisons of rendering quality on the Free~\cite{wang2023f2} dataset.}}
\label{fig:more_qualitative_free}
\end{figure*}

\begin{figure*}[t]
\centering
\small
\setlength{\tabcolsep}{1pt}
\renewcommand{\arraystretch}{1}
\resizebox{1.0\textwidth}{!} 
{
\begin{tabular}{ccccc}
\includegraphics[width=0.2\textwidth]{figures/results/road_112_0_gt.jpg} & \includegraphics[width=0.2\textwidth]{figures/results/road_112_0_mvsnerf.jpg} & \includegraphics[width=0.2\textwidth]{figures/results/road_112_0_mvsnerf_ft.jpg} & \includegraphics[width=0.2\textwidth]{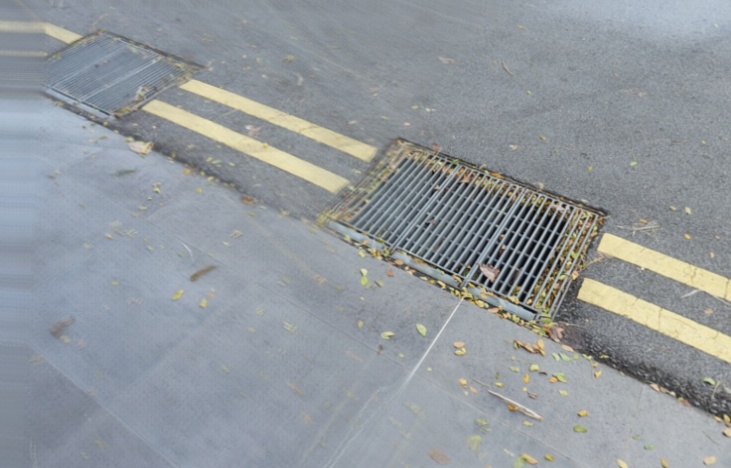} & \includegraphics[width=0.2\textwidth]{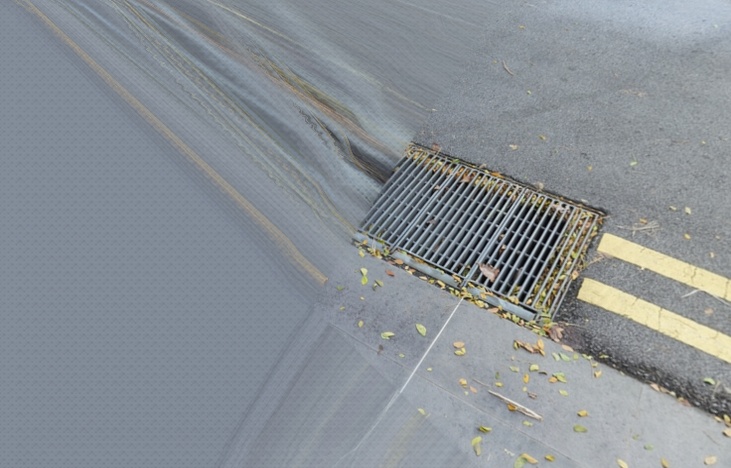} \\
Ground truth & MVSNeRF~\cite{chen2021mvsnerf} & MVSNeRF + ft & ENeRF~\cite{lin2022efficient} & ENeRF + ft \\
 & \includegraphics[width=0.2\textwidth]{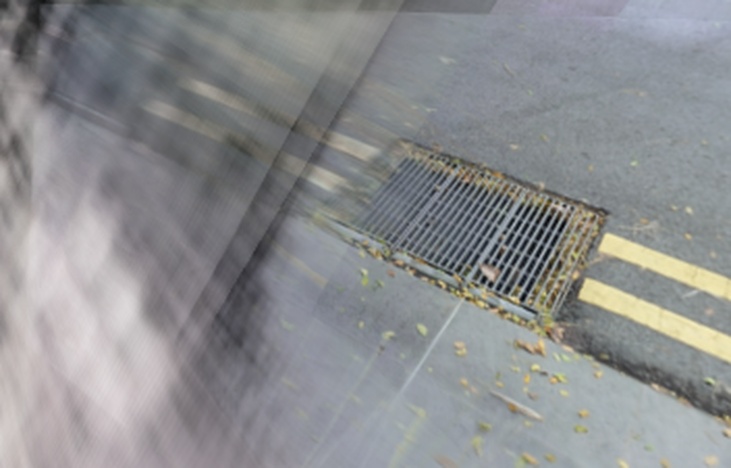} & \includegraphics[width=0.2\textwidth]{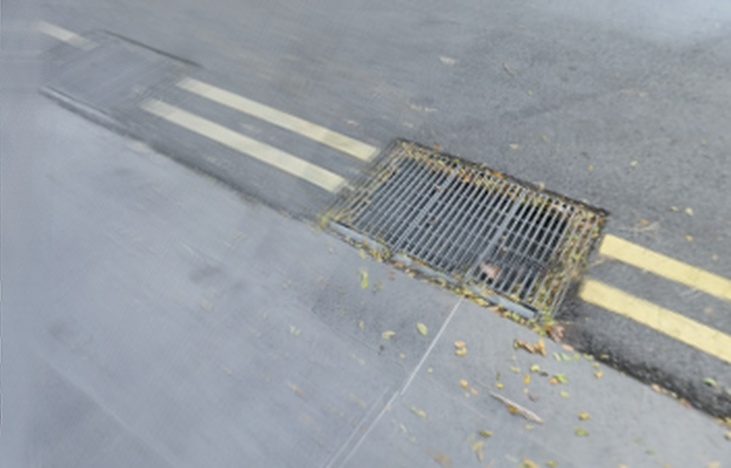} & \includegraphics[width=0.2\textwidth]{figures/results/road_112_0_enerf_ours.jpg} & \includegraphics[width=0.2\textwidth]{figures/results/road_112_0_enerf_ours_ft.jpg}\\
 & MVSNeRF + \textbf{ours} & MVSNeRF + \textbf{ours} + ft & ENeRF + \textbf{ours} & ENeRF + \textbf{ours} + ft \\
\includegraphics[width=0.2\textwidth]{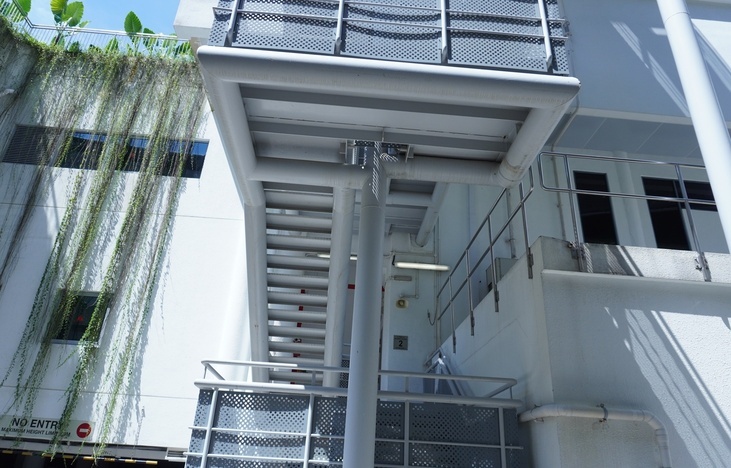} & \includegraphics[width=0.2\textwidth]{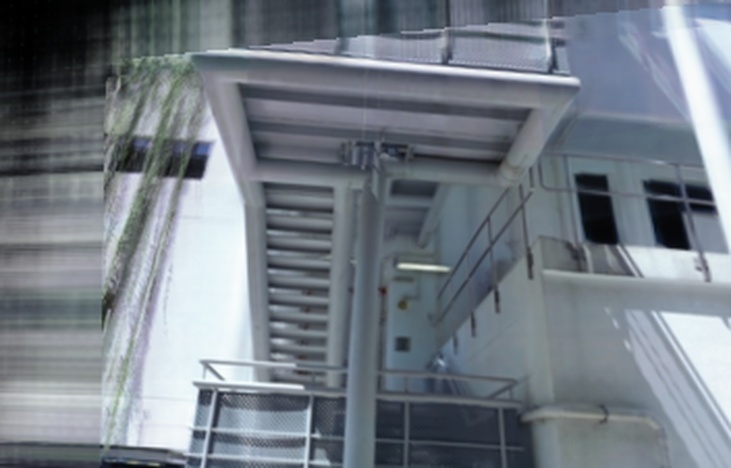} & \includegraphics[width=0.2\textwidth]{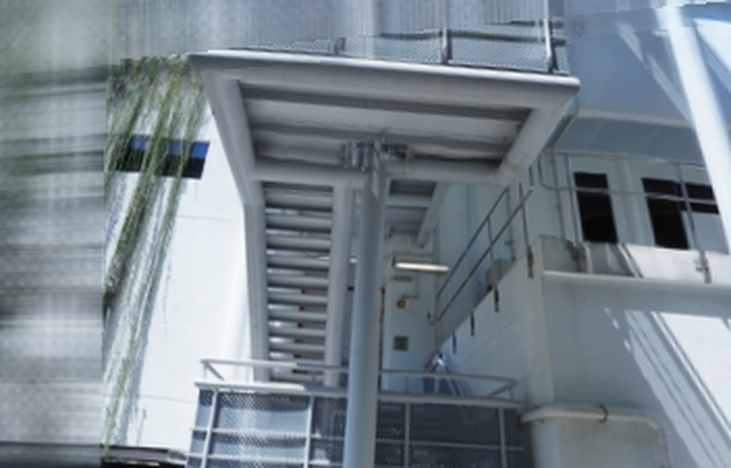} & \includegraphics[width=0.2\textwidth]{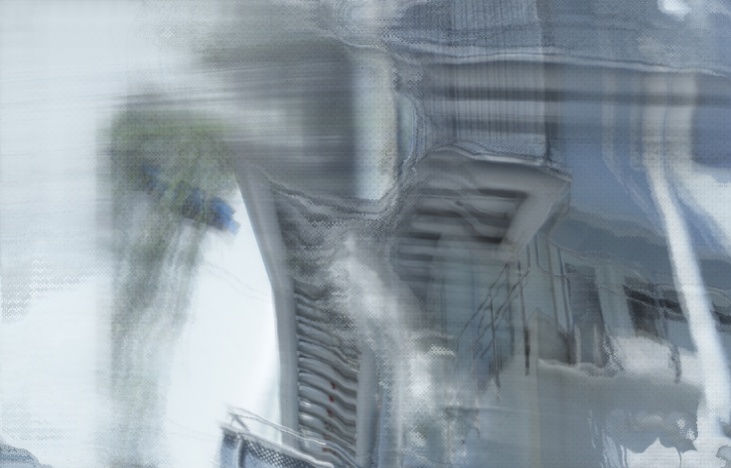} & \includegraphics[width=0.2\textwidth]{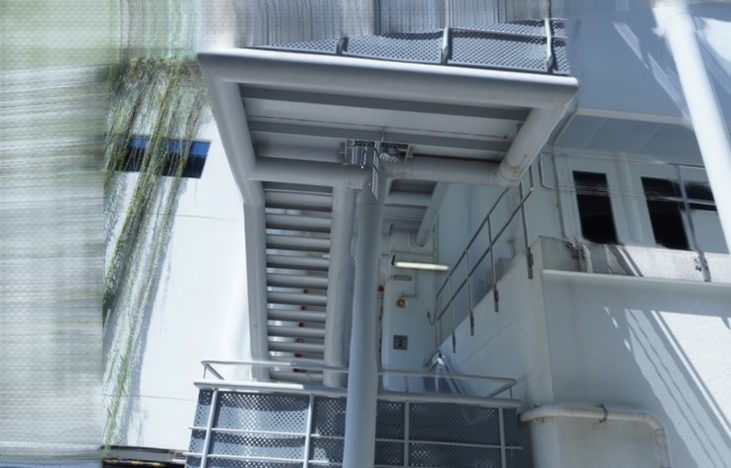} \\
Ground truth & MVSNeRF~\cite{chen2021mvsnerf} & MVSNeRF + ft & ENeRF~\cite{lin2022efficient} & ENeRF + ft \\
 & \includegraphics[width=0.2\textwidth]{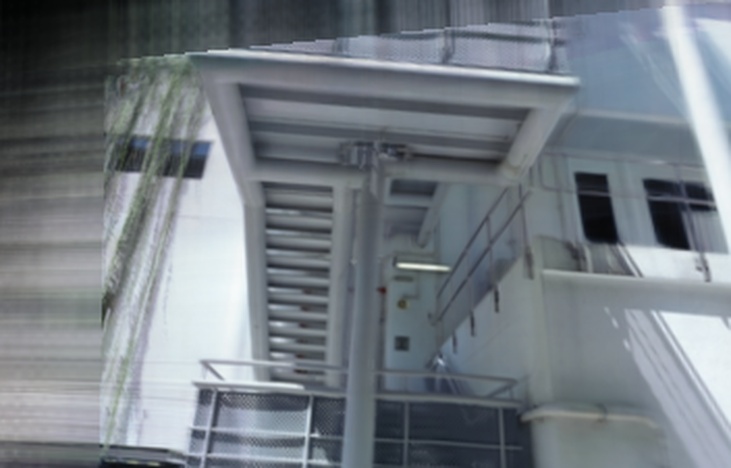} & \includegraphics[width=0.2\textwidth]{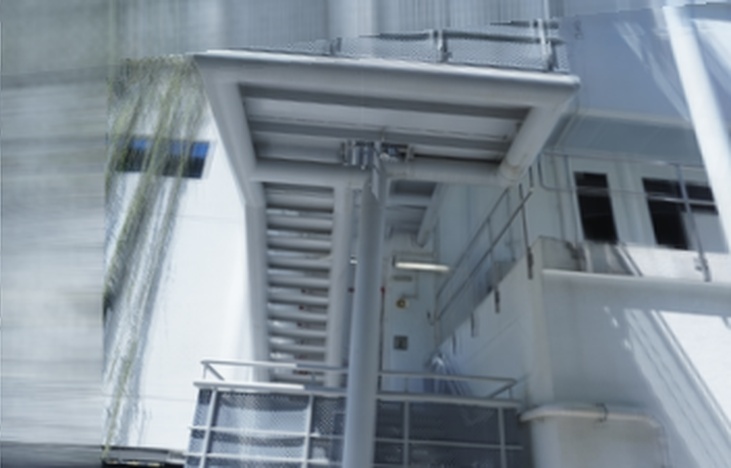} & \includegraphics[width=0.2\textwidth]{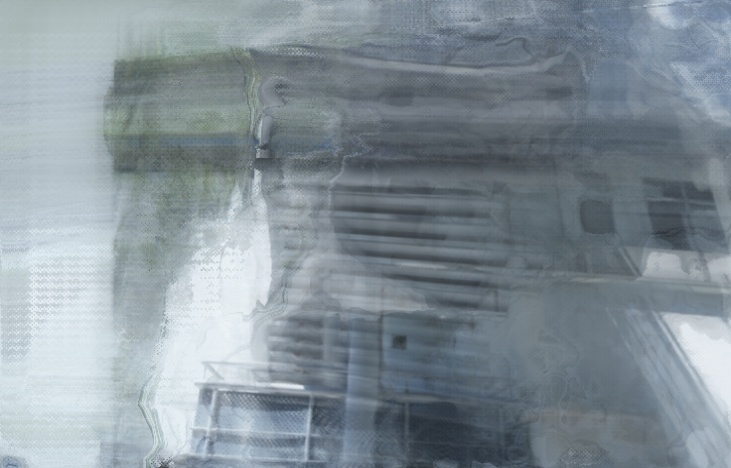} & \includegraphics[width=0.2\textwidth]{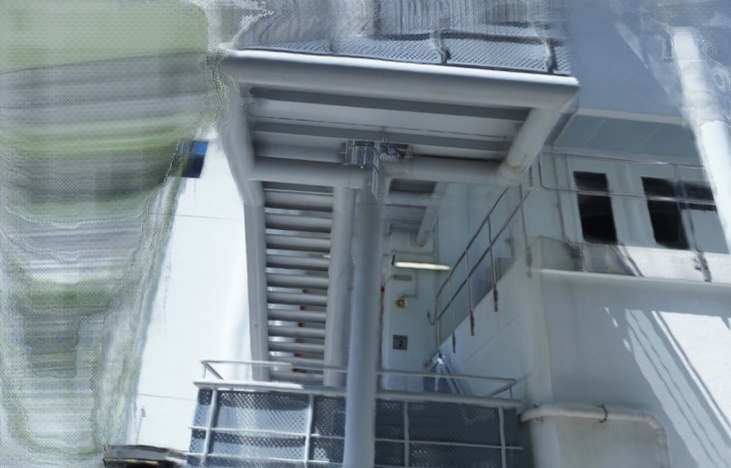}\\
 & MVSNeRF + \textbf{ours} & MVSNeRF + \textbf{ours} + ft & ENeRF + \textbf{ours} & ENeRF + \textbf{ours} + ft \\
\includegraphics[width=0.2\textwidth]{figures/results/lab_72_0_gt.jpg} & \includegraphics[width=0.2\textwidth]{figures/results/lab_72_0_mvsnerf.jpg} & \includegraphics[width=0.2\textwidth]{figures/results/lab_72_0_mvsnerf_ft.jpg} & \includegraphics[width=0.2\textwidth]{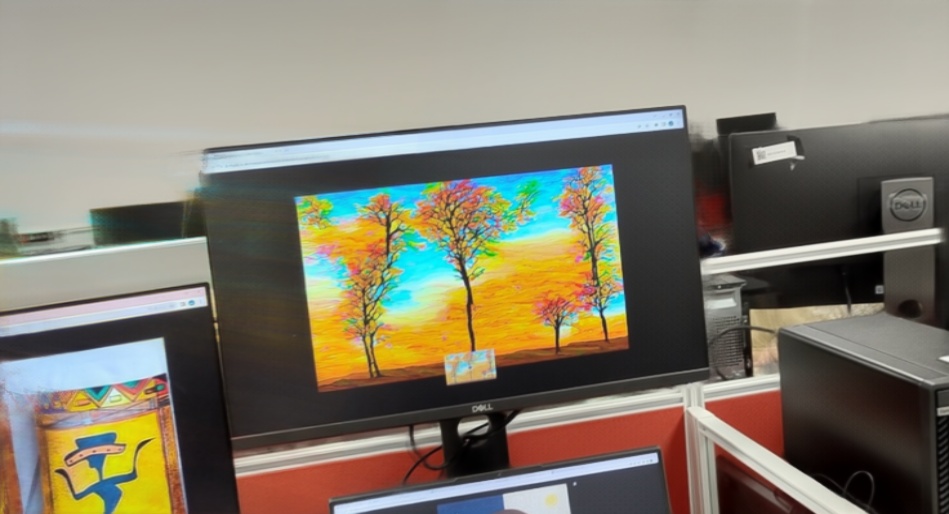} & \includegraphics[width=0.2\textwidth]{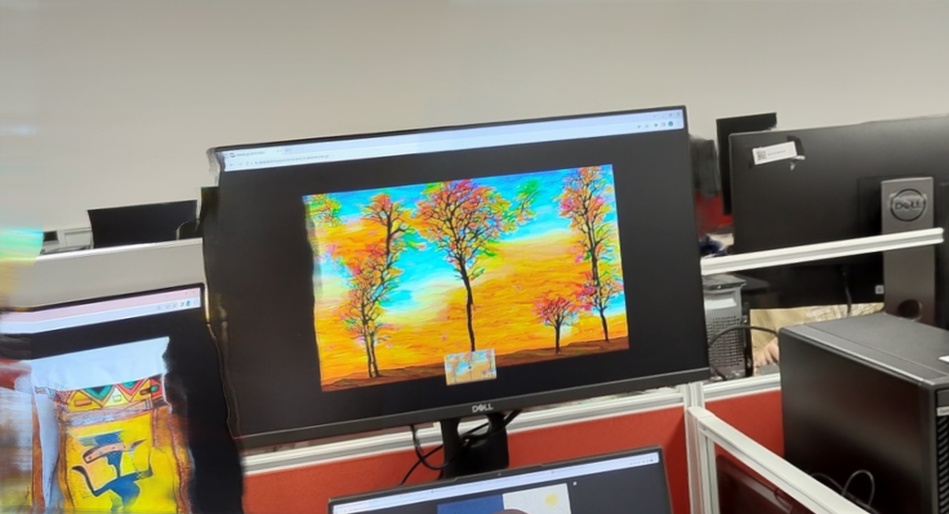} \\
Ground truth & MVSNeRF~\cite{chen2021mvsnerf} & MVSNeRF + ft & ENeRF~\cite{lin2022efficient} & ENeRF + ft \\
 & \includegraphics[width=0.2\textwidth]{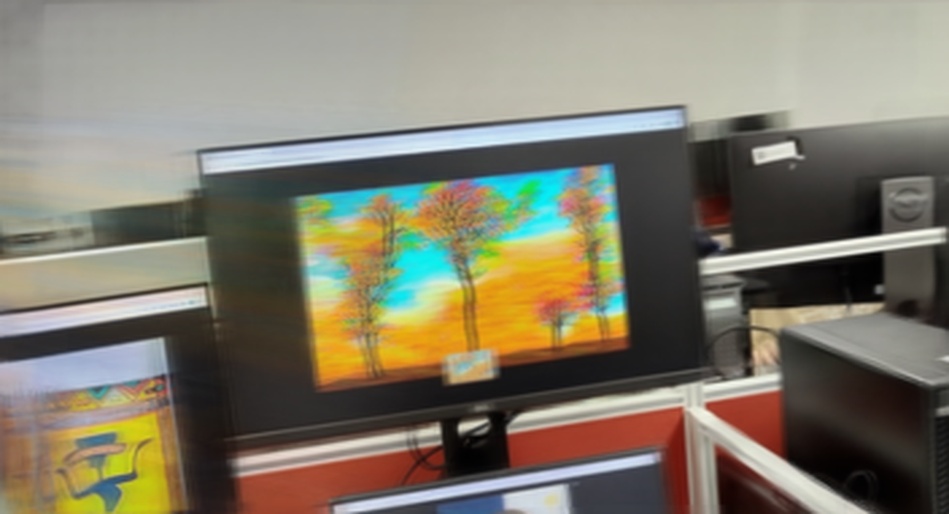} & \includegraphics[width=0.2\textwidth]{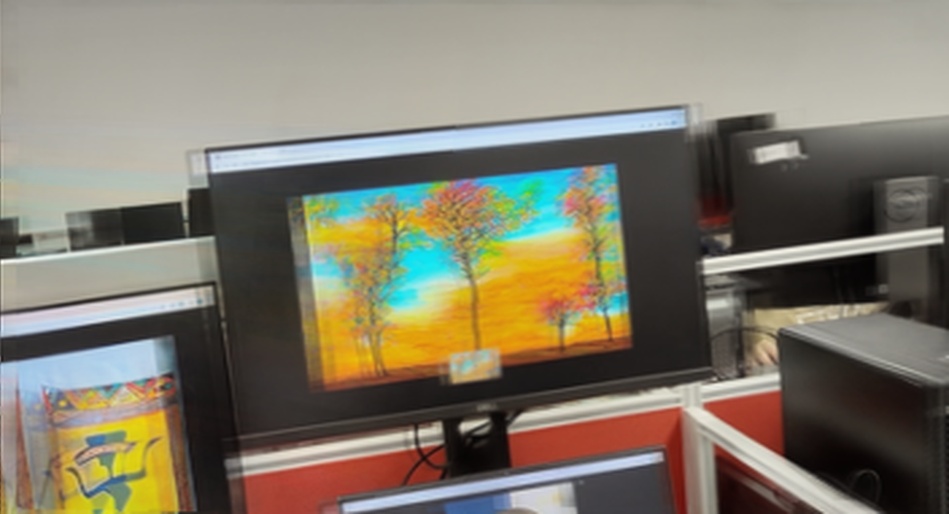} & \includegraphics[width=0.2\textwidth]{figures/results/lab_72_0_enerf_ours.jpg} & \includegraphics[width=0.2\textwidth]{figures/results/lab_72_0_enerf_ours_ft.jpg}\\
 & MVSNeRF + \textbf{ours} & MVSNeRF + \textbf{ours} + ft & ENeRF + \textbf{ours} & ENeRF + \textbf{ours} + ft \\
\includegraphics[width=0.2\textwidth]{figures/results/grass_128_0_gt.jpg} & \includegraphics[width=0.2\textwidth]{figures/results/grass_128_0_mvsnerf.jpg} & \includegraphics[width=0.2\textwidth]{figures/results/grass_128_0_mvsnerf_ft.jpg} & \includegraphics[width=0.2\textwidth]{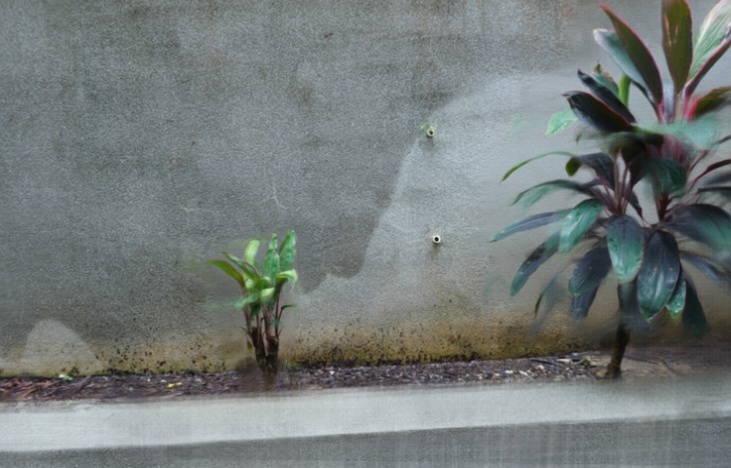} & \includegraphics[width=0.2\textwidth]{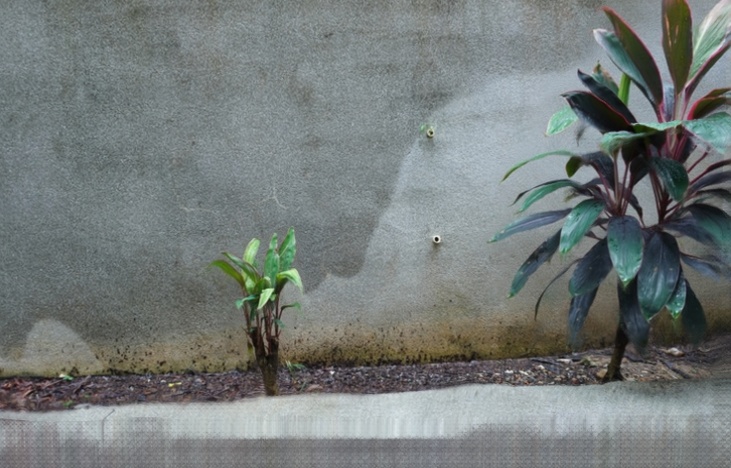} \\
Ground truth & MVSNeRF~\cite{chen2021mvsnerf} & MVSNeRF + ft & ENeRF~\cite{lin2022efficient} & ENeRF + ft \\
 & \includegraphics[width=0.2\textwidth]{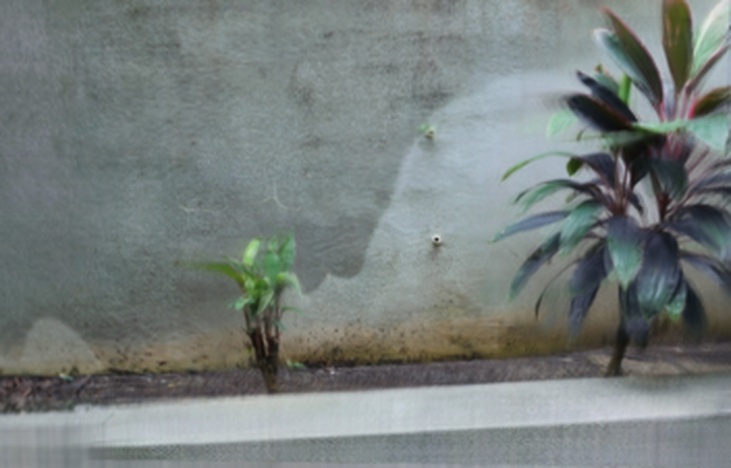} & \includegraphics[width=0.2\textwidth]{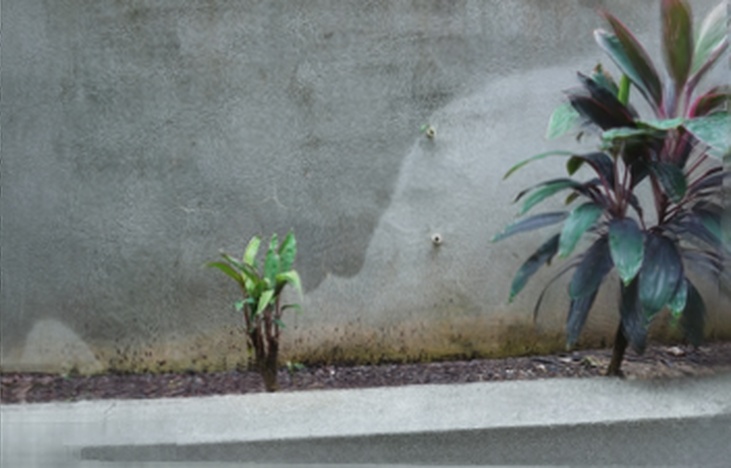} & \includegraphics[width=0.2\textwidth]{figures/results/grass_128_0_enerf_ours.jpg} & \includegraphics[width=0.2\textwidth]{figures/results/grass_128_0_enerf_ours_ft.jpg}\\
 & MVSNeRF + \textbf{ours} & MVSNeRF + \textbf{ours} + ft & ENeRF + \textbf{ours} & ENeRF + \textbf{ours} + ft \\
\end{tabular}%
}
\caption{\textbf{Additional qualitative rendering quality improvements of integrating our method into MVS-based NeRF methods on the Free dataset.}}
\label{fig:more_qualitative_boostmvs_free}
\end{figure*}

We show additional qualitative comparisons on the Free~\cite{wang2023f2} dataset in Fig.~\ref{fig:more_qualitative_free} and Fig.~\ref{fig:more_qualitative_boostmvs_free}.

\subsection{ScanNet dataset}
\begin{figure*}[t]
\centering
\small
\setlength{\tabcolsep}{1pt}
\renewcommand{\arraystretch}{1}
\resizebox{1.0\textwidth}{!} 
{
\begin{tabular}{c:ccc:ccccc}
\includegraphics[width=0.13\textwidth]{figures/results/scene0616_00_405_0_gt.jpg} & 
\includegraphics[width=0.13\textwidth]{figures/results/scene0616_00_405_0_mvsnerf.jpg} & 
\includegraphics[width=0.13\textwidth]{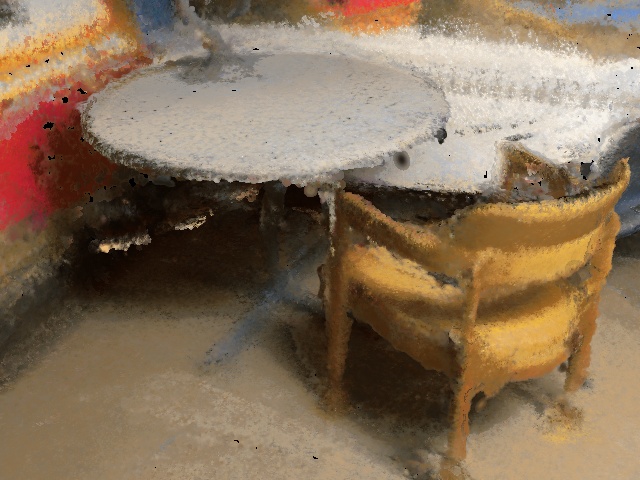} & 
\includegraphics[width=0.13\textwidth]{figures/results/scene0616_00_405_0_enerf_ours.jpg} & 
\includegraphics[width=0.13\textwidth]{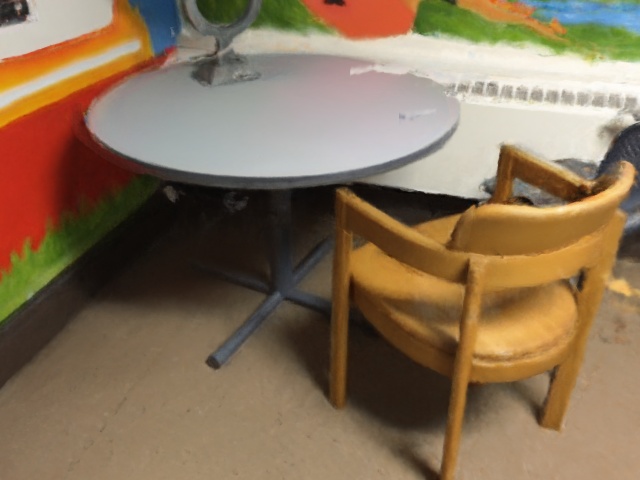} & 
\includegraphics[width=0.13\textwidth]{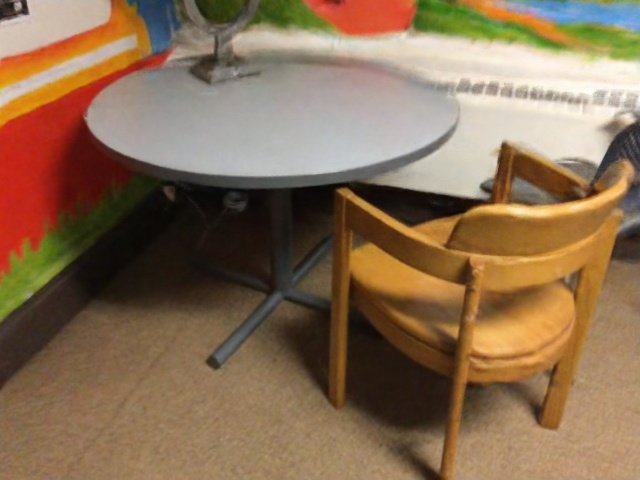} & 
\includegraphics[width=0.13\textwidth]{figures/results/scene0616_00_405_0_mvsnerf_ft.jpg} & 
\includegraphics[width=0.13\textwidth]{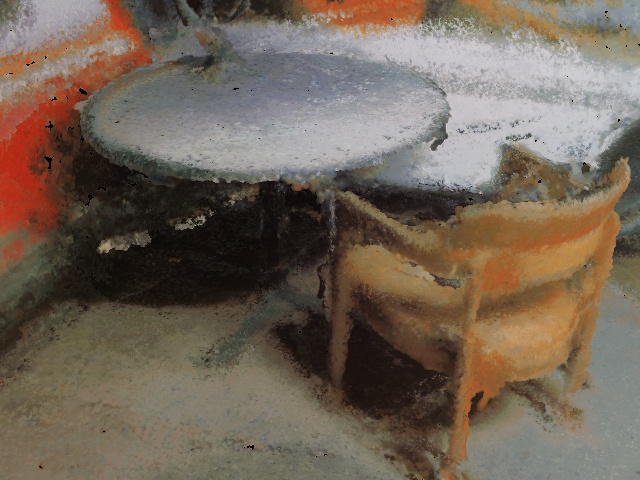} & 
\includegraphics[width=0.13\textwidth]{figures/results/scene0616_00_405_0_enerf_ours_ft.jpg} \\
\includegraphics[width=0.13\textwidth]{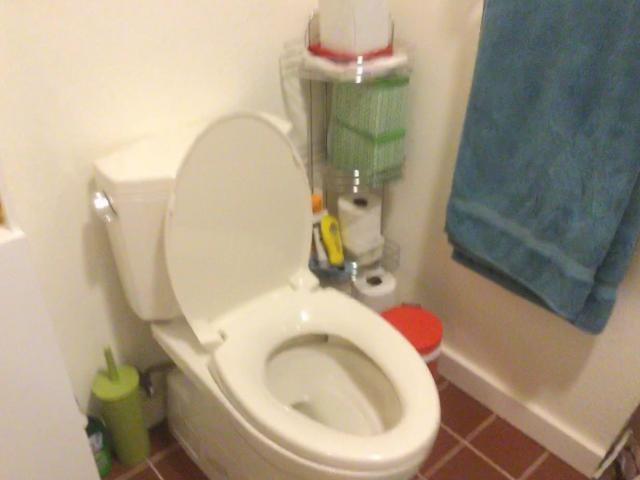} & 
\includegraphics[width=0.13\textwidth]{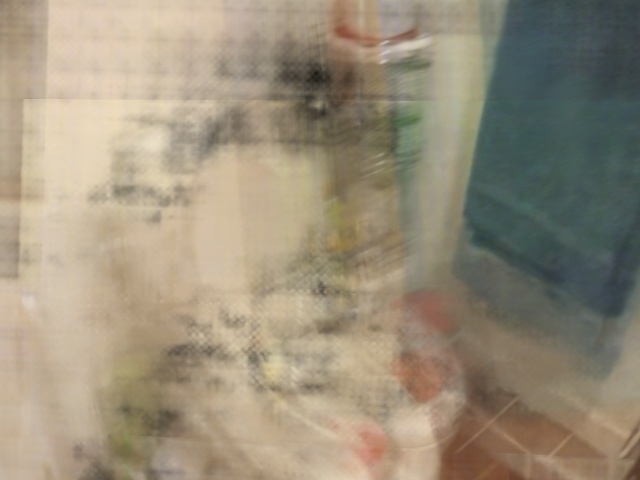} & 
\includegraphics[width=0.13\textwidth]{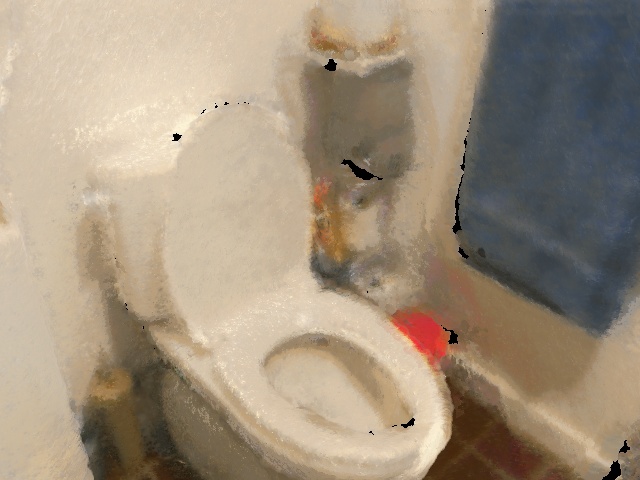} & 
\includegraphics[width=0.13\textwidth]{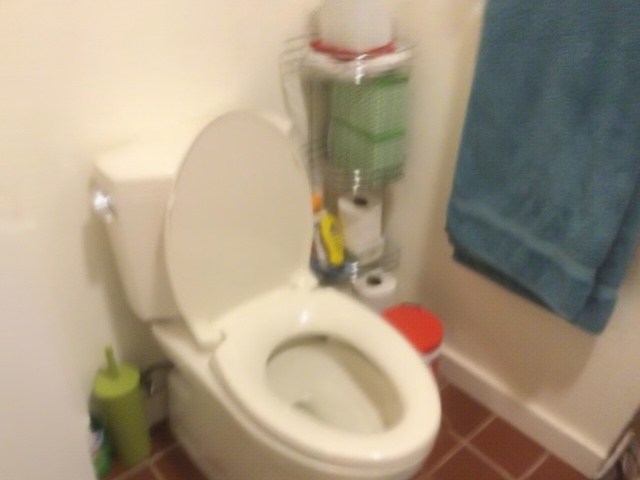} & 
\includegraphics[width=0.13\textwidth]{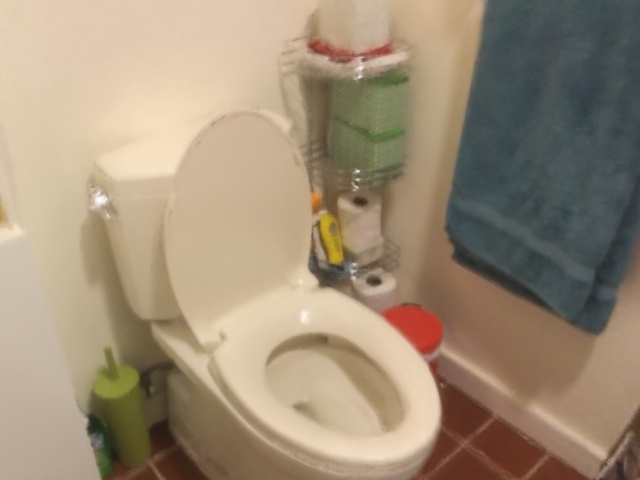} &
\includegraphics[width=0.13\textwidth]{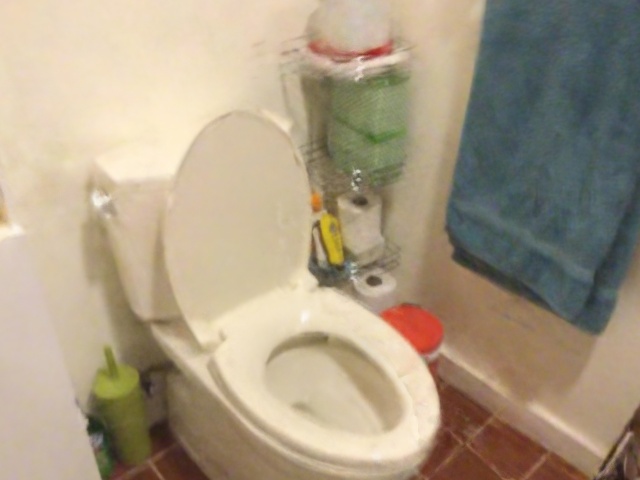} & 
\includegraphics[width=0.13\textwidth]{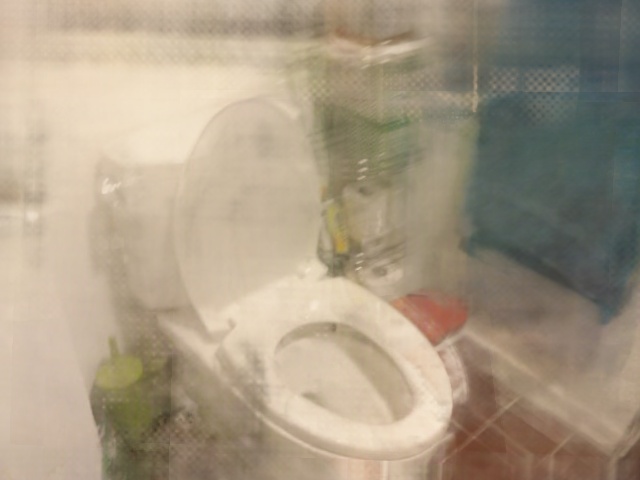} & 
\includegraphics[width=0.13\textwidth]{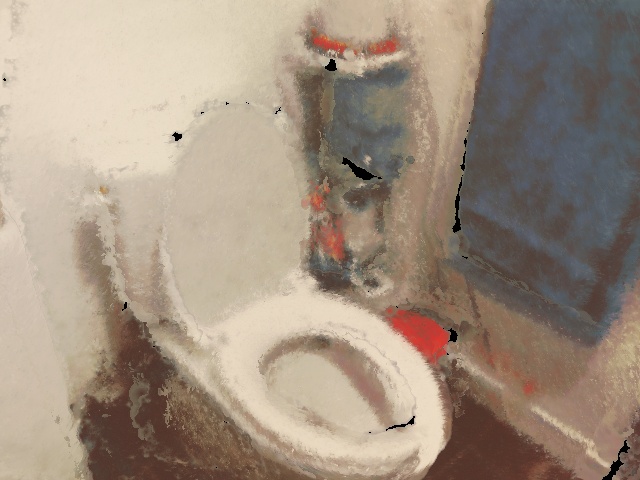} & 
\includegraphics[width=0.13\textwidth]{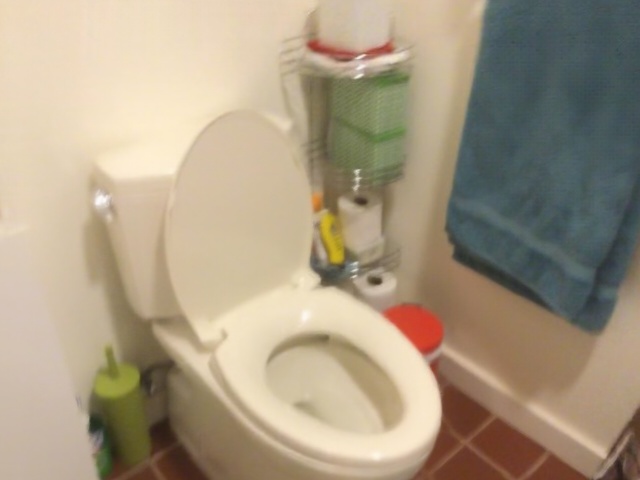} \\
\includegraphics[width=0.13\textwidth]{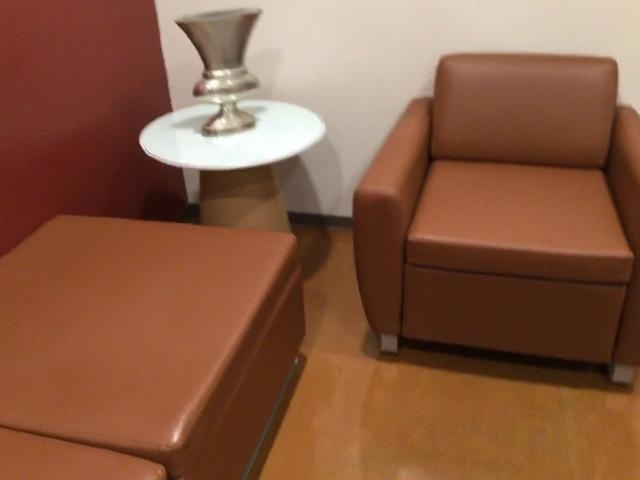} & 
\includegraphics[width=0.13\textwidth]{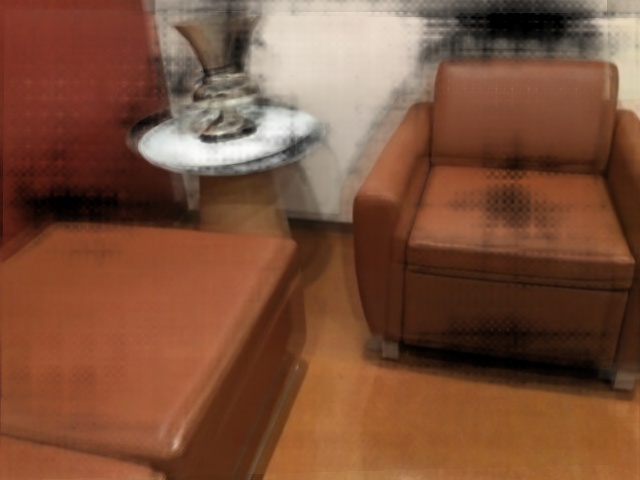} & 
\includegraphics[width=0.13\textwidth]{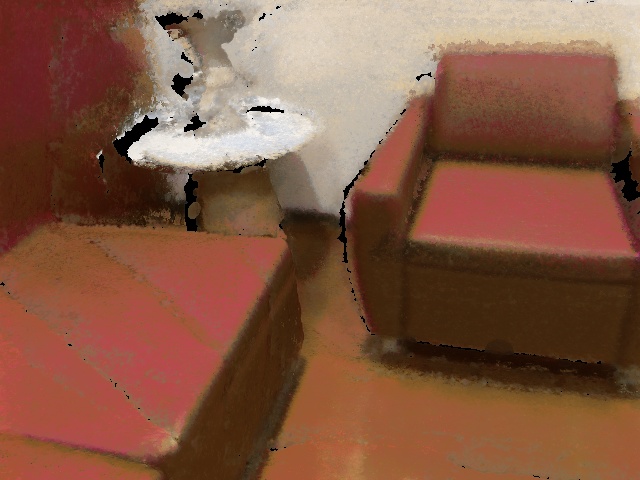} & 
\includegraphics[width=0.13\textwidth]{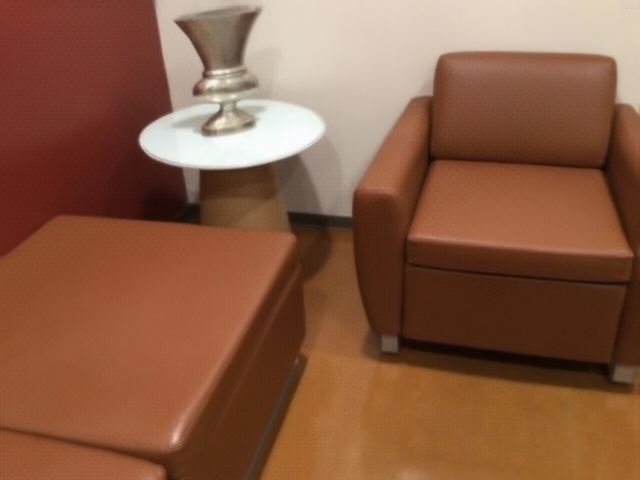} & 
\includegraphics[width=0.13\textwidth]{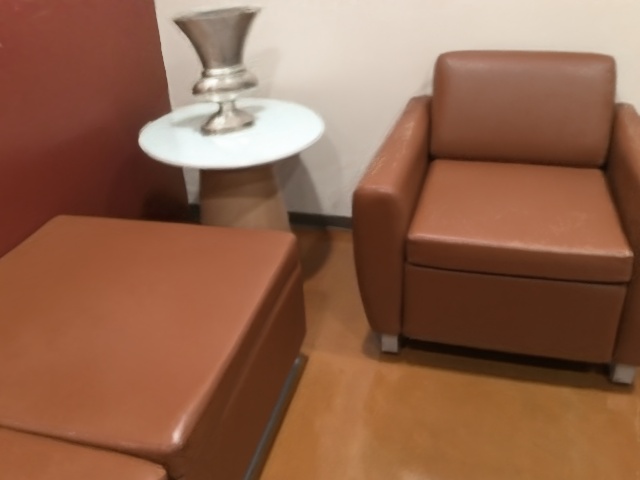} & 
\includegraphics[width=0.13\textwidth]{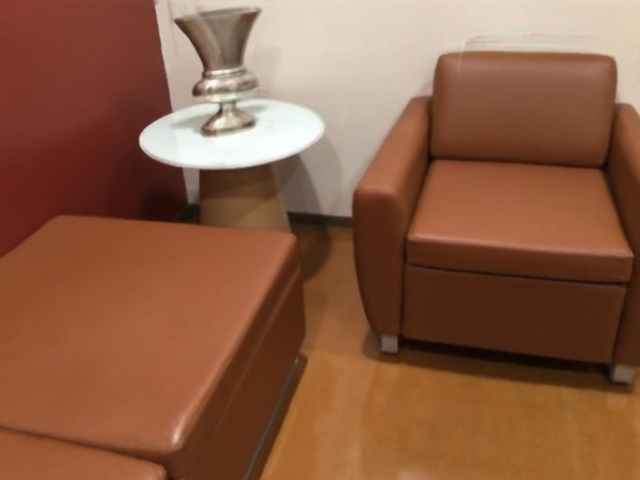} & 
\includegraphics[width=0.13\textwidth]{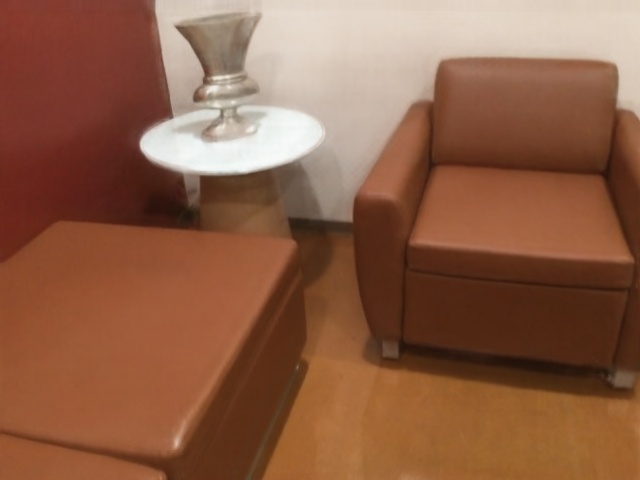} & 
\includegraphics[width=0.13\textwidth]{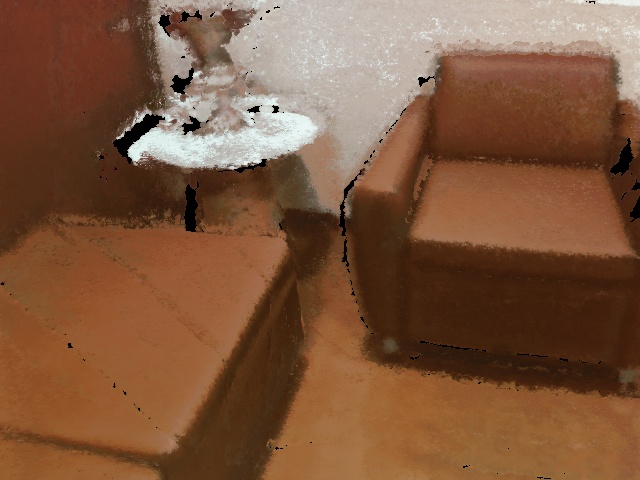} & 
\includegraphics[width=0.13\textwidth]{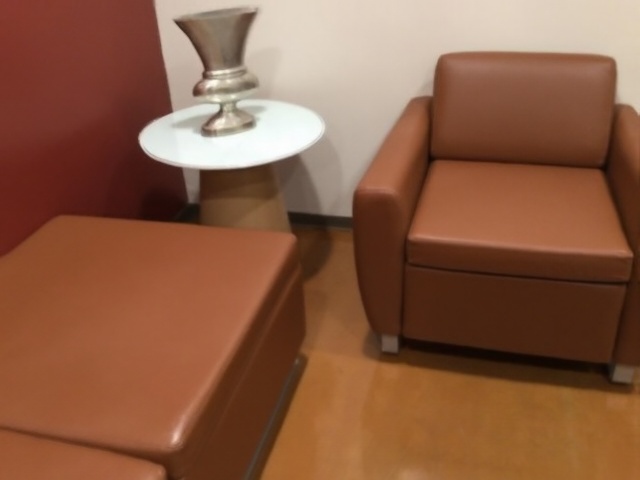} \\
\includegraphics[width=0.13\textwidth]{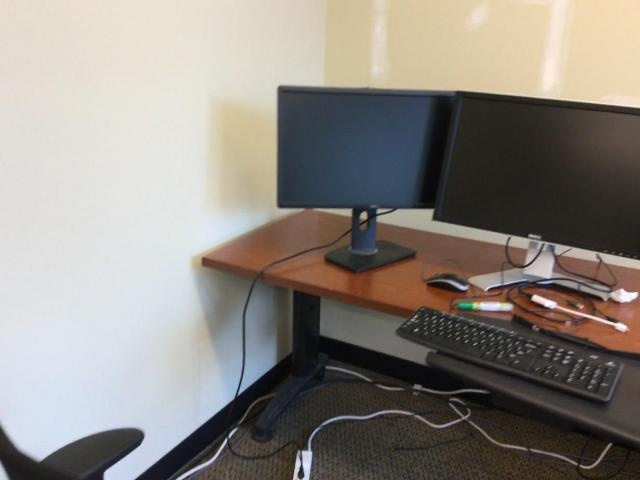} & 
\includegraphics[width=0.13\textwidth]{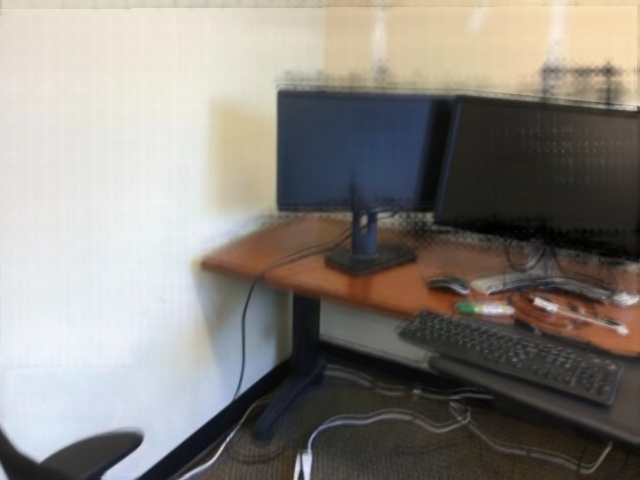} & 
\includegraphics[width=0.13\textwidth]{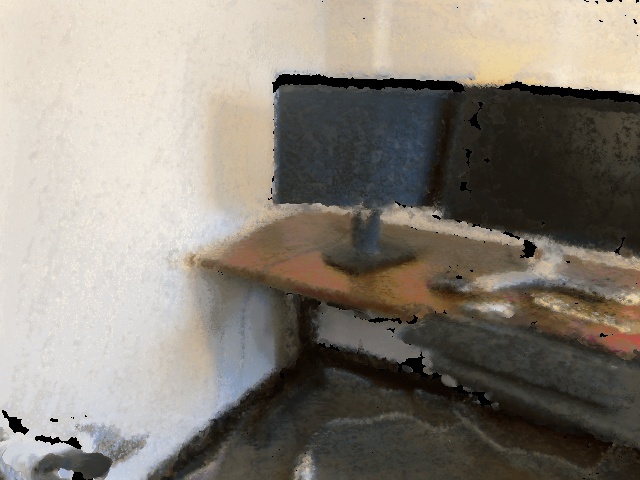} & 
\includegraphics[width=0.13\textwidth]{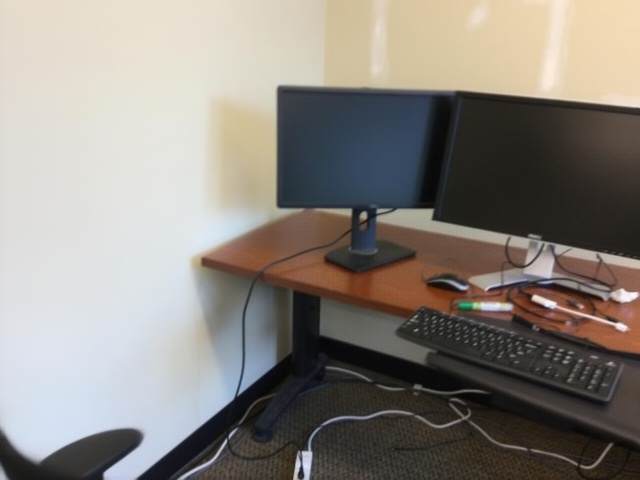} & 
\includegraphics[width=0.13\textwidth]{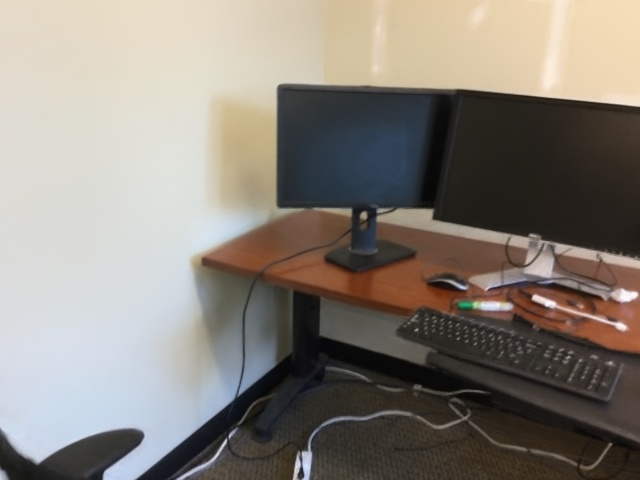} & 
\includegraphics[width=0.13\textwidth]{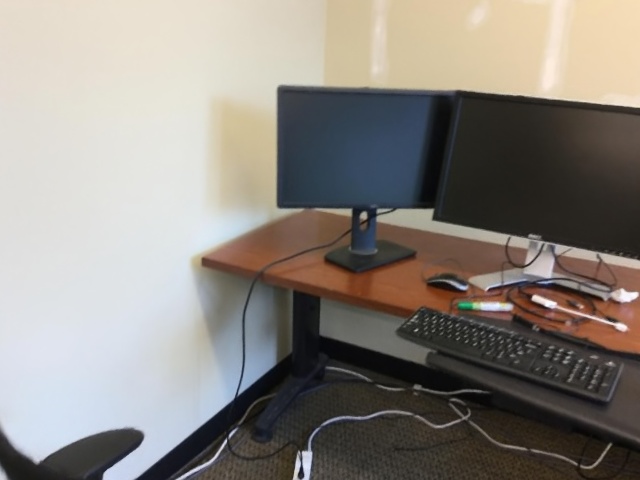} & 
\includegraphics[width=0.13\textwidth]{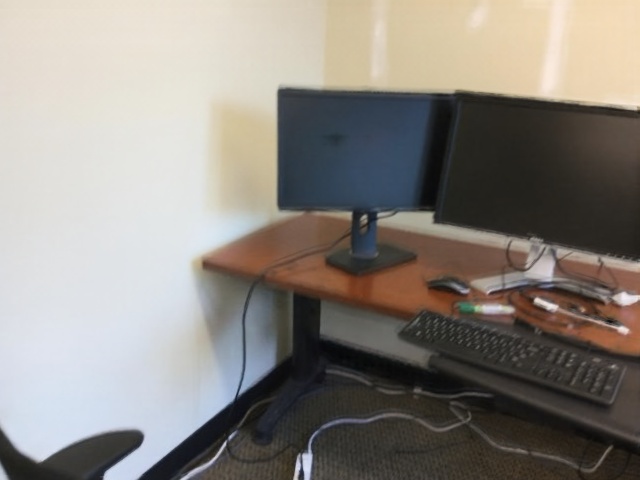} & 
\includegraphics[width=0.13\textwidth]{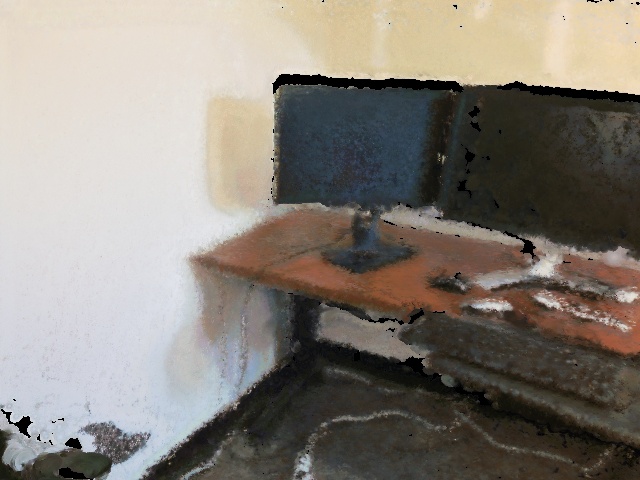} & 
\includegraphics[width=0.13\textwidth]{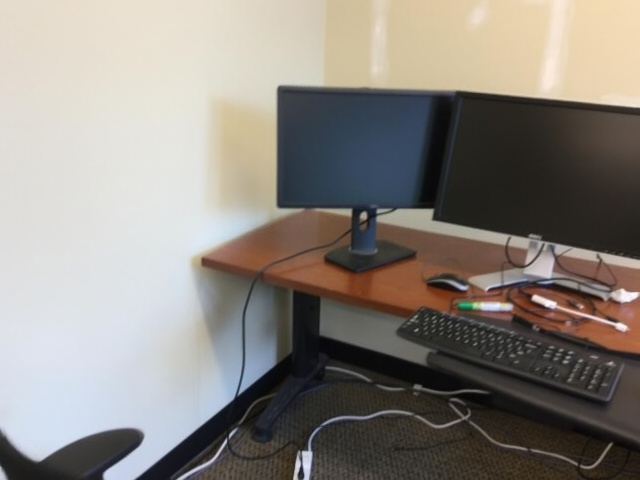} \\
\includegraphics[width=0.13\textwidth]{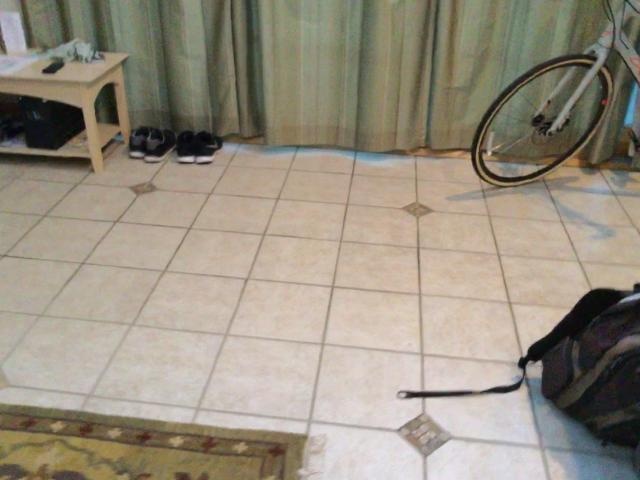} & 
\includegraphics[width=0.13\textwidth]{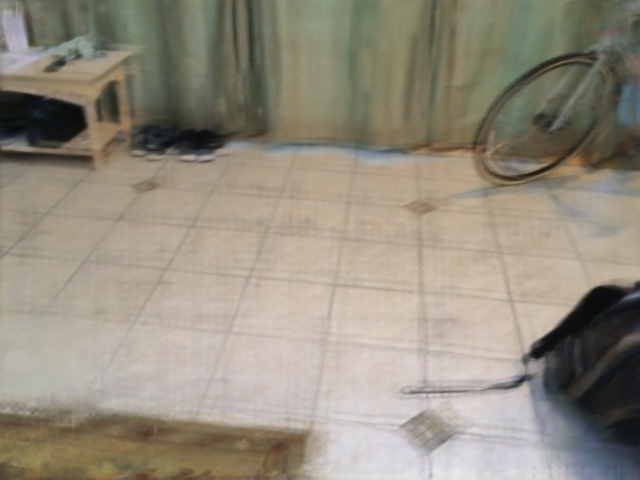} & 
\includegraphics[width=0.13\textwidth]{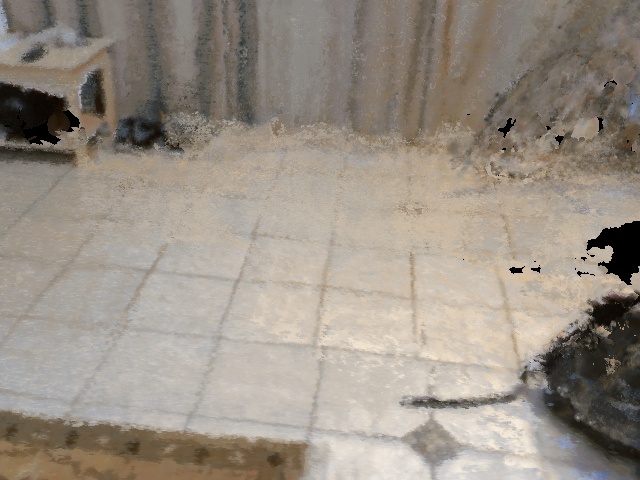} & 
\includegraphics[width=0.13\textwidth]{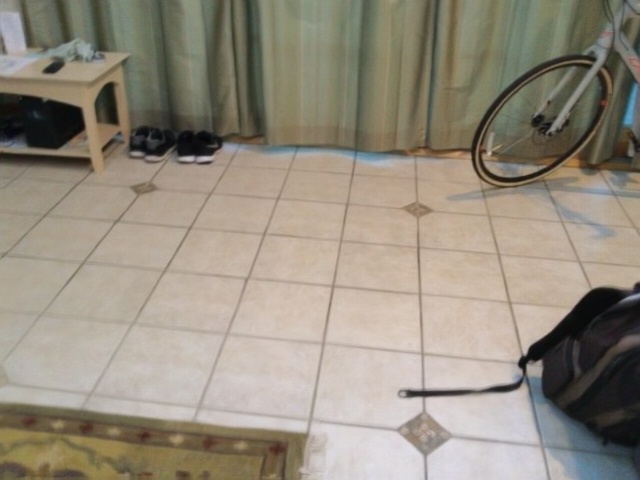} & 
\includegraphics[width=0.13\textwidth]{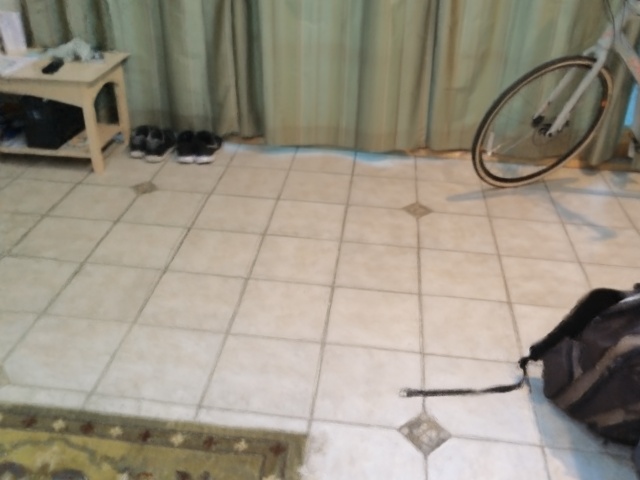} & 
\includegraphics[width=0.13\textwidth]{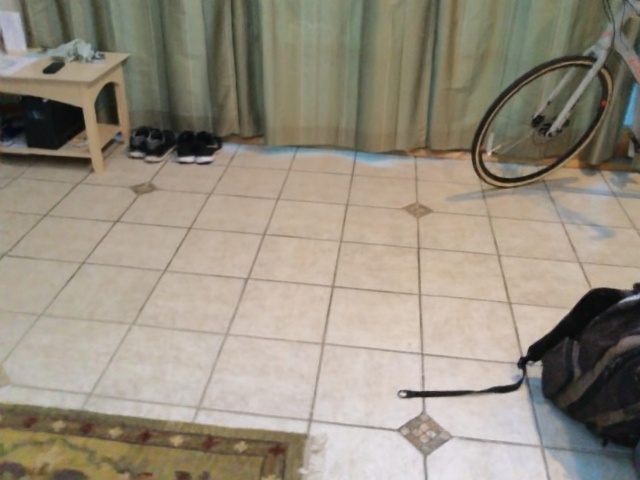} & 
\includegraphics[width=0.13\textwidth]{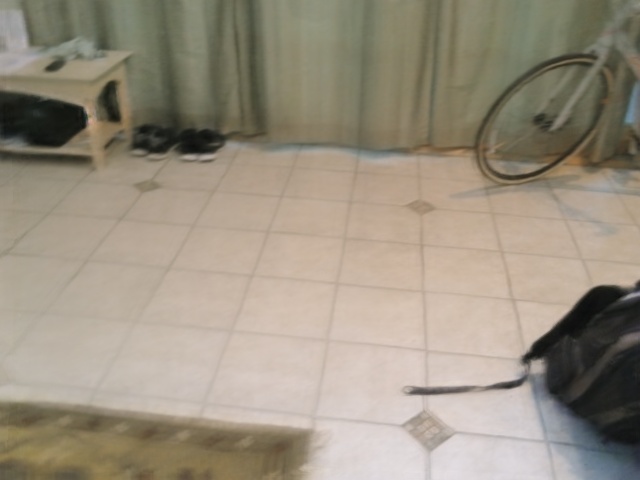} & 
\includegraphics[width=0.13\textwidth]{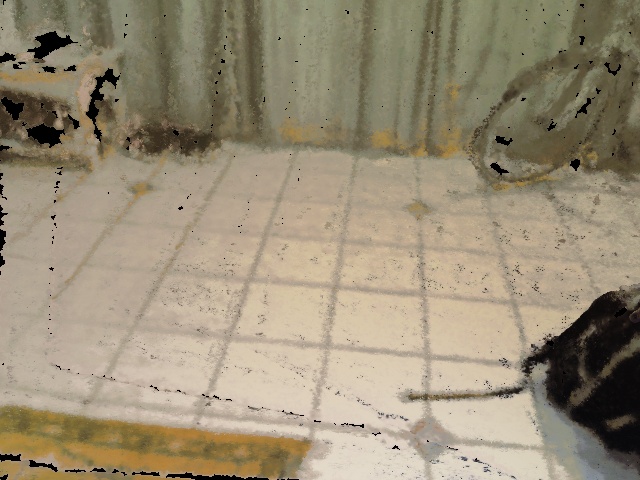} & 
\includegraphics[width=0.13\textwidth]{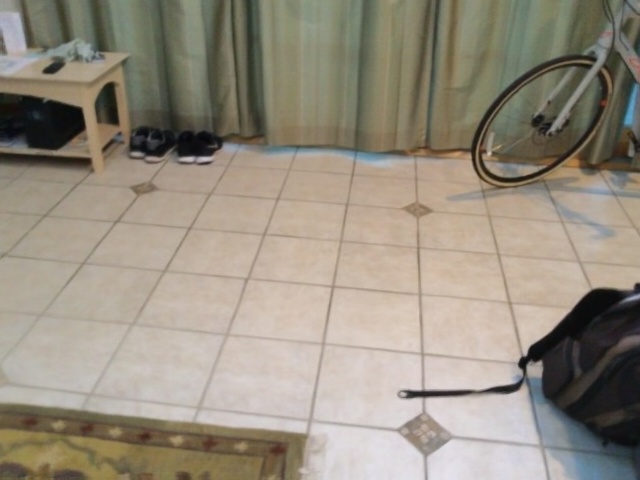} \\
\includegraphics[width=0.13\textwidth]{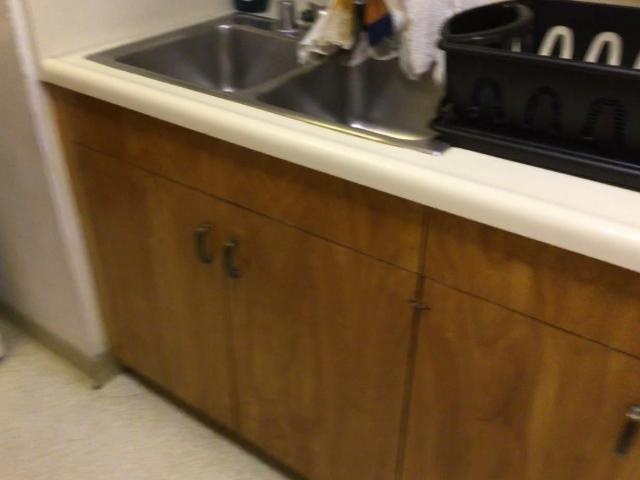} & 
\includegraphics[width=0.13\textwidth]{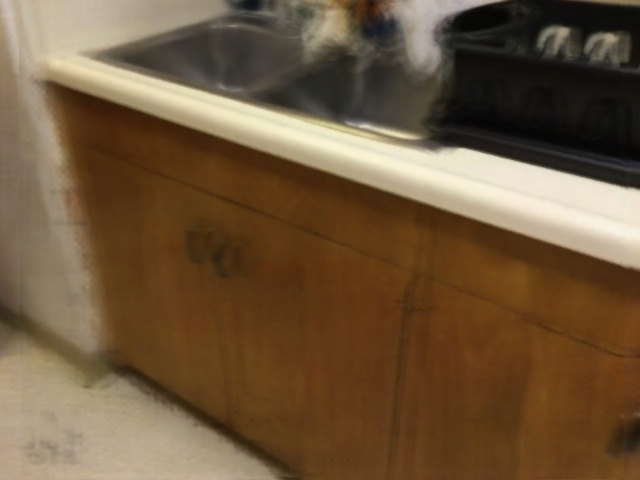} & 
\includegraphics[width=0.13\textwidth]{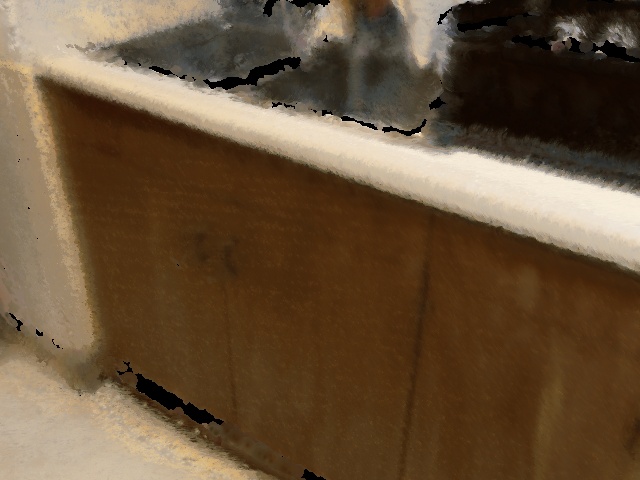} & 
\includegraphics[width=0.13\textwidth]{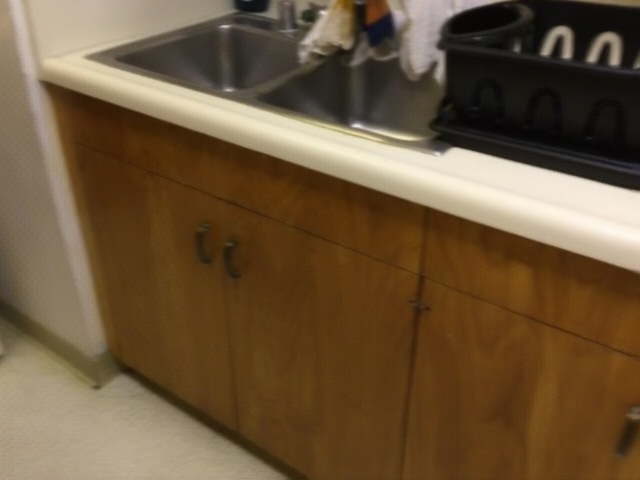} & 
\includegraphics[width=0.13\textwidth]{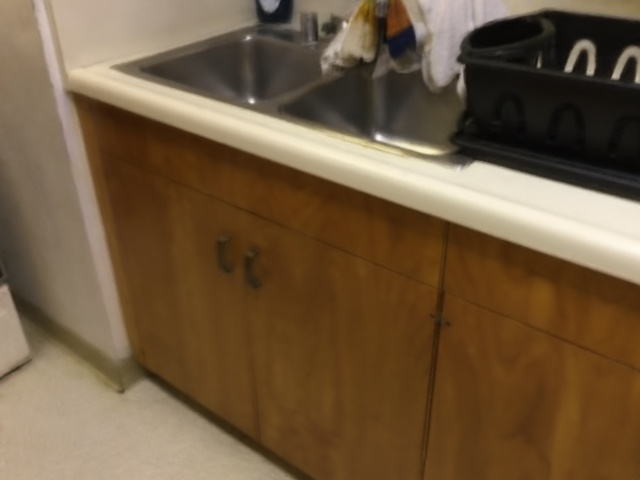} & 
\includegraphics[width=0.13\textwidth]{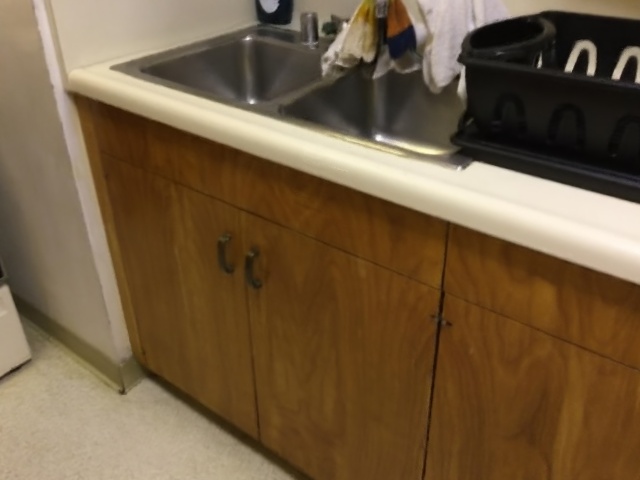} & 
\includegraphics[width=0.13\textwidth]{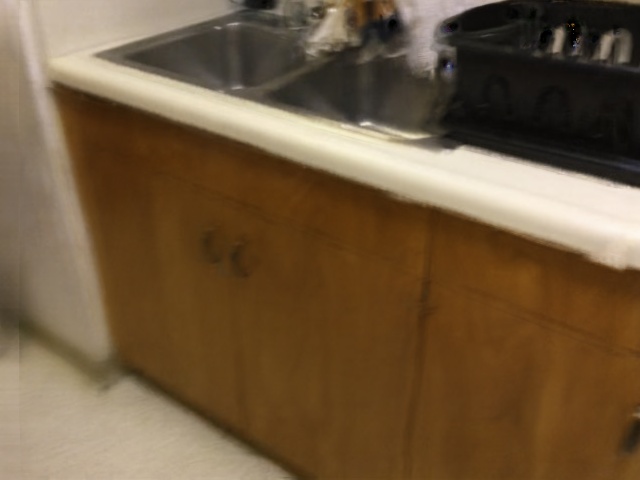} & 
\includegraphics[width=0.13\textwidth]{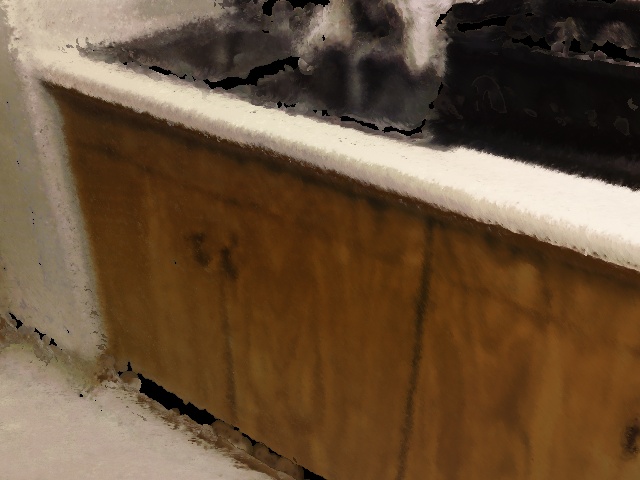} & 
\includegraphics[width=0.13\textwidth]{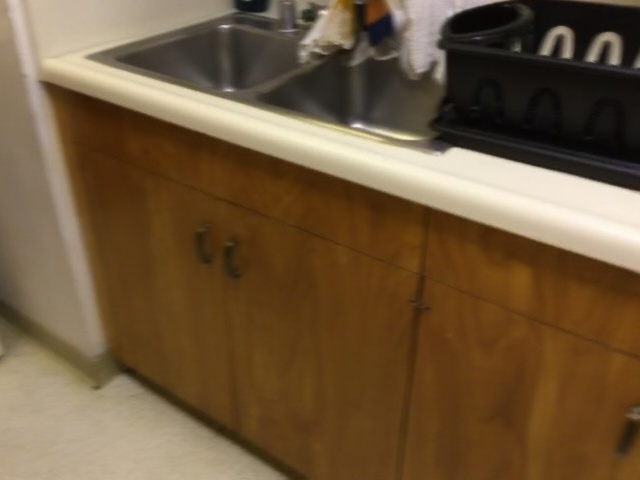} \\
\includegraphics[width=0.13\textwidth]{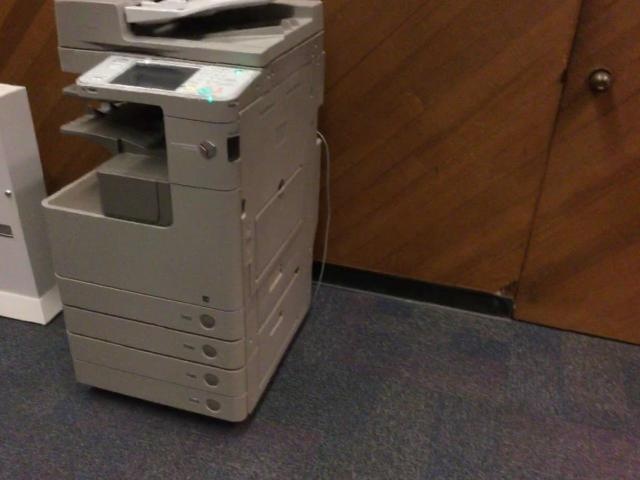} & 
\includegraphics[width=0.13\textwidth]{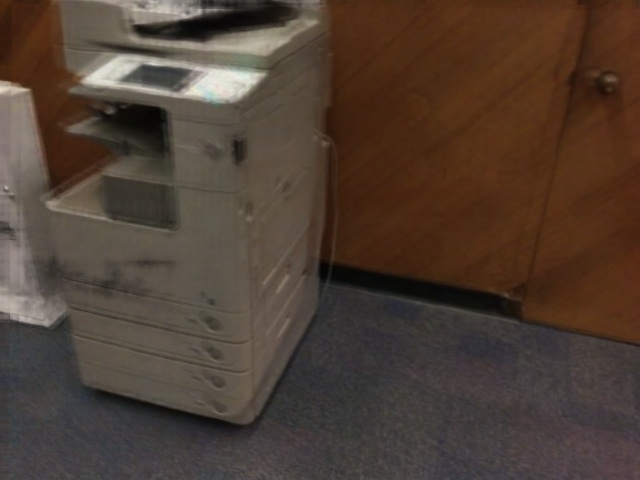} & 
\includegraphics[width=0.13\textwidth]{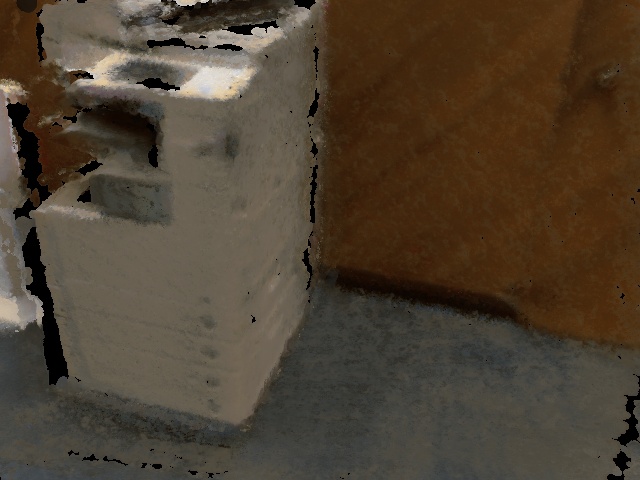} & 
\includegraphics[width=0.13\textwidth]{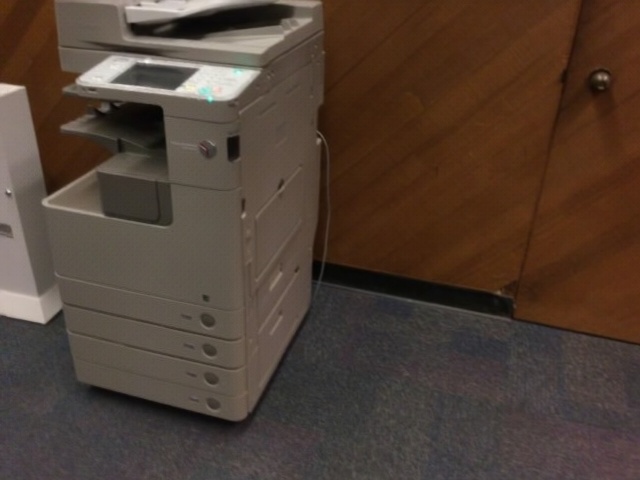} & 
\includegraphics[width=0.13\textwidth]{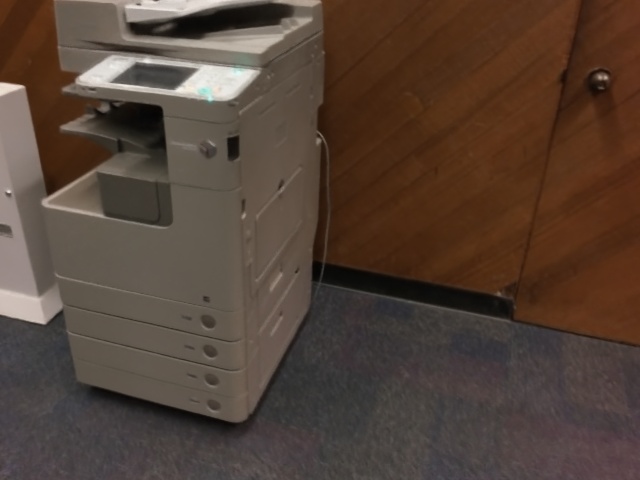} & 
\includegraphics[width=0.13\textwidth]{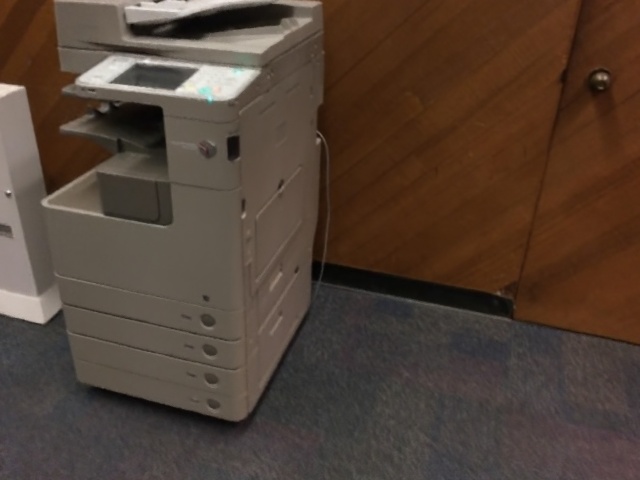} & 
\includegraphics[width=0.13\textwidth]{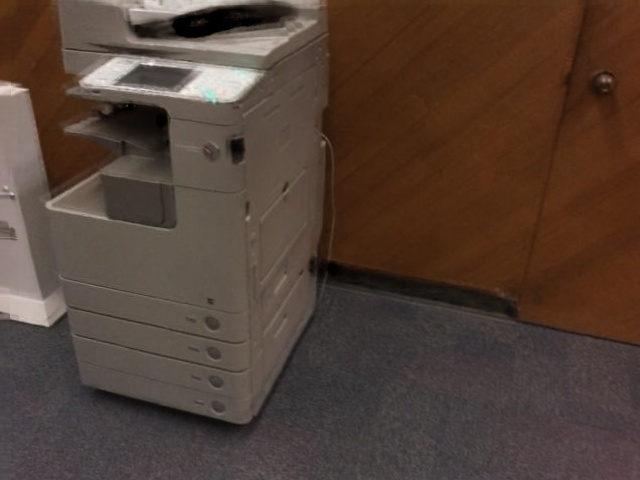} & 
\includegraphics[width=0.13\textwidth]{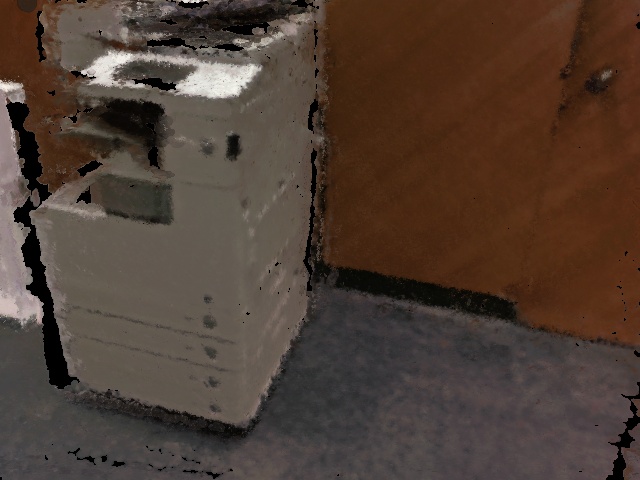} & 
\includegraphics[width=0.13\textwidth]{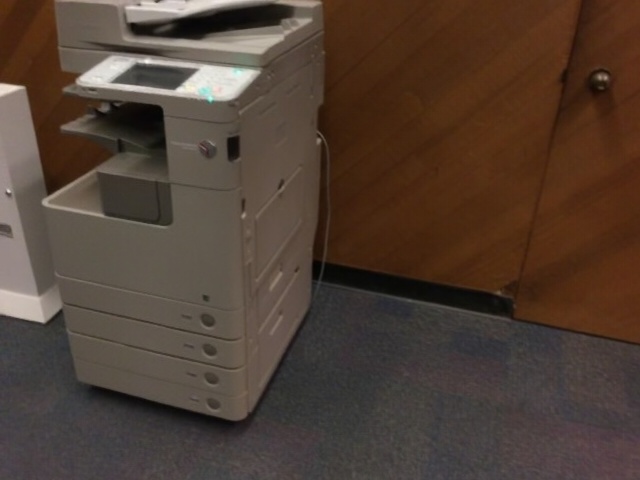} \\
\includegraphics[width=0.13\textwidth]{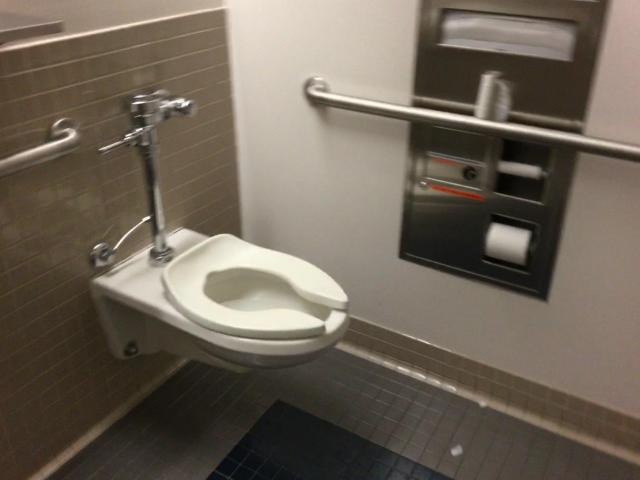} & 
\includegraphics[width=0.13\textwidth]{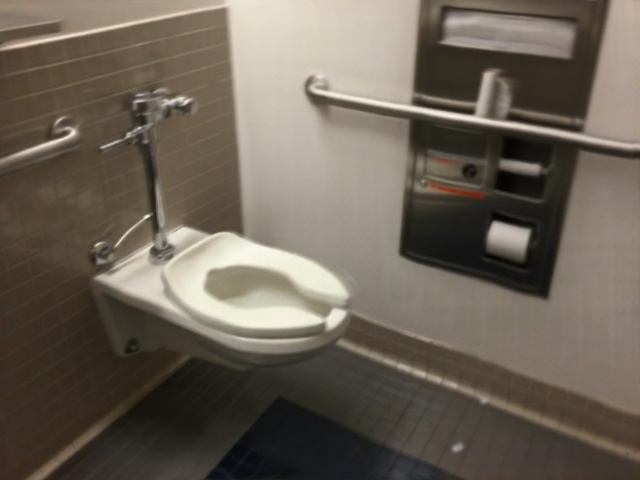} & 
\includegraphics[width=0.13\textwidth]{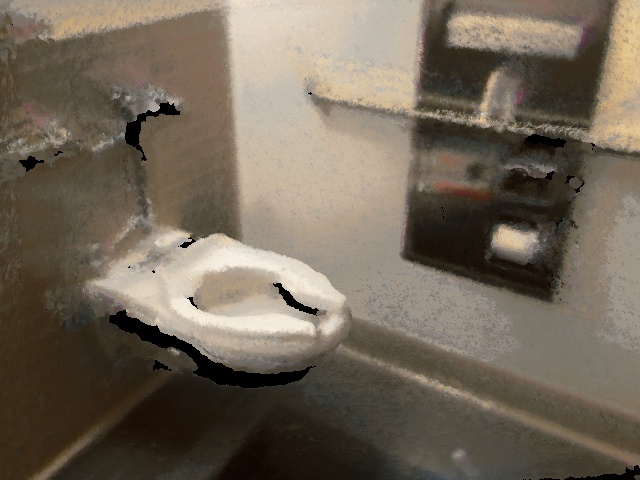} & 
\includegraphics[width=0.13\textwidth]{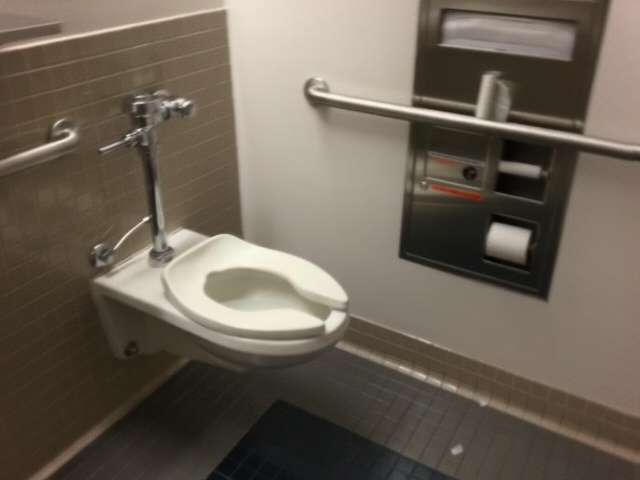} & 
\includegraphics[width=0.13\textwidth]{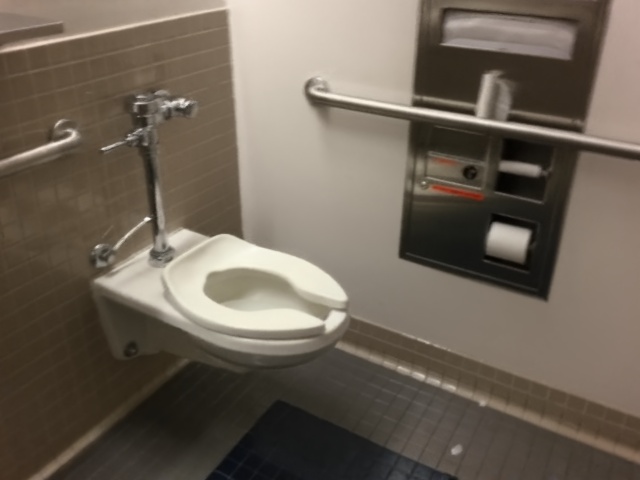} & 
\includegraphics[width=0.13\textwidth]{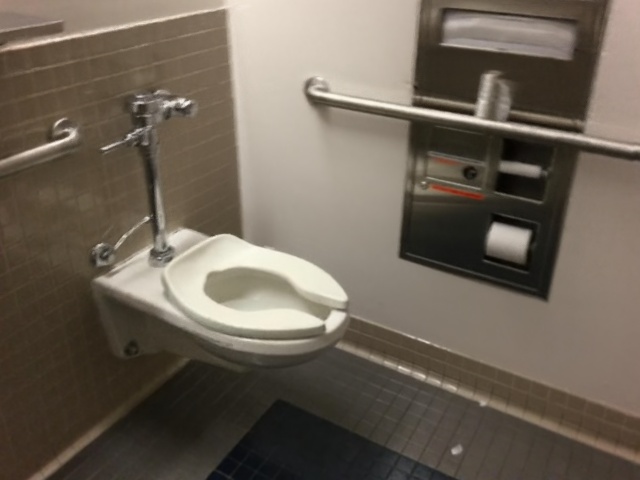} & 
\includegraphics[width=0.13\textwidth]{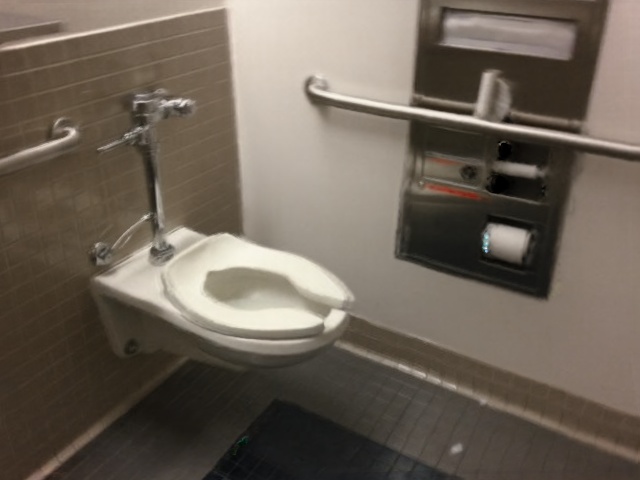} & 
\includegraphics[width=0.13\textwidth]{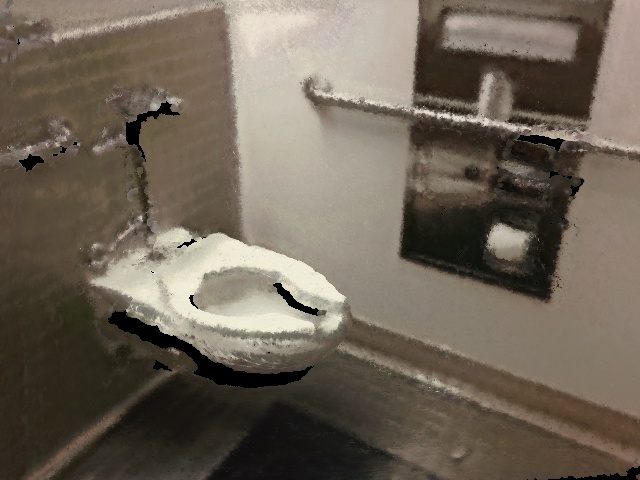} & 
\includegraphics[width=0.13\textwidth]{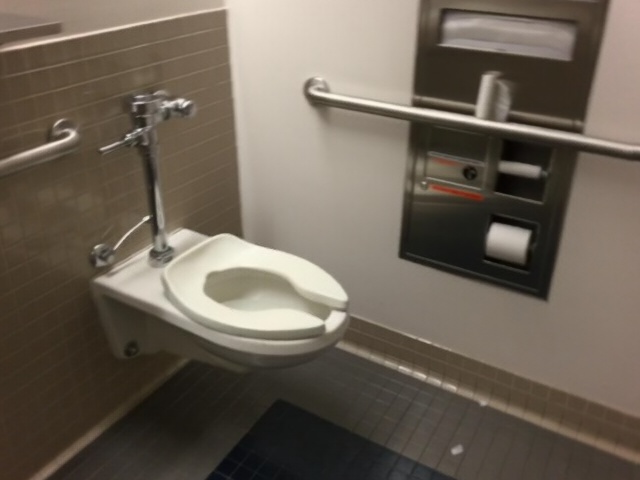} \\
Ground truth & MVSNeRF & SurfelNeRF & Ours & F2-NeRF & Zip-NeRF & MVSNeRF + ft & SurfelNeRF + ft & Ours + ft \\
& \cite{chen2021mvsnerf} & \cite{gao2023surfelnerf} & & 
\cite{wang2023f2} & \cite{barron2023zip} & \cite{chen2021mvsnerf} & \cite{gao2023surfelnerf}\\
\end{tabular}%
}
\caption{\textbf{Additional qualitative comparisons of rendering quality on the ScanNet~\cite{dai2017scannet} dataset.}}
\label{fig:more_qualitative_scannet}
\end{figure*}

\begin{figure*}[t]
\centering
\small
\setlength{\tabcolsep}{1pt}
\renewcommand{\arraystretch}{1}
\resizebox{1.0\textwidth}{!} 
{
\begin{tabular}{ccccc}
\includegraphics[width=0.2\textwidth]{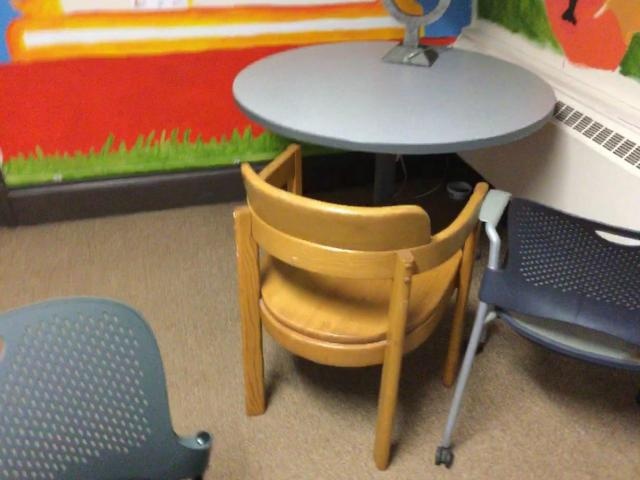} & \includegraphics[width=0.2\textwidth]{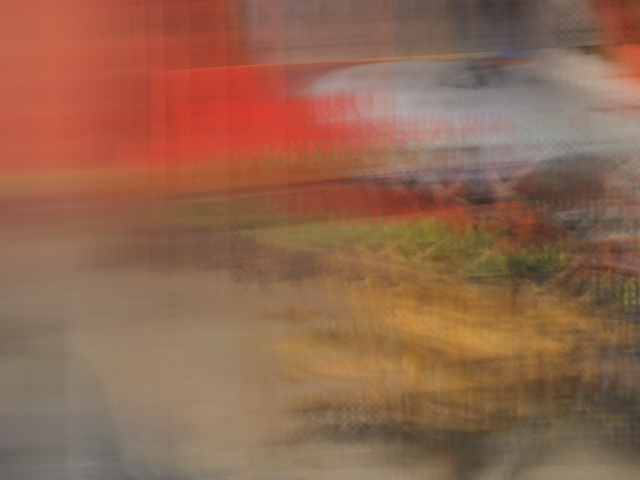} & \includegraphics[width=0.2\textwidth]{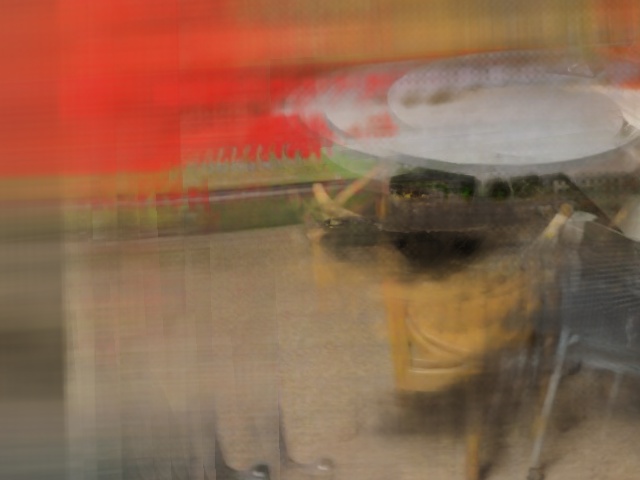} & \includegraphics[width=0.2\textwidth]{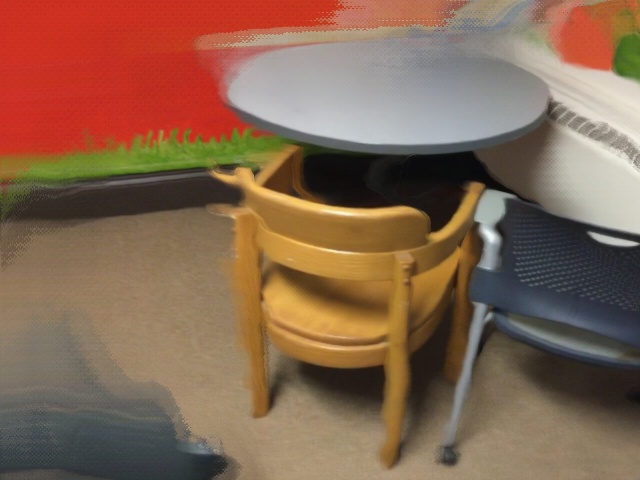} & \includegraphics[width=0.2\textwidth]{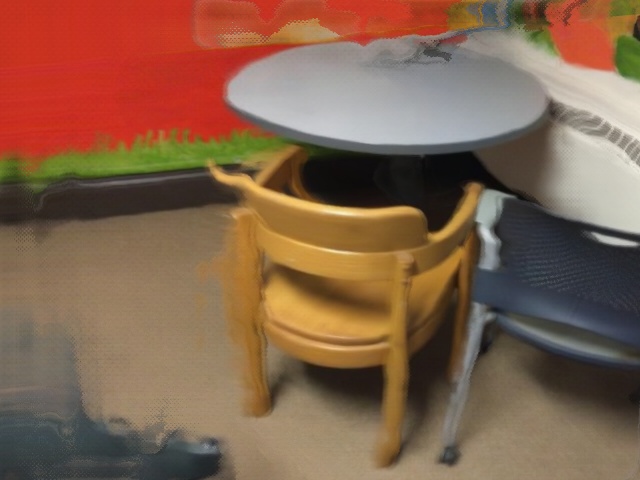} \\
Ground truth & MVSNeRF~\cite{chen2021mvsnerf} & MVSNeRF + ft & ENeRF~\cite{lin2022efficient} & ENeRF + ft \\
 & \includegraphics[width=0.2\textwidth]{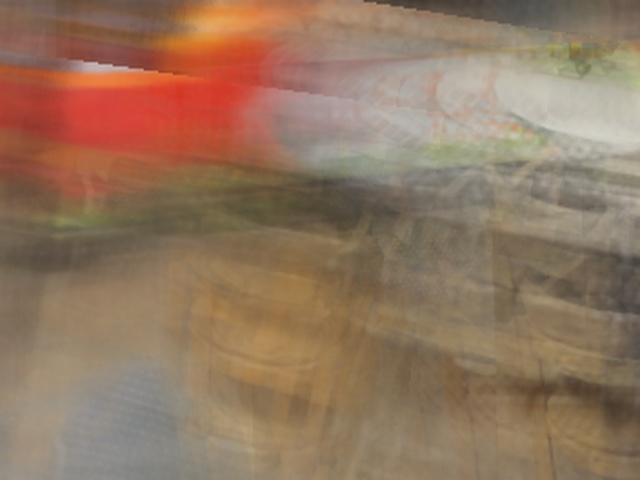} & \includegraphics[width=0.2\textwidth]{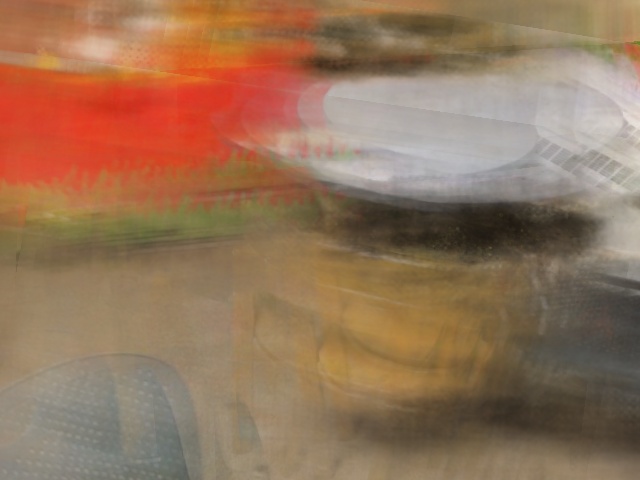} & \includegraphics[width=0.2\textwidth]{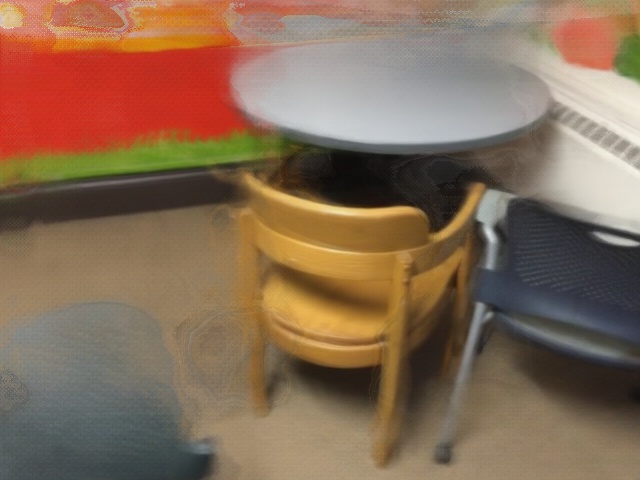} & \includegraphics[width=0.2\textwidth]{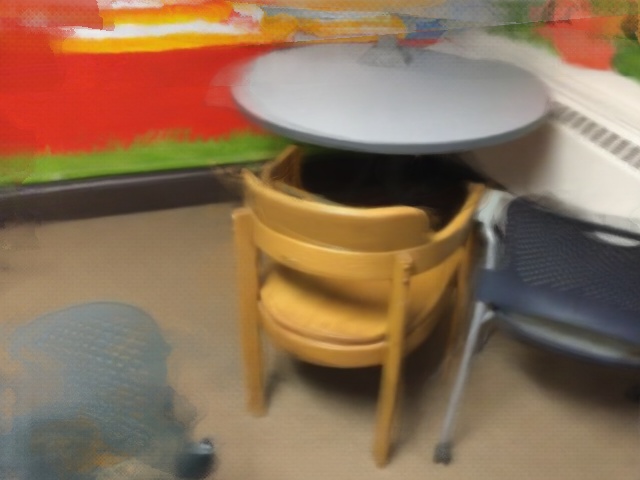}\\
 & MVSNeRF + \textbf{ours} & MVSNeRF + \textbf{ours} + ft & ENeRF + \textbf{ours} & ENeRF + \textbf{ours} + ft \\
\includegraphics[width=0.2\textwidth]{figures/results/scene0316_00_755_0_gt.jpg} & \includegraphics[width=0.2\textwidth]{figures/results/scene0316_00_755_0_mvsnerf.jpg} & \includegraphics[width=0.2\textwidth]{figures/results/scene0316_00_755_0_mvsnerf_ft.jpg} & \includegraphics[width=0.2\textwidth]{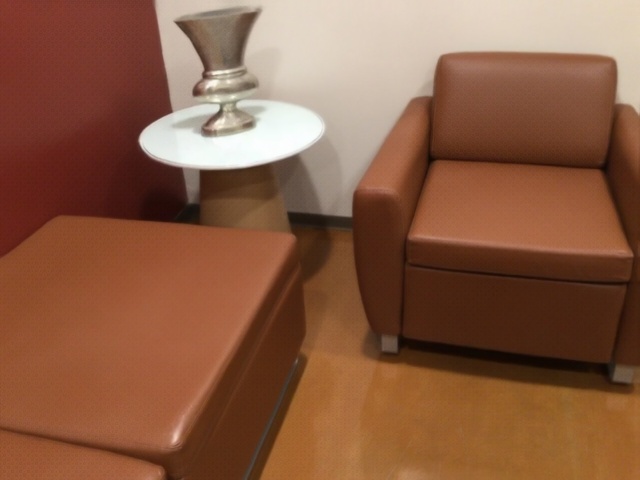} & \includegraphics[width=0.2\textwidth]{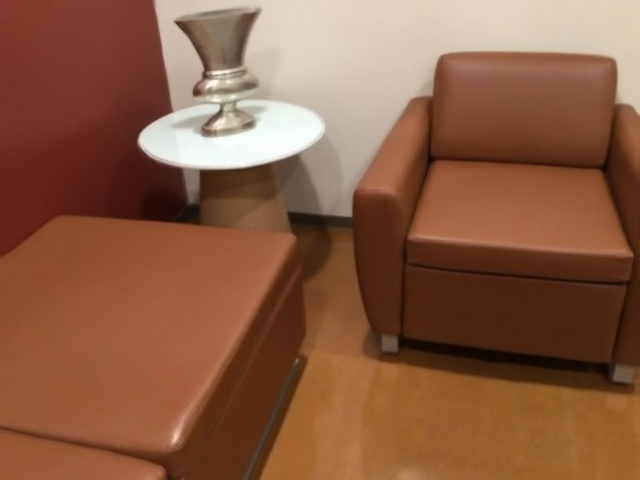} \\
Ground truth & MVSNeRF~\cite{chen2021mvsnerf} & MVSNeRF + ft & ENeRF~\cite{lin2022efficient} & ENeRF + ft \\
 & \includegraphics[width=0.2\textwidth]{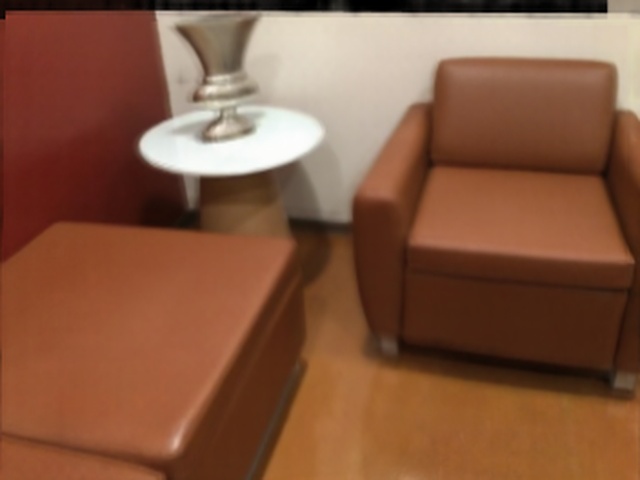} & \includegraphics[width=0.2\textwidth]{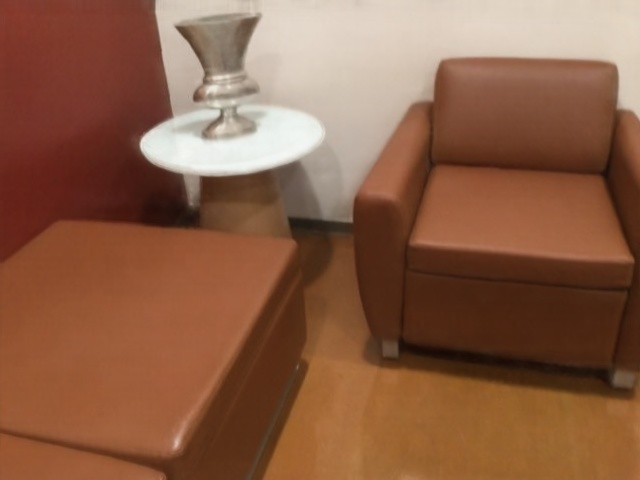} & \includegraphics[width=0.2\textwidth]{figures/results/scene0316_00_755_0_enerf_ours.jpg} & \includegraphics[width=0.2\textwidth]{figures/results/scene0316_00_755_0_enerf_ours_ft.jpg}\\
 & MVSNeRF + \textbf{ours} & MVSNeRF + \textbf{ours} + ft & ENeRF + \textbf{ours} & ENeRF + \textbf{ours} + ft \\
\includegraphics[width=0.2\textwidth]{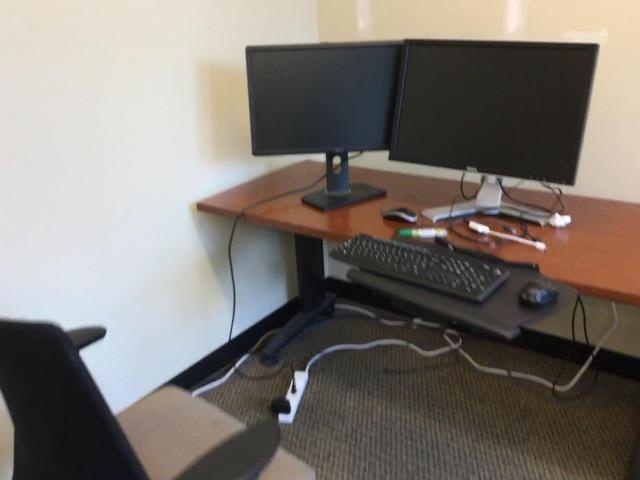} & \includegraphics[width=0.2\textwidth]{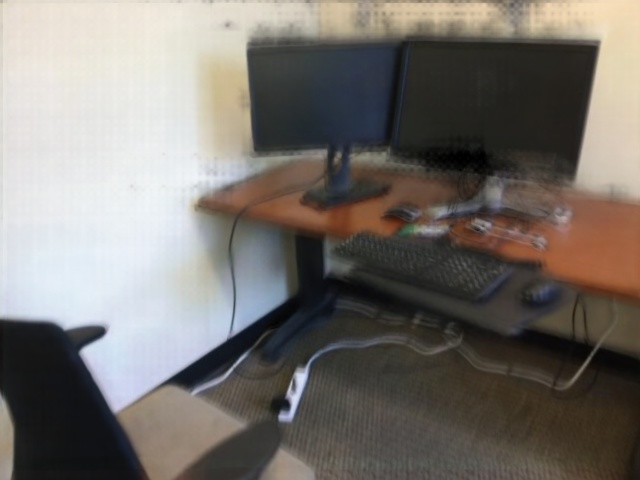} & \includegraphics[width=0.2\textwidth]{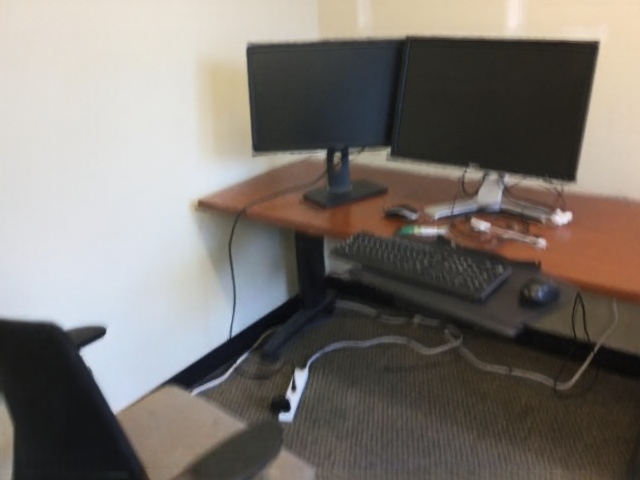} & \includegraphics[width=0.2\textwidth]{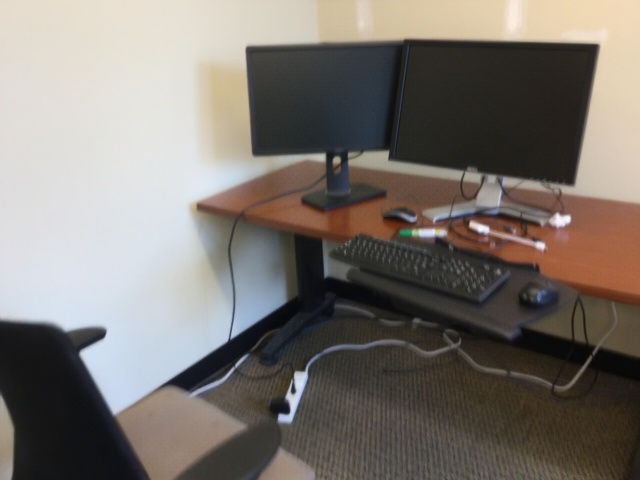} & \includegraphics[width=0.2\textwidth]{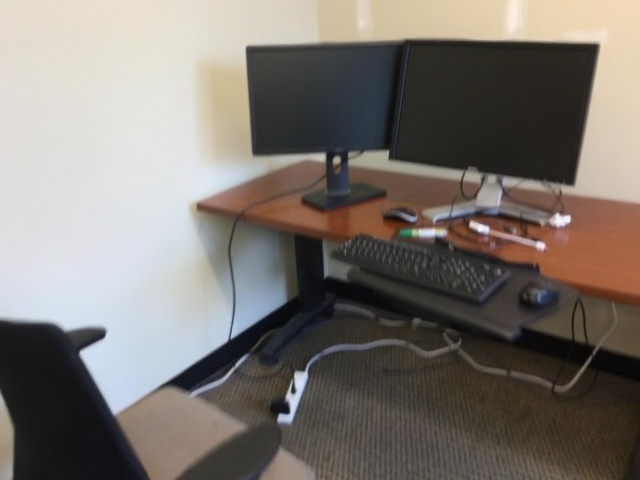} \\
Ground truth & MVSNeRF~\cite{chen2021mvsnerf} & MVSNeRF + ft & ENeRF~\cite{lin2022efficient} & ENeRF + ft \\
 & \includegraphics[width=0.2\textwidth]{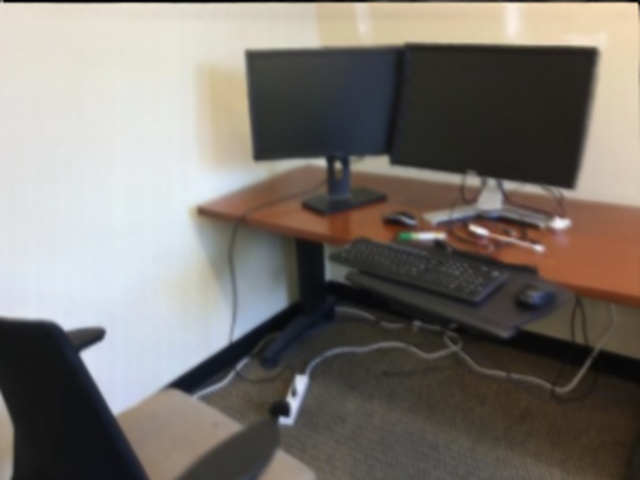} & \includegraphics[width=0.2\textwidth]{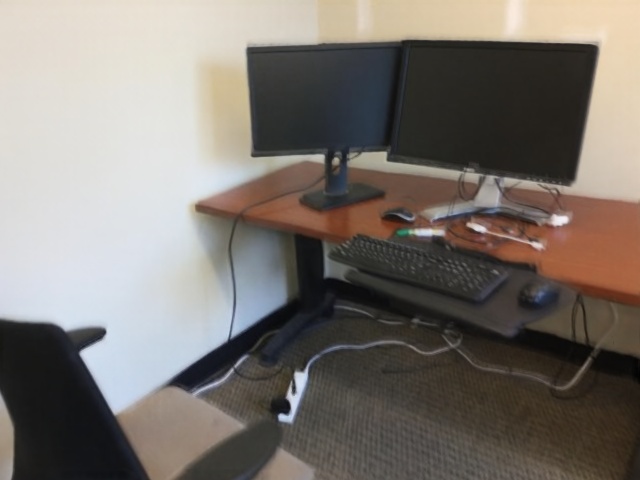} & \includegraphics[width=0.2\textwidth]{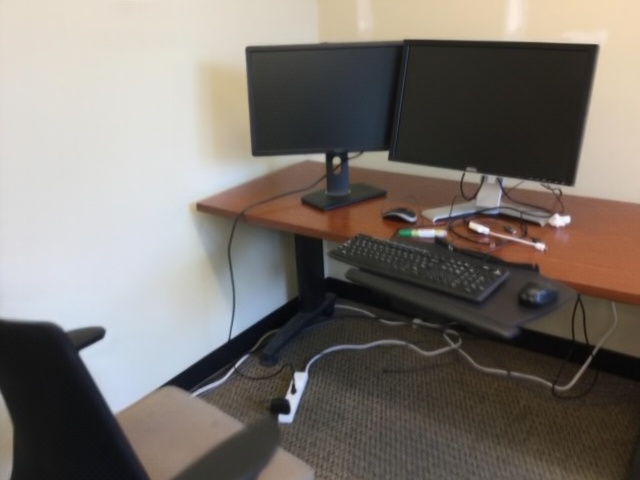} & \includegraphics[width=0.2\textwidth]{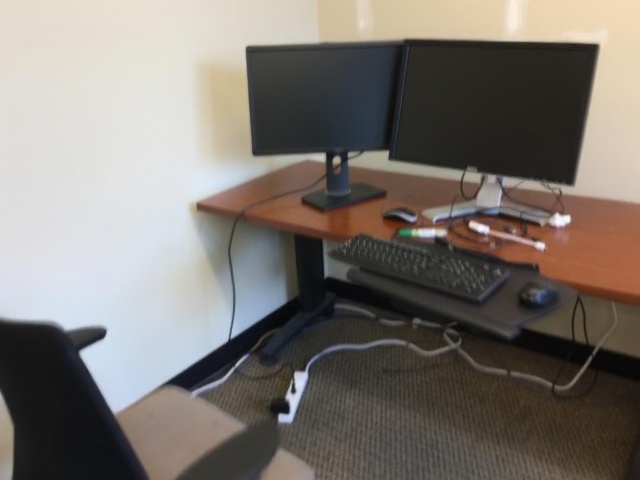}\\
 & MVSNeRF + \textbf{ours} & MVSNeRF + \textbf{ours} + ft & ENeRF + \textbf{ours} & ENeRF + \textbf{ours} + ft \\
\end{tabular}%
}
\caption{\textbf{Additional qualitative rendering quality improvements of integrating our method into MVS-based NeRF methods on the ScanNet dataset.}}
\label{fig:more_qualitative_boostmvs_scannet}
\end{figure*}

We show additional qualitative comparisons on the ScanNet~\cite{dai2017scannet} dataset, following the train/test split defined in NeRFusion, NerfingMVS, and SurfelNeRF, as depicted in Fig.~\ref{fig:more_qualitative_scannet} and Fig.~\ref{fig:more_qualitative_boostmvs_scannet}.

\end{document}